\documentclass[11pt]{article} 
\usepackage[left=1in, right=1in, top=1in, bottom=1in]{geometry}

\usepackage{amsmath}
\usepackage{amsfonts}
\usepackage{amssymb}
\usepackage{amsthm}

\usepackage{authblk}
\usepackage[linesnumbered,lined,boxed,ruled,vlined]{algorithm2e}
\usepackage{hyperref}
\usepackage{bbm}
\usepackage{url}
\usepackage{multirow}
\usepackage{comment}
\usepackage{wrapfig}
\renewcommand{\tilde}{\widetilde}

\newtheorem{lemma}{Lemma}
\counterwithin{lemma}{section}
\newtheorem{theorem}{Theorem}
\counterwithin{theorem}{section}
\newtheorem{claim}{Claim}
\counterwithin{claim}{section}
\theoremstyle{definition}
\newtheorem{definition}{Definition}
\counterwithin{definition}{section}
\newtheorem{assumption}{Assumption}
\counterwithin{assumption}{section}

\DeclareMathOperator*{\argmin}{arg\,min}

\newcommand \mb[1] {\mathbf{#1}}
\newcommand \sge[1] {#1}
\newcommand \sg[1] {#1}
\newcommand \js[1] {#1}

\newcommand{\ssize}{\mathbf{n}}

\usepackage{tikz}
\usetikzlibrary{matrix,decorations.pathreplacing,shapes,arrows.meta}

\usepackage[textsize=tiny,textwidth=2cm,color=green!50!gray]{todonotes} 

\newcommand{\cK}{\mathcal{I}}

\newcommand{\cX}{\mathcal{X}}

\newcommand{\x}{\mathbf{x}}

\usepackage[normalem]{ulem}
\usepackage{multirow}
\usepackage{tikz}

\newcommand{\bx}{\mathbf{x}}

\providecommand{\keywords}[1]
{
  \small	
  \textbf{\textit{Keywords---}} #1
}

\title{Algorithmic Challenges in Ensuring Fairness \\ at the Time of Decision}

\author[1]{Jad Salem}

\author[2]{Swati Gupta}

\author[3]{Vijay Kamble}

\affil[1]{\small US Naval Academy \protect \\
{\small \tt jsalem@usna.edu}
}
\affil[2]{\small Massachusetts Institute of Technology \protect \\
{\small \tt swatig@mit.edu}
}
\affil[3]{\small University of Illinois Chicago \protect \\
{\small \tt kamble@uic.edu}
}

\begin{document}
\maketitle

\begin{abstract}
Algorithmic decision-making in societal contexts, such as retail pricing, loan administration, recommendations on online platforms, etc., can be framed as stochastic optimization under bandit feedback, which typically requires experimentation with different decisions for the sake of learning. Such experimentation often results in perceptions of unfairness among people impacted by these decisions; for instance, there have been several recent lawsuits accusing companies that deploy algorithmic pricing practices of price gouging. Motivated by the changing legal landscape surrounding algorithmic decision-making, we introduce the well-studied fairness notion of {\it envy-freeness} within the context of stochastic convex optimization. Our notion requires that upon receiving decisions in the present time, groups do not envy the decisions received by any of the other groups, both in the present as well as the past. This results in a novel trajectory-constrained stochastic optimization problem that renders existing techniques inapplicable.

The main technical contribution of this work is to show problem settings where there is no gap in achievable regret (up to logarithmic factors) when envy-freeness is imposed. In particular, in our main result, we develop a near-optimal envy-free algorithm that achieves $\tilde{O}(\sqrt{T})$ regret for smooth convex functions that satisfy the Polyak-{\L}ojasiewicz (PL) inequality. This algorithm has a coordinate-descent structure, in which we carefully leverage gradient information to ensure monotonic sampling along each dimension, while avoiding overshooting the constrained optimum with high probability. This latter aspect critically uses smoothness and the structure of the {envy}-freeness constraints, while the PL inequality allows for sufficient progress towards the optimal solution. We discuss several open questions that arise from this analysis, which may be of independent interest.  
\end{abstract}%


\section{Introduction} 
In recent years, a wide range of organizations have deployed sophisticated data-driven algorithms that repeatedly output decisions that can have serious consequences for human stakeholders. For example, a recommendation engine shows job advertisements to a stream of arriving customers to maximize their click-through-rates and platform engagement \cite{bansal2017topic}; a retail pricing algorithm repeatedly offers a product at different prices over time to arriving customers to learn the most profitable price \cite{keskin2014dynamic}; a resume screening algorithm screens out unqualified candidates to provide the most lucrative subset of candidates to hiring managers \cite{sinha2021resume}. This widespread pursuit of efficiency and optimality often ignores a critical issue with decision-making with partial information: the perceptions of unfairness that may arise from {\it experimentation}  
that is necessary for learning good decisions in the long-run. Towards addressing this concern, the main contribution of this work is to tackle algorithmic challenges that arise from introducing the natural fairness notion of {\it envy-freeness} within the broadly applicable online decision-making framework of {\it stochastic convex optimization with bandit feedback} \cite{bubeck2015convex, hazan2016introduction}.

As a motivating example, consider a demand learning algorithm that may experiment with different prices over time to determine the optimal price for a good. While such experimentation is necessary to learn the optimal price, arbitrarily changing prices may create a sense of unfairness amongst customers, e.g., a customer may receive a price much higher than a previous customer who arrived before her for no apparent reason \cite{atlantic}.  Indeed, there has been extensive research in behavioral sciences investigating consumer perception of price fairness, which has found that notions of price fairness essentially stem from comparison: without explicit explanations, customers think they are similar to other customers buying the same item, and thus should pay equal prices \cite{darke2003fairness, bolton2003consumer, Heyman2008PerceptionsOF, haws2006dynamic, xia2004the}. In the above scenario, there is likely no palatable explanation that the firm can provide to justify the temporal disparities that arise in the search for the profit-maximizing price. 

\begin{figure}[!t] 
\centering
\begin{minipage}{\textwidth} 

  \begin{minipage}[c]{0.55\textwidth} 
    \includegraphics[trim={0 0 0 0},clip,width=\textwidth]{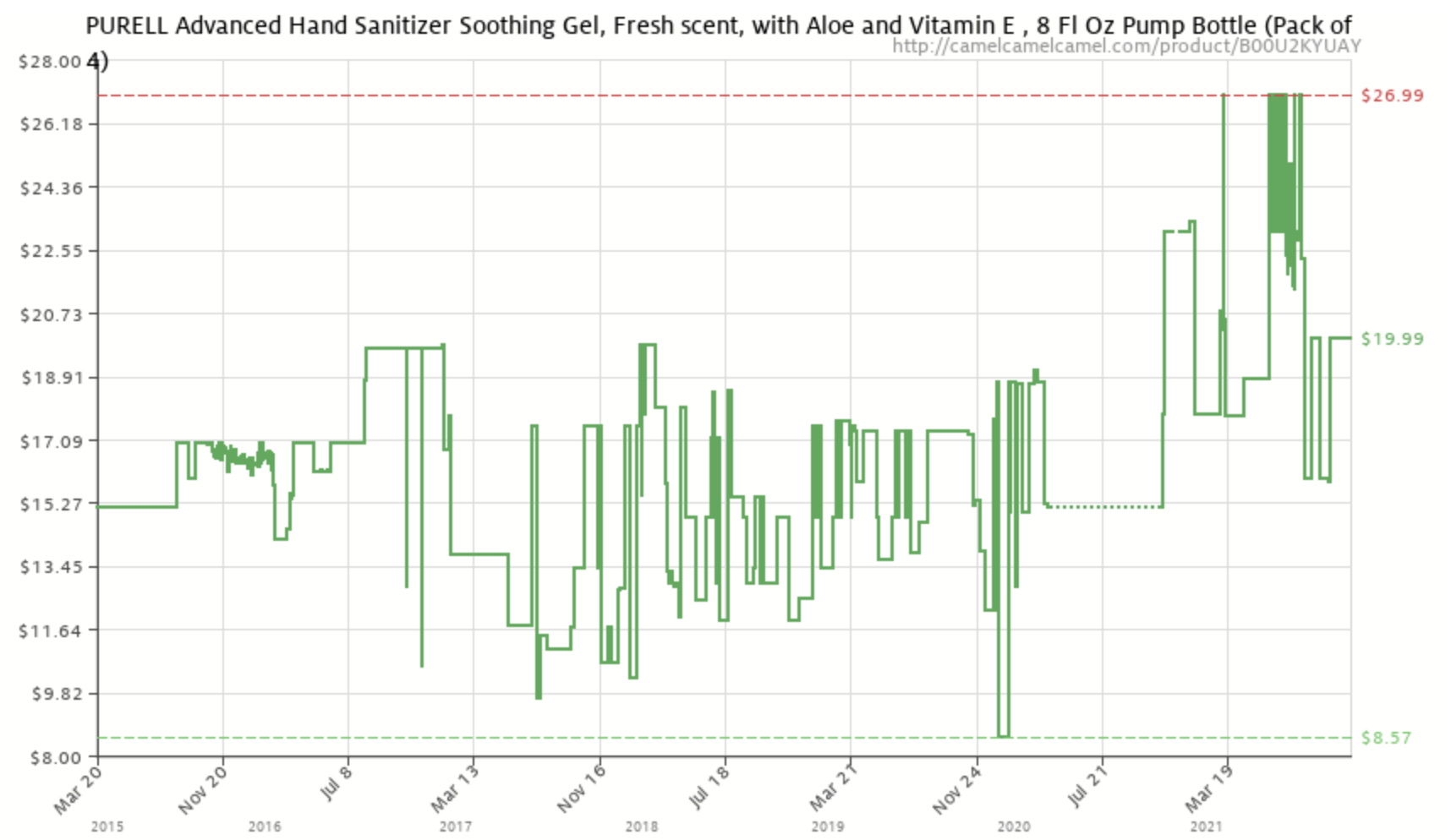} 
  \end{minipage}
  \hfill 
  \begin{minipage}[c]{0.40\textwidth} 
    \caption{The price path of Purell hand sanitizer on Amazon from March 2015 to March 2021 (from \url{https://camelcamelcamel.com/product/B00U2KYUAY}).}
    \label{fig:price-path-amazon}
  \end{minipage}

\end{minipage}
\end{figure}

Such issues with increased experimentation are further complicated by the growing litigation and policy making surrounding algorithmic decisions; e.g., Amazon was recently sued for allegedly price-gouging during the pandemic, with the claim that they drastically increased the price of essential goods (see Figure~\ref{fig:price-path-amazon} for the price path of Purell hand sanitizer on Amazon) compared to previously seen prices ({\it McQueen and Ballinger v. Amazon.com}\footnote{McQueen and Ballinger v. Amazon.com, Inc., 422 U.S. \texttt{Case 4:20-cv-02782} (2020).}). For example, this complaint alleges that Amazon prices for face masks increased by over 500\%. Such misalignment of business practices geared towards profitability, and evolving consumer protection laws (e.g., \cite{ccpa}), raises a natural question for various operational tasks: {\it how can firms enforce appropriate fairness desiderata in their online decision-making systems and what is the impact of such constraints on operational performance?} This question is non-trivial since, as we show, embedding even the simplest fairness notions in online decision-making can result in novel algorithmic challenges. 

\subsection{Modeling \sge{Envy-Freeness in}  Stochastic Convex Optimization}\label{sec:stoc-ftd} 
To study the aforementioned question, we study a canonical problem of stochastic convex optimization where a (scalar) decision is repeatedly taken for each one of a finite number of groups of individuals. In particular, we consider a sequence of decision vectors $\mb{x}_1, \mb{x}_2, \hdots, \mb{x}_T \in  \mathcal{X}^N$ taken at times $t = 1, \hdots, T$, where $\mathcal{X}=[x_{\min},x_{\max}]\subseteq \mathbb{R}$, for $N$ groups. We interpret each dimension in the decision vector as corresponding to a group, and the corresponding value is the scalar decision received by the group. This interpretation models practical decisions in societal settings, such as interest rates offered to different groups over time to minimize the default rate, salaries offered to potential hires over time to maximize the utility of a firm, price discounts offered on a product over time to maximize revenue of an unknown demand function, etc. We assume that there is a directionality in the conduciveness of decisions from the perspective of the groups, and without loss of generality, we assume that higher decisions are perceived to be better by any group (e.g., a higher salary, higher discount).

The total cost of the decisions is given by $f(\mathbf{x}) = \sum_{i\in N} f_i(\mathbf{x})$, where $f_i(\mathbf{x})$ models the cost (or equivalently, the negative utility) associated with each group. The functions $f_i$ are a priori unknown, but are known to belong to some known subclass $\mathcal{F}$ of convex functions. Our goal will be to design an algorithm to output decisions $\{\x_t\}_{t}$ that minimize $f(\mathbf{x})$ while constraining the trajectory that $\x_t$ can take over time due to certain fairness considerations. This notion of constraining trajectories is novel in stochastic convex optimization, and an important conceptual contribution of our work is to exemplify the resulting algorithmic challenges in a simple setting where the trajectory constraints are motivated by the fairness notion of \sge{envy}-freeness, as explained next.

\paragraph{\bf \sge{Envy}-Free Trajectory Constraints:} We propose a natural fairness constraint, called {\it \sge{envy}-freeness at the time of decision} (EFTD), which requires that a group, upon receiving a decision, should not {\it \sge{envy}} a decision received by any group in the present as well as the past, at least up to an allowable slack. Envy-freeness is a commonly studied notion of fairness in the literature on social choice and fair division of resources, which requires that an entity should not \sge{envy} an allocation received by someone else, i.e., prefer someone else's allocation over their own allocation \cite{foley1966resource, stromquist1980cut, moulin_2016}. This notion has also been applied in the context of fair machine learning \cite{balcan2019envy}.

In our setting, since higher decisions are perceived to be better by any group, each group will \sge{envy} every group that has received a higher decision at any point in the past. 
Instead of eliminating such \sge{envy}, we allow for certain limits on the \sge{envy} between any two groups by defining an allowable ``slack'' on ordered pairs of groups: $s(\cdot,\cdot):\mathcal{I}\times\mathcal{I}\rightarrow \mathbb{R}^{\geq 0}$. We assume that $s(i,i)=0$ for all $i\in\mathcal{I}$.

\begin{definition}
A sequence of decisions made over time is said to satisfy {\it \sge{envy} freeness at the time of the decision} (EFTD) if
\vspace{-0.1cm}
$\x_t(i) \geq \x_{t'}(j) - s(i,j) \textrm{ for all }i,\,j\in\cK,\textrm{ and }1\leq t'\leq t \leq T.$
\end{definition}


Note that the slack need not be symmetric, i.e., $s(i,j)$ may not necessarily equal $s(j,i)$. 
We note that satisfying EFTD is equivalent to satisfying dimension-wise monotonicity constraints while ensuring that the decisions lie in the \sge{envy}-freeness (EF) polyhedron within each period.

\begin{claim}
The EFTD constraints are equivalent to the following constraints
\begin{align}
\text{(i) } \mathbf{x}_t(i) &\geq \mathbf{x}_{t^\prime}(i) \text{ for all } t \geq t^\prime, i \in [N], \text{ (monotonicity) }\label{eq:mono}\\
\text{(ii) }\mathbf{x}_t(i) &\geq \mathbf{x}_{t}(j) + s(i, j), \text{ for all } t, i \in [N], j \in [N] \text{ (\sge{envy}-freeness (EF) within each period).}\label{eq:envy2}
\end{align}
\end{claim}
For the sake of non-degeneracy, we assume that the EF polyhedron has positive volume. Otherwise, one can always combine groups until this condition is satisfied. E.g., if there are three groups, and $s(1,2) = s(2,1) = 0$, then the EF polyhedron has zero volume and $\x(1) = \x(2)$ is enforced. In this case, one can simply combine groups 1 and 2 into a single group to avoid this redundancy.

\paragraph{\bf Objective:} We assume that at each time period $t$, the algorithm produces a decision $\bx_t \in \mathcal{X}^N$ and observes {\it bandit} feedback, i.e., the function values for each $i$ at the chosen decision $\x_{t}(i)$, potentially corrupted with noise. In particular, we assume that the feedback observed is a random variable $Y_{i,t} = f_i(\x_{t}(i)) + \varepsilon_{i,t}$, where $(\varepsilon_{i,t})_{1\leq t\leq T}$ for each $i\in\cK$ is a sequence of random variables representing the noise in the feedback. We will consider two cases: (a) the case of {\it noiseless feedback}, in which $\varepsilon_{i,t} = 0$ for all $i$ and $t$, and (b) the case of {\it noisy feedback}, in which we assume that the noise sequence $\{\varepsilon_{i,t}\}$ is sub-Gaussian, an assumption commonly made in the literature.

Our objective is to design an algorithm that minimizes regret, i.e., 
\begin{align}
\sum_{t=1}^T \sum_{i\in\cK} \mathbb{E}(f_i(\x_{t}(i))) - T\min_{\bx \in \text{EF}} \sum_{i\in \cK} f_i(\x(i)),
\end{align}
in the worst-case over all functions in class $\mathcal{F}$, while constraining the trajectory of $\x_t$ to satisfy EFTD. This results in a trajectory-constrained stochastic optimization problem, in which each $\mathbf{x}_t$ is constrained to be coordinate-wise monotone with respect to all $\mathbf{x}_{t^\prime}$ for $t^\prime \leq t$ (i.e., \eqref{eq:mono}), while being feasible in $\text{EF} = \{\x(i) \geq \x(j) + s(i,j), \forall i, j \in [N]\}$ at each time period (i.e., \eqref{eq:envy2}). 

Ensuring that decisions lie within the EF polyhedron at each time falls well within the scope of standard stochastic convex optimization approaches. If it were only this constraint, then several algorithms exist (e.g., see \cite{agarwal2011stochastic}) that attain the regret guarantee of $\mathcal{O}(\sqrt{T})$ assuming that the cost functions are convex, which is known to be the best attainable \cite{hazan2016introduction}. Thus, the challenging feature in our setting is the requirement of ensuring monotonicity of decisions over time for each dimension (i.e., \eqref{eq:mono}).

We will refer to algorithms that satisfy these constraints as {\it EFTD algorithms}. None of the existing regret-optimal algorithms for online convex optimization satisfy such monotonicity, as we extensively discuss in Section~\ref{sec:related}. Although this work considers the EF polyhedron arising from \sge{envy}-free constraints, the question about achievable regret for general trajectory-constrained stochastic optimization remains open (see Section \ref{sec:discussion}).

\subsection{Overview of results} 
 Throughout the paper, unless otherwise noted, we assume that $\mathcal{F}$ is the class of smooth, convex functions that satisfy the Polyak-Lojasiewicz (PL) inequality; this condition is similar to but strictly weaker than strong convexity, and is commonly made in the literature \cite{karimi2016linear}.

\begin{assumption}\label{asp:st-sm}
\js{Let $\alpha > 0$ and $\beta > 0$. We assume that $\mathcal{F}$ is the class of all \js{separable} convex functions $f(\x) = f_1(\x(1)) + \cdots + f_N(\x(N)) : \mathcal{X}^N \to \mathbb{R}$ such that each component $f_i$ is}
\begin{enumerate}
    \item \emph{$\beta$-smooth}, i.e., $f_i(y) \leq f_i(x) + \nabla f_i(x) (y-x) + \frac{\beta}{2} (y-x)^2$ for all $x,y \in  \mathcal{X}$, and,
    \item \emph{$\alpha$-{PL}, i.e., $f_i(y) - f_i(x^*) \leq \frac{1}{2\alpha} \nabla f_i(y)^2$} 
    for all $y \in \mathcal{X}$, where $x^*$ is the left-most unconstrained optimum of $f_i$.
\end{enumerate}
\end{assumption}

Our main results are as follows, which are summarized in Table \ref{tab:reorganized_label}.

\paragraph{\it Noiseless Feedback.} First, we consider the case of noiseless feedback. In this case, we first show how to design a monotone algorithm for the single-dimensional setting ($N=1$
) that achieves $\mathcal{O}(1)$ regret, which is the best achievable without the monotonicity constraint on decisions.  This algorithm performs gradient descent using gradients estimated from monotonic two-point function evaluations. The key challenge is to ensure that the decisions do not excessively overshoot the optimum. We eliminate overshooting by (a) tailoring the step size to the smoothness parameter and (b) making the descent jump from a lagged point.

 We then use the single dimensional algorithm as a building block to design a coordinate descent algorithm that yields a $\widetilde{\mathcal{O}}(1)$ regret for the $N>1$ case. This algorithm critically embeds the ability to iteratively detect tight inequalities at constrained optimality and move to strict faces of the EF polyhedron while monotonically approaching optimality. In Appendix~\ref{sec:relaxing-PL}, we show that for the case of $N=1$ and noiseless feedback, we can relax the PL condition and achieve a $\widetilde{\mathcal{O}}(1)$ regret for the class of smooth and convex functions.

\paragraph{\it Noisy Feedback.} Next, we consider the case of noisy feedback. In this case, the algorithm only observes the function value at any decision point corrupted by zero-mean sub-Gaussian noise. We make several algorithmic innovations in this case to ultimately achieve a $\tilde{O}(\sqrt{T})$ regret bound, which is the best achievable (up to logarithmic factors) without the EFTD constraint for this setting \cite{cope2009regret}. Similar to the noiseless setting, we first consider monotonic algorithm design for the 1-dimensional case, and then the 2-dimensional case; these algorithms are then used as building blocks to design an algorithm for the general $N$ dimensional case. While the high-level dynamics mirror the noiseless setting, the presence of noise makes gradient estimation challenging, especially as one approaches the optimum. We introduce several novel and intricate design elements to ensure that the sample complexity of gradient estimation does not increase regret while tackling the challenge of ensuring the monotonicity of the decision iterates. Our analysis illustrates design trade-offs that arise in satisfying monotonicity while ensuring low impact on the regret.

\begin{table}[t]
\centering
\footnotesize 
\caption{Optimal worst-case regret bounds as $T\rightarrow \infty$ for monotonic and non-monotonic bandit optimization of a separable $N$-dimensional function over the \sge{envy} freeness polyhedron under multiple settings. All bounds are tight, up to polylogarithmic factors.}
\label{tab:reorganized_label}
\begin{tabular}{|l||l|l|}
\hline
& \textbf{Without monotonicity constraint} & \textbf{With monotonicity constraint} \\ \hline \hline

\multicolumn{3}{|l|}{\rule{0pt}{3ex}\textbf{Noiseless Settings}} \\ \hline \rule{0pt}{10pt}
{Smooth, convex, PL inequality} & ${\mathcal{O}}(1)$ \cite{bubeck2015convex}  & ${\mathcal{O}}(1)$ ($N=1$) and ${\widetilde{\mathcal{O}}}(1)$ ($N>1$) {\bf[This work]} \\ \hline \rule{0pt}{10pt}
{Smooth and convex} & $\widetilde{\mathcal{O}}(1)$ \cite{bubeck2015convex}  & $\widetilde{\mathcal{O}}(1)$ ($N=1$) {\bf[This work]}\\
&&$N>1$: Open\\
\hline

\multicolumn{3}{|l|}{\rule{0pt}{3ex}\textbf{Noisy Settings}} \\ \hline \rule{0pt}{15pt}
{Unimodal, Lipschitz} & $\widetilde{\mathcal{O}}(T^{2/3})$ ($N=1$) \cite{kleinberg2004nearly} & \begin{tabular}{@{}l} $\widetilde{\mathcal{O}}(T^{3/4})$ ($N=1$) \\ \cite{jiaetal,chen2021multi} \end{tabular} \\ \hline \rule{0pt}{15pt}
{Smooth, convex, PL inequality} & \begin{tabular}{@{}l} $\widetilde{\mathcal{O}}(T^{1/2})$, \\ \cite{kiefer1952stochastic,cope2009regret} \end{tabular} & $\widetilde{\mathcal{O}}(T^{1/2})$ ($N$ general) {\bf[This work]} \\ \hline \rule{0pt}{10pt}
{Smooth and convex} & $\widetilde{\mathcal{O}}(T^{1/2})$ \cite{agarwal2011stochastic} &  Open  \\ \hline

\end{tabular}
\end{table}

\paragraph{\it Organization.} The rest of the paper is organized as follows. In Section~\ref{sec:eftd}, we discuss the EFTD constraints and give examples of practical applications that naturally fall into EFTD constrained stochastic optimization. Next, we present a discussion of related literature in Section~\ref{sec:related}. \js{In the following two sections (Sections~\ref{sec:noiseless}-\ref{sec:noisy}), we present our main algorithmic contributions, culminating in our algorithm for the case of noisy feedback and arbitrary dimensions. We discuss simpler settings first for the purpose of presenting algorithmic ideas in a modular fashion. We begin in Section~\ref{sec:noiseless} by presenting an algorithm for noiseless feedback in a single dimension, which allows us to isolate the ideas (e.g., the use of lagged points) necessary for handling monotonicity constraints. We then consider the multi-dimensional noiseless setting, which allows us to isolate the ideas (e.g., the coordinate descent dynamic) necessary for handling multiple dimensions. Finally, in Section~\ref{sec:noisy}, we discuss noisy feedback and focus our discussion on the ideas (e.g., the use of adaptive lags) required to handle noise.} 
In Section~\ref{sec:numerical-experiments}, we present numerical experiments validating our approaches using a retail pricing dataset. We finally discuss related open problems and extensions in Section~\ref{sec:discussion}, in which we highlight a rich space of research opportunities at the interface of temporal fairness and online optimization. The detailed proofs of all our results can be found in the Appendix.

\section{Applications of EFTD}
\label{sec:eftd}

EFTD provides a basic safeguard against discrimination in high-impact societal decision-making settings. It is a natural temporal notion of fairness that goes beyond ensuring the fairness of decisions across groups within each time period, which is often insufficient to meaningfully capture fairness desiderata in dynamic environments. Imagine a pricing algorithm that, while satisfying \sge{envy} freeness in each pricing interval, offers a drastically worse price to a group than what was offered to the same group only some time ago. Clearly, a compelling fairness notion for dynamic settings must address such violations. EFTD squarely addresses this concern by eliminating comparative fairness concerns relative to all past decisions. 

At the same time, it is important to note that under EFTD, while \sge{envy}-freeness is guaranteed for the decision made at time $t$ relative to all decisions made {\it before} this time, this is not guaranteed relative to decisions made {\it after} this time. In particular, given current decisions, it may turn out that a group may \sge{envy} the decision that will be received by some group in the future. EFTD is thus weaker than the more stringent notion of \sge{envy}-freeness across all time, where a decision must be deemed fair not only relative to past decisions but also the future decisions. Such a strong requirement eliminates any possibility of experimenting with decisions, which may be necessary to learn the optimal fair decisions.  
In contrast, because of the directional slack in the constraints, EFTD generally allows room for (constrained) experimentation and learning. This observation was first made in a different online learning context in \cite{gupta2019individual} and is applicable in the context of our setting as well.

We now discuss example applications of EFTD in practical problems of interest and compare it to prevailing fairness notions for these applications in the literature.

\begin{enumerate} 

\item {\bf Fair pricing.} Consider a dynamic product pricing scenario with $N$ customer segments. Let $\x_t(i)$ be the price offered to segment $i$ at time $t$. To ensure comparative fairness from the perspective of segment $i$, we may require certain upper bounds on the price offered to $i$ depending on the prices offered to the other segments in the past. For instance, we may require that $\x_t(i) \leq \x_{t'}(i') + s(i,i')$, for $t'\leq t$, where $s(i,i')\in \mathbb{R}$ represents a permissible additive slack that depends on the two segments in comparison. For example, if the two segments are youth (1) and adults (2), we may require that $\x(1)\leq \x(2)$, i.e., the price offered to the youth segment must be at most that offered to the adults. Similarly, we may require that  $\x(2)\leq \x(1) +5$, i.e., while the adults can be priced higher than the youth, the difference cannot be larger than \$$5$. Such a notion of price fairness has been recently proposed in \cite{cohen2022price} for the static complete information setting where the revenue functions are known, under the assumption that the slacks are non-negative and symmetric (so that the constraints amount to requiring that $|\x(i) - \x(i')|\leq s(i,i')= s(i',i)$ for all segments $i,\,i'$). 

In our setting, the natural extension to EFTD results in an online stochastic optimization to maximize unknown concave revenue function with the following trajectory constraints: 
\begin{align}
\x_t(i) - \x_t(i') &\leq s(i,i'), \quad \forall i, i^\prime \in [N],\, t\in[T],\\ 
\x_t(i) &\leq \x_{t^\prime}(i), \quad ~\forall i \in [N], \forall t^\prime \leq t \leq T. 
\end{align}

Note that the sense of conduciveness in this case is that lower prices are preferred, and a concave revenue function is maximized. This is easily modeled in our setting. 

\item {\bf Fair pay.} Consider a firm determining the pay levels across various roles depending on the experience level of a candidate. In this case, we can assume $\x_t(i)$ is the annual compensation offered to segment $i$ in year $t$, where a segment corresponds to a particular role and an experience level. It is natural to require certain comparative \sge{envy}-freeness constraints across these segments. For example, it may be necessary to ensure that the pay for a role at a particular experience level is never lower than what has been offered to a lower experience-level candidate for the same role in the past. It may also be necessary to ensure that the pay of a more advanced role must be higher than that offered for a less advanced role in the past. There could be more such constraints that can be modeled by asymmetric slacks under EFTD. In many scenarios, the employer may not know the optimal compensation levels that balance requisite skills against remuneration costs. EFTD constraints, similar to the first example, would allow the employer to fairly explore and learn the optimal pay levels in such scenarios. 

\item {\bf Fair assortment planning.} \cite{chen2022fair} recently considered the scenario where a platform needs to make assortment planning decisions over $N$ products/sellers to show to a customer. Each product $i$ has some weight $\mathbf{w}(i) \geq 0$ that measures how popular the product is to customers; in particular, upon offering the customers an assortment $\mathcal{S}$ of products, the customers purchase product $i \in \mathcal{S}$ according to the Multinomial Logit Choice (MNL) model, with probability $\mathbf{w}(i)/(1+\mathbf{w}(\mathcal{S}))$ where $\mathbf{w}(\mathcal{S}) = \sum_{j\in\mathcal{S}} \mathbf{w}(j)$. Given a probabilistic assortment choice of the platform, suppose that $\mathbf{y}(i)$ is the probability that $i$ is included in the assortment, referred to as the {\it visibility} offered to product $i$. \cite{chen2022fair} propose a notion of fairness that requires that for every product $i$ and for some $\delta>0$, 
$$\mathbf{y}(i) \geq \frac{\mathbf{w}({j})}{\mathbf{w}({i})} \mathbf{y}(j) - \delta, \textrm{ for all }i, j \in [N].$$
That is, product $i$ must get at least as much visibility (with a $\delta$ slack) as product $j$ after adjusting for their relative differences in popularity. \cite{chen2022fair} assume a complete information setting where the platform's objective function, which is the expected revenue from an assortment, is known. However, one can imagine an incomplete information setting where the objective differs from expected revenue (e.g., customer welfare) and is a priori unknown. In this case, the following temporal extension corresponding to the EFTD constraint can allow the platform to optimize the assortment over time in a fair manner.  
\begin{align*}
\mathbf{w}(i)\mathbf{y}_t(i) &\geq \mathbf{w}({j}) \mathbf{y}_t(j) - \delta\mathbf{w}(i), \textrm{ for all }i, j \in [N],\, t\in[T],\\
\mathbf{w}(i)\mathbf{y}_t(i) &\geq \mathbf{w}(i)\mathbf{y}_{t^\prime}(i), \quad \forall t^\prime \leq t \leq T.
\end{align*}

This constraint can be seen as EFTD constraint by considering $\x(i) =  \mathbf{y}(i)\mathbf{w}({i})$ to be the decision corresponding to each group, and the slack $s(i,j)$ is defined to be $\delta \mathbf{w}(i)$ for all $j\neq i$. That is, 
\begin{align}
\x_t(i) - \x_t(j) &\geq \delta \mathbf{w}(i), \quad \forall i, j \in [N],\, t\in[T],\\ 
\x_t(i) &\geq \x_{t^\prime}(i), \quad ~\forall i\in [N], ~t^\prime \leq t \leq T. 
\end{align}

\end{enumerate}

\section{Related Work}
\label{sec:related}
{\it Stochastic convex optimization.} This work introduces the concept of trajectory-constrained stochastic convex optimization under bandit feedback motivated through the lens of fairness. In general, existing algorithms from the convex optimization toolkit cannot be easily modified to satisfy monotonicity of iterates to satisfy EFTD. In the noiseless bandit feedback setting, Kiefer gave an $\mathcal{O}(1)$-regret algorithm, now well known as \textsc{Golden-Section Search}, for minimizing a one-dimensional convex function \cite{kiefer1953sequential}. This algorithm iteratively uses three-point function evaluations to ``zoom in'' to the optimum, by eliminating a point and sampling a new point in each round. Its mechanics render it infeasible to implement it in a fashion that respects the monotonicity of decisions. For higher dimensions, a $\mathcal{O}(1)$-regret algorithm has been designed by \cite{nesterov2017random}. This algorithm is based on gradient-descent using a one-point gradient estimate constructed by sampling uniformly in a ball around the current point. This key idea recurringly appears in several works on convex optimization with bandit feedback (e.g., \cite{spall1992multivariate}, \cite{flaxman2005online}, \cite{hazan2014bandit}). However, due to the randomness in the direction chosen to estimate the gradient, such an approach does not satisfy monotonicity.

For the case of noisy bandit feedback, \cite{cope2009regret} has shown a lower bound of $\Omega(\sqrt{T})$ in the single-dimensional setting that holds for smooth and strongly convex functions and showed that an appropriately tuned version of the well-known Kiefer-Wolfowitz \cite{kiefer1952stochastic} stochastic approximation algorithm achieves this rate. This algorithm uses two-point function evaluations to construct gradient estimates, which are then utilized to perform gradient-descent. Our near-optimal algorithm in the single dimensional case is the most related to this approach, where we tackle the significant additional challenge that the two-point evaluations need to consistently satisfy monotonicity over time while attaining the same regret bounds (up to logarithmic factors). For the case of convex functions, Agrawal et al. designed an algorithm that achieves the $\tilde{\mathcal{O}}(\sqrt{T})$  bound under bandit feedback \cite{agarwal2011stochastic}. In the one-dimensional case, their approach is the most related to the golden-section search procedure of \cite{kiefer1953sequential}, and as such, is infeasible to implement in a monotonic fashion. 

Though not motivated by fairness concerns, two recent parallel works have considered the problem of ensuring monotonicity of decisions in stochastic optimization under bandit feedback (\cite{jiaetal} and \cite{chen2021multi}). Assuming that the functions are one-dimensional and known to be Lipschitz and unimodal, the authors show that the optimal achievable regret is $\tilde{\Theta}(T^{3/4})$, which is higher than the $\tilde{\Theta}(T^{2/3})$ regret achievable without the monotonicity requirement. In contrast, Theorem~\ref{prop:lgd-noisy-bandit-dynamic-lags} implies that under the assumption that the cost functions are smooth, convex, and they satisfy the PL-inequality, the unconstrained optimal regret bound of $\tilde{\mathcal{O}}(\sqrt{T})$ is attainable while ensuring monotonicity of decisions (up to logarithmic terms). We also note that, because the settings considered by these other works are more pessimistic in their view of the possible cost functions, algorithm-design for optimizing the worst-case turns out to be simpler. In particular, both \cite{jiaetal} and \cite{chen2021multi} show that we cannot do better than sequentially traversing the decision space in a fixed direction and using a fixed step size until the utility keeps increasing (as determined by a sequence of two-point hypothesis tests). In contrast, the problem of leveraging gradient information while maintaining monotonicity, while also ensuring negligible impact on regret, results in several novel and non-trivial aspects of our algorithm design.

\paragraph{\it Fairness in static decision-making.} Notions of fairness in static decision-making typically ensure that various stakeholders are treated equitably \cite{sen20136}. Two broad varieties of notions have appeared in this regard.  First is the popular notion of {\it individual fairness} \cite{Dwork2012}, which requires that similar individuals must receive similar treatments. Formally, if $\x(i)$ is the decision received by individual $i$, then individual fairness requires that $|\x(i)-\x(j)|\leq Kd(i,j)$ for any two individuals $i$ and $j$, where $K>0$ is some constant, and $d(i,j)$ is a similarity metric that captures some publicly acceptable notion of ``distance" between two individuals that dictates the allowable disparity between their decisions. The second notion is {\it group fairness}, also known as {\it statistical parity} \cite{zafar2017fairness,calders2009building}, under which one demands parity in certain statistics of the decisions across multiple groups of people, e.g., requiring that two groups of populations in the same role have similar average salaries in an organization.  
Such parity notions have been proposed in the context of multi-segment pricing recently by \cite{cohen2022price} and \cite{kallus2021fairness}. For example, the notion of price fairness considered by \cite{cohen2022price} imposes bounds on the price disparity seen across segments, i.e., they require that $|\mathbf{p}(i)-\mathbf{p}(j)|\leq C_{ij}$ for any two customer segments $i$ and $j$ where $C_{ij}>0$ is some constant. The common aspect of these two notions is that they impose bounds on the degree of {\it \sge{envy}} felt by entities -- either individuals or groups -- when comparing their decision to others. The notion of \sge{envy}-freeness we consider is a generalization of these notions of fairness in that we allow for asymmetric bounds on the disparity in \sge{envy}.

\paragraph{\it Temporal notions of fairness in dynamic decision-making.} Broadly, four different types of temporal notions of fairness have appeared in the literature on online decision-making. The first type of constraint is a {\it time-independent} fairness constraint in which a static fairness constraint must be satisfied independently in each period. For example, \cite{chen2021fairness} and \cite{cohen2021dynamic} consider a multi-segment dynamic pricing problem where the price fairness constraint of \cite{cohen2022price} must be satisfied in each time period. 

In contrast, the second type of fairness constraint is what we refer to as a {\it time-dependent decision-adaptive fairness constraint}. FTD is an example of this type of temporal constraint. Such constraint changes depending on the decisions taken in the past, but is agnostic to the outcomes of the decisions. Similar constraints have appeared in recent and parallel literature. For example, \cite{cohen2021dynamic} requires that the price offered to a particular segment must be close to the prices offered to the same segment over the past $k$ time periods (in addition to satisfying the price fairness constraint across segments every time period). Such a sliding-window fairness constraint was also considered by \cite{heidari2018preventing} in the context of adaptive supervised learning. 

In a similar vein, and the most related to our paper, \cite{gupta2019individual} define the notion of {\it individual fairness in hindsight} in the context of contextual bandit problems. This is a temporal version of the static notion of individual fairness that guarantees individual fairness at the time of decision based on some underlying similarity metric. Our objective of EFTD is a significant generalization of this notion that goes beyond individual fairness objectives for static decision-making (e.g., unlike individual fairness, we can model asymmetric allowable disparities across groups).  Moreover, the technical analysis of \cite{gupta2019individual} is restricted to an abstract contextual bandit setup where the decision-maker tries to distinguish between a finite set of utility models. Our framework, on the other hand, considers a general stochastic convex optimization formulation that is applicable in a broader variety of settings. 

The third type of temporal fairness constraint is what we refer to as a \emph{time-dependent outcome-adaptive} fairness constraint, where the constraint can change as the outcomes from the decisions change. For example, \cite{joseph2016fairness} has introduced a fairness constraint in a contextual bandit setting requiring that preferential treatment be given to an arriving entity only if it is known with high enough certainty that such preferential treatment is justified by current estimates of the mean rewards resulting from the treatment. Thus, as entities are administered decisions over time, certainty in their reward distributions increases, and they can eventually be treated differently. A Bayesian version of this idea was proposed by \cite{liu2017calibrated} in the non-contextual bandit setting, in which decisions are constrained to be fair with respect to the posterior reward distributions of the arms at any time. In this way, as the posterior distributions converge to the true distributions, violations of the desired fairness constraint should diminish. In a similar vein, \cite{salem2019closing} consider a problem of ensuring group fairness in hiring, where uncertainty is modeled using a partial order and decisions are constrained to be consistent with the partial order at any time. 

The fourth type of temporal fairness constraint asks for fairness of certain decision statistics, either ex ante or ex post, across arriving entities over time. For example, \cite{manshadi2021fair} considers a sequential resource allocation setting where entities arrive over time with certain demands and the goal is to maximize the minimum fill rate. \cite{ma2020group} similarly considers the problem of maximizing the minimum service rate across different groups in an online bipartite matching setting. The main difference from the other three types of constraints is that one doesn't necessarily seek to guarantee comparative fairness for an entity arriving at a point in time, but rather one guarantees fair treatment of groups of entities across time. A general framework for satisfying such time-average fairness constraints in the context of kidney transplantation has been proposed by \cite{bertsimas2013fairness}.

\section{Regret-optimal EFTD algorithms without noise}
\label{sec:noiseless}

In this section, we discuss algorithm design for the problem introduced in Section~\ref{sec:stoc-ftd} in the case of noiseless feedback, i.e., under the assumption that $\varepsilon_{i,t} = 0$ for all $i \in [N]$ and $t \in [T]$. 

\js{To build some intuition, imagine that we had noiseless first-order information. Then, by applying gradient descent with jumps of $-\frac{1}{\beta} \nabla f(\x_t)$, one would converge monotonically toward the optimum. However, in the case of noisy bandit setting, one must sample multiple points to obtain a gradient estimate, and it is non-trivial to do so in a monotonic fashion.  In this section, we handle the case of the bandit feedback without noise, and then discuss how to handle problems stemming from noisy feedback in Section \ref{sec:noisy}.}


Recall from Assumption~\ref{asp:st-sm} that $\mathcal{F}$ is the class of convex and $\beta$-smooth functions which additionally satisfy the Polyak-{\L}ojasiewicz (PL) inequality with parameter $\alpha$. For expositional clarity, we first discuss the single-group setting (Section~\ref{subsec:noiseless-single-group}), and then the multi-group setting (Section~\ref{subsec:noiseless-multi-group}).

\subsection{Algorithm design for \texorpdfstring{$N=1$}{N=1} group}
\label{subsec:noiseless-single-group}

We first consider algorithm design in the single-group noiseless bandit setting: given a decision $x \in \cX$, the algorithm observes $f(x)$. In this setting, the EFTD constraints reduce to monotonicity in decisions; that is, decisions are weakly increasing over time.

\begin{algorithm}[t]
\footnotesize
\DontPrintSemicolon
\SetKwInOut{Input}{input}
\Input{smoothness parameter $\beta$, $T$, initial lagged point $x_1'$, lag size $\delta=T^{-1/2}$}
Set $x_1 \leftarrow x_1' + \delta$, and observe $f(x_1')$ and $f(x_1)$ \;
\For{$t=1,\ldots,T/2$}{
    Let $\tilde \nabla_t \leftarrow \frac{f(x_t)-f(x_{t}')}{x_t-x_{t}'}$ \;
    \uIf{$ - \frac{1}{\beta} \tilde
    \nabla_t \geq (1+\gamma)\delta$}{
        Sample $f(\cdot)$ at $x_{t+1}^\prime = x_{t}' - \frac{1}{\beta} \tilde \nabla_t-\delta$ (lagged iterate)\; 
        Sample $f(\cdot)$ at $x_{t+1} = x_{t+1}^\prime+\delta$ (non-lagged iterate)
    }
    \Else{
        Exit from loop and stabilize at $x_t$ \;
    }
}
\caption{Lagged Gradient Descent ({\sc Lgd}) (noiseless bandit) \label{alg:Lagged-2pt-Gradient-Descent-Backtracking}}
\end{algorithm}

We present a monotonic procedure called {\sc Lagged Gradient Descent} (\textsc{Lgd}) (Algorithm \ref{alg:Lagged-2pt-Gradient-Descent-Backtracking}) for this setting, which is a variation on classical gradient descent. At each round, we use two queries (one at $x_t$ and one at a ``lagged'' point $x_t' =x_t-\delta$) to estimate the gradient. With some abuse of notation, we will let $x_1, x_2, \ldots$ denote non-lagged iterates in this process. The algorithm therefore terminates when $T$ total samples have been taken, or at decision $x_{\lfloor T/2 \rfloor}$.

The lag size $\delta$ is a parameter which will be tuned for minimizing the regret of the overall scheme. Since we need to ensure monotonicity of iterates, we need to sample first at the lagged point $x_t'$ and next at $x_t$ to get an estimate of the gradient. We then move by an amount proportional to the estimated gradient.

While increasing decisions, even in this non-noisy case, we need to be careful about not excessively overshooting the optimum point, which could result in high regret due to the monotonicity constraint that disallows backtracking. To avoid overshooting, we move proportional to the gradient {\it from the lagged point $x_t'$ instead of from $x_t$}. Since the estimated gradient $\tilde{\nabla}_t$ in {\sc Lagged Gradient Descent} is less steep than the true gradient at $x_t'$, the smoothness of $f$ allows us to ensure that we never overshoot. However, since we jump from $x_t'$ instead of $x_t$, a small jump (of a magnitude smaller than $x_t-x'_t$) would violate monotonicity. To avoid this, we jump forward only if the gradient is steep enough; in particular, if the magnitude of the estimated gradient is at least $\beta (1+\gamma)\delta$, for some $\gamma>1$ (see Theorem \ref{cor:lgd-one-stopping-criteria-noiseless}), we stay at the current decision for the remaining time.

Lemma \ref{prop:lagged-gradient-descent-backtrack-convergence-pl} below shows that the decisions resulting from {\sc Lagged Gradient Descent (Lgd)} are monotonic, avoid overshooting, and converge at an exponential rate. 

\begin{lemma} \label{prop:lagged-gradient-descent-backtrack-convergence-pl}
Let $f : \mathcal{X} \to \mathbf{R}$ be a function satisfying Assumption~\ref{asp:st-sm}. 
Let $x_1,\ldots,x_{T/2}$ be the non-lagged points generated by {\upshape {\sc Lagged Gradient Descent}} (Algorithm \ref{alg:Lagged-2pt-Gradient-Descent-Backtracking}), and assume that $x_1 \leq x^* = \arg\,\min_{x \in \mathbb{R}} f(x)$. Then, for $\gamma > 1$, the following hold:
\begin{enumerate}
    \item Decisions increase monotonically toward the optimum: $x_1 \leq x_2 \leq x_3 \leq \cdots \leq x_{T/2} \leq x^*$;
    \item Convergence rate up to halting is: $h_t = f(x_t) - f(x^*) \leq h_1e^{-c(t-1)}$
    , where $c = \frac{1}{2\beta} - \frac{1}{(1+\gamma)\beta}$. 
\end{enumerate}
\end{lemma}

To prove the first statement of Lemma~\ref{prop:lagged-gradient-descent-backtrack-convergence}, we use smoothness and the fact that we jump using a shallow gradient to show that the iterates do not overshoot. To show the convergence rate, we begin by exploiting smoothness to show that $h_{t+1}-h_t \leq -c \Vert \widetilde{\nabla}_t \Vert^2$. We then use the PL condition to relate $\Vert\widetilde{\nabla}_t \Vert^2$ to $h_t$, which ultimately results in a contraction in $h_t$ (see Appendix \ref{app:single-group}). 

With this convergence rate established, we can calculate a regret bound for {\sc Lgd}. We can break the regret of this procedure into three categories: regret during exploration, regret due to stopping (i.e., regret incurred after the ``for loop'' has ended), and regret due to potential overshooting (which is 0 by Lemma \ref{prop:lagged-gradient-descent-backtrack-convergence-pl}). We balance these three to obtain an $\mathcal{O}(1)$ regret bound. 

\begin{theorem} 
\label{cor:lgd-one-stopping-criteria-noiseless-pl}
Assume that $x^* = \arg\,\min_{x \in \mathbf{R}} f(x)$
, the initial lagged point satisfies $x_1' < x^*$, and fix $\delta = T^{-1/2}$ and $\gamma = 1 + \frac{1}{\log T}$. Then {\upshape {\sc Lagged Gradient Descent}} (Alg. \ref{alg:Lagged-2pt-Gradient-Descent-Backtracking}) is a $\mathcal{O}(1)$-regret EFTD algorithm for optimizing a convex and $\beta$-smooth function in the noiseless bandit setting.
\end{theorem}

Theorem \ref{cor:lgd-one-stopping-criteria-noiseless-pl} shows that imposing monotonicity essentially has no effect on the optimal asymptotic regret of the noiseless setting: a non-monotonic constant-regret procedure is already known \cite{kiefer1953sequential}, and {\sc Lgd} is a {\it monotonic} $\mathcal{O}(1)$-regret procedure. Moreover, if the objective function is only known to be smooth and convex, then one can tweak the analysis to show that {\sc{Lgd}} achieves $\tilde{\mathcal{O}}(1)$-regret by setting $\delta = T^{-1}$ (see Theorem~\ref{cor:lgd-one-stopping-criteria-noiseless} in Appendix~\ref{sec:relaxing-PL}).


\subsection{Algorithm design for \texorpdfstring{$N\geq 2$}{N>1} groups}
\label{subsec:noiseless-multi-group}

In this section, we detail algorithm design for $N$ groups assuming noiseless bandit feedback. Recall that the objective function $f$ is \js{separable, convex, smooth, and satisfies the PL inequality (Assumption~\ref{asp:st-sm}).} 
We will extend the 1-dimensional algorithm to multi-dimensions. Ensuring EFTD in the presence of multiple groups imposes two types of constraints across decisions over time, as discussed before:  

\begin{enumerate}
\item {Coordinate-wise monotonicity,} i.e.  
$\x_t(i) \geq \x_{t'}(i)$ for $1\leq t'\leq t \leq T$ and for $i \in [N]$, 
\item {Cross-coordinate EF constraints,} i.e., 
$\x_t(i) \geq \x_{t}(j) - s(i,j)$ for all $1\leq t \leq T$ and $i,j \in [N]$.
\end{enumerate}

\paragraph{\bf Algorithmic approach.} We will start monotonically optimizing each group in isolation, switching to the next group whenever we are close enough to a cross-coordinate constraint, or a group-specific optimum is attained. In other words, we employ a form of coordinate descent across the group dimensions while ensuring we always satisfy the EF constraints. This approach, however, does not converge to the constrained optimum, just as stated. For example, consider the scenario in Figure~\ref{fig:combining-needed} (a). In this scenario, the first coordinate $\x_t(1)$ cannot move toward its optimum in isolation, since $\x_t(1)$ is at a tight constraint. At the same time, $\x_t(2)$ is at its optimum, so it will not move in isolation under gradient descent mechanics. Thus, monotonic coordinate descent will not converge to the joint constrained optimum $\x^*_C$.

\begin{figure}
\begin{center}
\scalebox{.8}{
\begin{tikzpicture}

\draw[white, fill=gray, fill opacity=.1] (-5.5,3) -- (-4,3) -- (-4,2.5) -- (-5.5,2.5) -- cycle;
\node at (-2.5,2.75) {Feasible region};
\draw[white, fill=gray, fill opacity=.1] (-3,-.9) -- (1,3.1) -- (2,2.1) -- (-2,-1.9) -- (-3,-1.9) -- cycle;
\draw (-3,-.9) -- (1,3.1);
\draw (-2,-1.9) -- (2,2.1);

\draw[->] (-3,-1.9) -- (2.6,-1.9) node[right]{$\x(1)$};
\draw[->] (-3,-1.9) -- (-3,1.5) node[above]{$\x(2)$};

\draw[->] (-1,-.9) -- node[fill=white]{$-\nabla f(\x_t)$} (1.6, -.9);

\tikzstyle{vertex}=[draw, circle, color = black, fill = black, text opacity = 1, inner sep = 1.4pt]

\node[vertex] (z1) at (-1,-.9) {};
\node[vertex] (z3) at (1,1.1) {};
\node[vertex] (z3) at (3,-.9) {};

\node at (-1.4,-.6) {$\x_t$};
\node at (2.6,-.6) {$\x^*$};
\node at (.6,1.4) {$\x^*_C$};

\node at (0, -2.4) {\textbf{(a)}};

\draw[dashed] (4,3) -- (4,-2);

\begin{scope}[shift={(8,0)}]
\draw[white, fill=gray, fill opacity=.1] (-3,-.9) -- (1,3.1) -- (2,2.1) -- (-2,-1.9) -- cycle;
\draw (-3,-.9) -- (1,3.1);
\draw (-2,-1.9) -- (2,2.1);

\tikzstyle{vertex}=[draw, circle, color = black, fill = black, text opacity = 1, inner sep = 1.4pt]
\node[vertex] (z3) at (1.6,1.7) {};
\node[vertex] (z3) at (2.2,1.1) {};

\node at (2.5,.8) {$\x^*$};
\node at (1.2,2) {$\x^*_C$};

\draw[ultra thick, dashed] (-2.3,-1.6) -- (-1.7,-1.6) -- (-1.7,.4) -- (.3,.4) -- (.3,1.1) -- (1,1.1) -- (1.6,1.7);

\node at (0, -2.4) {\textbf{(b)}};
\end{scope}

\end{tikzpicture}
}
\end{center}
\caption{\label{fig:combining-needed}
\textbf{(a)} Example of a scenario in which monotonic coordinate descent does not converge to the constrained optimum $\x^*_C$. Here, $\x_t(1)$ cannot be optimized further, since it is at a tight constraint, and $\x_t(2)$ cannot be optimized further, since it is already at its optimum. Yet, $\x_t$ still has not attained its joint constrained optimum of $\x^*_C$. \textbf{(b)} Trajectory of the continuous-time algorithm.}
\end{figure}

To address this issue, our idea is to \emph{combine groups} whenever they are unable to move towards their individual optima due to EF constraints (in other words, by combining groups, we jump to the appropriate face of the EF polyhedron) and then continue with coordinate descent on these clusters of groups. For illustrative purposes, we first describe our $N$-group method as a continuous-time algorithm with access to perfect gradient feedback. Our method involves two mechanisms: a coordinate descent mechanism, in which we optimize each group separately, and a cycle-then-combine mechanism, which dictates how the clusters are combined. 

\begin{enumerate}
\item {\it Coordinate descent dynamics (continuous-time).}  Starting with $\x_{0}(1) = \cdots = \x_{0}(N) = x_{\min}$, pick an arbitrary coordinate, say $i$, and increase it while keeping the other coordinates fixed until either (a) $\nabla f_i(\x_{t}(i)) = 0$, or (b) $\x_{t}(i)$ has hit some constraint. Switch to the next coordinate and repeat until no coordinates can move. Once that happens, go to Step 2. 
\item {\it Cycle-then-combine dynamics (continuous-time).} If $\nabla f_i(\x_{t}(i)) = 0$ for all $i\in [N]$, then we are done -- the EF constraint is not binding, and the unconstrained optimum is the same as the constrained optimum. Else, there is some $i,j \in [N]$ such that $\nabla f_i(\x_{t}(i)) = 0$ and $\x_{t}(j) = \x_{t}(i)  + s(i, j)$ (see Claim~\ref{claim:existence-of-low-gradient}). At that point, we can deduce that the corresponding EF constraint will be tight at the optimum, which means we can combine groups $i$ and $j$ into a cluster $\{i,j\}$ and optimize their joint objective $f_{\{i,j\}}(\x_{t}(i), \x_{t}(j)) = f_i(\x_{t}(i)) + f_j(\x_{t}(i)  + s(i, j)) \triangleq \psi_{\{i,j\}}(\x_t(i))$, where we will refer to $i$ as cluster $\{i,j\}$'s representative. Return to Step 1, replacing groups $i$ and $j$ with cluster $\{i,j\}$. 
\end{enumerate}

See Figure~\ref{fig:combining-needed} (b) for an example of the trajectory of this continuous-time algorithm for two groups. Convergence of the above continuous-time method hinges on the ability to combine clusters when they get stuck. This is argued below. 

\begin{claim}[continuous-time algorithm does not get stuck] \label{claim:existence-of-low-gradient}
Suppose the continuous-time algorithm has partitioned $[N]$ into clusters $C_1,C_2,\ldots,C_k$ at time $t$; in other words, for all $t' \geq t$, all $j \in [k]$, and all $i_1,i_2 \in C_j$, the gap in decisions of $i_1$ and $i_2$ is held constant (i.e., these decisions are ``locked" relative to each other):   
\[
\x_{t'}(i_1) - \x_{t'}(i_2) = \x_t(i_1) - \x_t(i_2).
\]
Let $f_{C_j}(\x) = \sum_{i \in C_j} f_i(\x(i)) = \psi_j(\x(i_j))$ be the objective associated to cluster $C_j$, where $i_j$ is a representative group from cluster $C_j$ (note that the decisions of all other groups from $C_j$ are fully determined as a function of $\x(i_j)$ given the relative gaps that are held constant). Now suppose that no cluster can move according to the gradient descent mechanism; that is, for every cluster $C_j$, either (1) the cluster is at its optimum, i.e., $\psi_j'(\x_t(i_j))=0$, or 
(2) $C_j$ is at a tight constraint, i.e., $\x_t(i_1) = \x_t(i_2) + s(i_1,i_2)$ for some $i_1 \in C_j$ and $i_2 \not\in C_j$. Then one of the following holds:
\begin{enumerate}
    \item[(a)] every cluster is at its optimum (i.e., $\psi_j'(\x_t(i_j))=0$ for all $j \in [k]$), or
    \item[(b)] there exists clusters $C_{j_1}$, $C_{j_2}$ such that $C_{j_2}$ is at its optimum and $C_{j_1}$ is constrained by cluster $C_{j_2}$ (i.e., $\psi'_{j_2}(\x_t(i_{j_2}))=0$ and $\x_t(i_1) = \x_t(i_2) + s(i_1,i_2)$ for some $i_1 \in C_{j_1}$ and $i_2 \not\in C_{j_2}$).
\end{enumerate}
\end{claim}

The proof of this claim has the following structure. First, a directed graph $D=(\{C_1,\ldots,C_k\}, A)$ is created, where an arc $(C_i,C_j)$ means that some group in $C_i$ is constrained by some group in $C_j$. If not all clusters are at their optima, then this graph is non-empty. Moreover, one can show that it is acyclic, by the non-degeneracy of slacks. This implies the existence of a node with positive in-degree and zero out-degree, which satisfies condition (b) in the claim. See Appendix~\ref{app:n-dim-noiseless} for the full proof.

To show that the continuous-time algorithm converges to the constrained optimum, we need to additionally show that the algorithm never overshoots the constrained optimum. To that end, observe that in the course of the continuous-time algorithm, no cluster overshoots its unconstrained optimum, since the gradient descent mechanics will cause cluster $C_j$ to stop when $\psi'_j(\cdot)$ is 0. The following crucial claim then implies that no cluster can overshoot the constrained optimum.

\begin{claim}[no overshooting the constrained optimum] \label{claim:continuous-time-overshooting}
Let $f$ be an objective function satisfying Assumption~\ref{asp:st-sm} with unconstrained optimum $\x^*$ and constrained optimum $\x^*_C$, where the feasible set is an EF polyhedron. Then for any feasible $\x$ satisfying $\x \leq \x^*$, it must be that $\x \leq \x^*_C$.
\end{claim}
Note that this claim does not hold generally for polyhedra other than EF. Claim~\ref{claim:existence-of-low-gradient} and Claim~\ref{claim:continuous-time-overshooting} imply that the continuous-time approach above, with access to perfect gradient information, converges to the constrained optimum. In our bandit setting with discrete time, however, decision steps are discrete. We thus adapt the above continuous-time procedure to a discrete-time procedure. In this discrete time procedure, we will assume that we obtain perfect gradient feedback for each decision. This assumption is costless from the perspective of the asymptotic regret guarantee since one can leverage the noiseless bandit feedback to construct arbitrarily accurate gradient estimates using arbitrarily close two-point function evaluations.

\begin{itemize}
    \item \emph{Coordinate descent dynamics.} Suppose we are optimizing Group $i$. We begin by observing the gradient $f'_i$ at the current iterate $\x(i)$ and calculating the next gradient-descent iterate $\x(i)-\frac{1}{\beta} f'_i(\x(i))$. The coordinate $\x(i)$ is then increased as much as possible in the direction of gradient-descent update while remaining feasible. Move onto the next group once one of the following happens: a gradient-descent update has been performed, the group is at a constraint, or $f'_i(\x(i))$ is close to 0.
\end{itemize}

As with the continuous-time approach described above, a cluster-combination mechanic can be introduced to avoid clusters getting stuck prior to reaching the constrained optimum.
\begin{itemize}
    \item \textit{Cycle-then-Combine procedure.} At the beginning of the algorithm, place each group in its own set, which we call a \emph{cluster}: $\{1\}, \{2\}, \ldots, \{N\}$. When we reach a point at which no groups can move in isolation, we find two groups $i,j$ for which (1) group $i$ is at its optimum, and (2) group $j$ is constrained by group $i$. We then combine the two groups into one cluster $\{i,j\}$, and optimize their joint objective $f_i(\x(i)) + f_j(\x(j))$ for the rest of the time horizon. We can combine clusters containing multiple groups each in a similar fashion. When optimizing a cluster, we treat the cluster as a single (one-dimensional) group, moving all component groups by the same amount each iteration. Each time this procedure is performed, the number of clusters decreases by one. 
\end{itemize}

\js{A high-level description of the algorithm, called \textsc{NC}$^2$-\textsc{Lgd}, can be found in Algorithm~\ref{alg:multi-segment-noiseless}. A more detailed description (Algorithm~\ref{alg:multi-segment-noiseless-long}) and} a description of all notation used in the algorithm can be found in Appendix~\ref{app:n-dim-noiseless}. As we will argue, this approach converges to the optimum and results in logarithmic regret. 

\begin{algorithm}[!t]
\footnotesize
\DontPrintSemicolon
\SetKwInOut{Input}{input}
\Input{Number of groups $N$, smoothness parameter $\beta$ of $f(\x(1),\ldots,\x(N)) = \sum_{i=1}^N f_i(\x(i))$, time horizon $T$, non-negative EF slacks $s$, \js{$x_{\min} \in \mathbb{R}$}}
\js{Maintain a set $S$ of points to be queried, starting with $x_{\min}$}\; 
Initialize the partition $\Pi = \big\{ \{1\},\ldots,\{N\} \big\}$ of groups\;
For any cluster $A \subseteq [N]$, define $\psi_A(x) = \sum_{j \in A} f_j(x + b_j)$, where $b_j$ for $j \in A$ is defined by the tight constraints of cluster $A$ (see Appendix~\ref{app:n-dim-noiseless}); 

\While{fewer than $T$ samples have been taken}{
    \If{no cluster can move}{
        \uIf{each cluster is constrained by some other cluster}{
            Combine tight clusters to get a coarser partition $\Pi$. 
        }
        \Else{
            There is a cluster $C$ that is at a tight constraint imposed by some group in cluster $D$, and $|\psi'_D(\x(\min D))| < T^{-1}$ \;
            Combine clusters $C$ and $D$ and update the partition $\Pi$. 
        }
    }
    {\bf Sampling:} Select a cluster to sample \js{and select its} representative group $i$\;
    \If{there are points to sample in the queue for cluster $i$}{
        sample point $\x(i)$ for group $i$\;
        Calculate $\psi_A'(\x(i))$ \;
               Add $\x(i) - \frac{1}{\beta} \psi_A'(\x(i))$ to queue for group $i$ \;
    }  
    \If{$|\psi'_B(\x(\min B))| < T^{-1}$ for every cluster $B \in \Pi$}{
        Exit the while loop and remain at point $\x$ for the remaining iterations
    }
}
\caption{Noiseless Cycle-then-Combine Lagged Gradient Descent ({\sc NC$^2$-Lgd})}
\label{alg:multi-segment-noiseless}
\end{algorithm}

\begin{theorem} \label{prop:multi-group-noiseless}
Suppose the objective function \js{$f$ satisfies Assumption~\ref{asp:st-sm}.} 
The multi-group algorithm assuming perfect gradient access, Algorithm ~\ref{alg:multi-segment-noiseless}, attains a regret of $\mathcal{O}(N^2 \log T)$.
\end{theorem}

\paragraph{\bf Proof sketch.} To prove the claim, we must first argue that the algorithm is correct (i.e., the steps are implementable), and then argue that the regret is constant. For correctness, we must argue that whenever the algorithm gets stuck (i.e., no cluster can move in isolation), then two clusters satisfying the following property can be found: one cluster is at its optimum, and the other is constrained by the first. This follows from Claim~\ref{claim:existence-of-low-gradient}.

To bound the regret of the algorithm, we first note that the number of incomplete gradient descent jumps (i.e., those which cannot be fully made due to the EF constraints) can be bounded by a number which depends on the EF constraints and is polylogarithmic in $T$. To see this, first note that convergence under Assumption~\ref{asp:st-sm} is exponential (see the proof of Theorem~\ref{cor:lgd-one-stopping-criteria-noiseless-pl} in Appendix~\ref{app:single-group}), so it takes $\mathcal{O}(\log T)$ updates until a cluster reaches its small-gradient threshold of $T^{-1}$. Thus, for a particular cluster $C$, the total number of iterations (over all clusters) until $C$ reaches the small-gradient threshold is $\mathcal{O}(N\log T)$. Since there are at most $N$ clusters, we can conclude that the total number of incomplete gradient descent jumps is $\mathcal{O}(N^2\log T)$. The remaining jumps therefore 
advance toward the current constrained optimum at an exponential rate (i.e., $h_t \in \mathcal{O}(e^{-ct})$), for some constant $c$.

Since the small-gradient threshold $1/T$ is positive, the algorithm may erroneously combine clusters. When this happens, the algorithm locks the two clusters together, preventing convergence to the constrained optimum. However, the new constrained optimum (that is, the constrained optimum which incorporates this new constraint) is at most $\mathcal{O}(1/T)$ away from the old constrained optimum. Thus, even if clusters are combined erroneously, the ultimate constrained optimum $x$ will have instantaneous regret on the order of $1/T^2$ (by PL). 
The algorithm therefore incurs a regret of $o(1)$ (ignoring factors of $N$) from these erroneous combinations. Since we have exponential convergence at all but logarithmically many iterations, and the regret from erroneous combinations is $o(1)$, the total regret of the algorithm is $\mathcal{O}({N^2\log T})$. \qed

One important aspect of the noiseless setting is that gradient estimation is easy: one simply requires two samples, and additionally, the gap $\delta$ between the samples can be arbitrarily small, resulting in arbitrarily accurate gradients. As we will discuss in the next section, when noise is introduced, there is a trade-off between the magnitude of $\delta$ and the number of samples required to accurately estimate the gradient. This tension ultimately results in a higher regret bound.

\section{Regret-optimal EFTD algorithms in the presence of noise}
\label{sec:noisy}
In this section, we tackle the challenges that arise in designing EFTD algorithms in the presence of noise. Recall once again from Assumption~\ref{asp:st-sm} that $\mathcal{F}$ is the class of convex and $\beta$-smooth functions, which additionally satisfy the Polyak-{\L}ojasiewicz (PL) inequality with parameter $\alpha$. Furthermore, we make the following assumption on the noise sequences that is commonly made in the literature.
\begin{assumption}\label{asp:noise}\emph{(Sub-Gaussian Noise)}
For each $i$, and a sequence of decisions $\x_{1}(i),\ldots, \x_{T}(i)$, the random variables $\varepsilon_{i,1},\ldots,\varepsilon_{i,T}$ are independent, have zero mean, and are sub-Gaussian: there exists a constant $c>0$ such that $\mathbb{P}(|\varepsilon_{i,t}| \geq s) \leq 2 e^{-cs^2}$ for all $s$. We also assume that they have bounded norm $\max_{i,t} \Vert \varepsilon_{i,t} \Vert_{\psi_2} \leq E_{\max}$, where $\Vert \varepsilon_{i,t} \Vert_{\psi_2} = \inf \big\{s > 0 : \mathbb{E}[\exp(\varepsilon_{i,t}^2/s^2)] \leq 2 \big\}$.
\end{assumption}

Note that the distribution of $\varepsilon_{i,t}$ in this case can potentially depend on $\x_{t}(i)$. Since several ideas will be necessary for designing a near-optimal-regret algorithm that satisfies EFTD with noisy bandit feedback, we break down this problem based on the number of groups. In Section~\ref{subsec:noisy-1-group}, we will consider the single-group setting; then, in Section~\ref{subsec:noisy-two-group}, we will consider the case of $N=2$ groups; finally, in Section~\ref{subsec:noisy-n-group}, we will build upon these ideas for $N > 2$.

\subsection{Algorithm design for \texorpdfstring{$N=1$}{N=1} group}
\label{subsec:noisy-1-group}

We begin by discussing the challenges inherent to decision-making under noisy bandit feedback even in the simplest single dimensional case.

\begin{algorithm}[tb]
\footnotesize
\DontPrintSemicolon
\SetKwInOut{Input}{input}
\Input{smoothness $\beta \geq 1$, time horizon $T$, feasible set $\mathcal{X}=[x_{\min},x_{\max}]$, initial lag size $\delta_1$, lag transition parameter $\gamma = 16\beta$, $q \in (0,1)$, noise sub-Gaussian norm bound $E_{\max}$, and gradient adjustments $\varepsilon_1(x) = (1+\beta)x$ and $\varepsilon_2(x) = x$} 
$\delta_i \leftarrow q^{i-1} \delta_1$ for $i \geq 2$, $\xi \leftarrow 1 - q$, $n(d) = \frac{2E_{\max}^2  \log \frac{2}{p}}{d^4}$ for any $d$, $x_1 \leftarrow x_{\min} + \delta_1$, $t \leftarrow 1$, $i \leftarrow 0$ \;

\Repeat{$T$ samples have been taken}{
    \Repeat{$- g_t^{(i)} > \gamma \delta_i$}{
        $\overline{f}(x_t - \delta_i) \leftarrow$ average of $n(\xi \delta_i)$ samples at $x_t - \delta_i$ \tcp*[f]{estimate $f(x_t-\delta_i)$} \;
        $\overline{f}(x_t - \delta_{i+1}) \leftarrow$ average of $n(\xi \delta_i)$ samples at $x_t - \delta_{i+1}$ \tcp*[f]{estimate $f(x_t-\delta_{i+1})$}\;
        $g_t^{(i)} \leftarrow \frac{\overline{f}(x_t - \delta_{i+1}) - \overline{f}(x_t - \delta_{i}) }{\xi \delta_i} + \varepsilon_1(\delta_i)$ \tcp*[f]{compute the approximate secant}\;
        \If{$-g_t^{(i)} \leq \gamma \delta_i$}{
            $i \leftarrow i + 1$
        }
    }
    $\overline{f}(x_t) \leftarrow$ average of $n(\delta_i)$ samples at $x_t$ \tcp*[f]{estimate $f(x_t)$} \;
    $\tilde{\tilde{\nabla}}_t \leftarrow \frac{\overline{f}(x_t) - \overline{f}(x_t - \delta_{i})}{\delta_i} + \varepsilon_2(\delta_i)$ \tcp*[f]{compute the approximate secant} \;
    Compute $x_{t+1} \leftarrow x_t-\delta_i - \frac{1}{\beta} \tilde{\tilde{\nabla}}_t$\;
    \If{$x_{t+1} > x_{\max}$}{
        Set $x_s \leftarrow x_{\max}$ for all $t+1 \leq s \leq T$
    }
    $t \leftarrow t + 1$ \;
}
\caption{Adaptive Lagged Gradient Descent ({\sc Ada-Lgd})}
\label{alg:continual-lgd-convex}
\end{algorithm}

\paragraph{\it Challenges due to noisy bandit feedback.}\label{sec:bandit-noisy} Let us try to replicate the approach in {\sc Lagged Gradient Descent}, in which we utilize secant calculations with a fixed lag size $\delta$ to estimate the gradient step, and stop updating the iterate when the gradient becomes $\mathcal{O}(\delta)$ in magnitude. In the absence of noise, the secant is always sandwiched between the true gradients at $x_t$ and the lagged point $x_t-\delta$ by the mean value theorem. Thus, moving from the lagged point using the secant ensures that the algorithm never overshoots the optimum. But when the function evaluations are noisy, it takes $\Theta(1/\delta^4)$ function evaluations at $x_t-\delta$ and $x_t$ to evaluate the secant $\frac{f(x_t)-f(x_t-\delta)}{\delta}$ up to an error of $\mathcal{O}(\delta)$, by Hoeffding's inequality. Such high sampling rates may be acceptable when the algorithm iterates are very close to the optimum, but will lead to a high (linear) regret when the iterates are farther from the optimum. Thus, $\delta$ cannot be set to be too small. To see this, suppose one uses $\delta=T^{-1}$, the number of samples required at each point is $\Omega(T^4)$ which is too high.

At the same time if $\delta$ is too large then the algorithm may stop far from the optimum, since the local gradient may prematurely become small enough relative to the lag size that jumping from a lagged point may violate monotonicity. By the PL inequality, stopping at a point $x_t$ with $|x^*-x_t|<\delta$ implies that $h_t = f(x_t)-f(x^*) \leq c\delta^2$ for some constant $c>0$ at the stopping point. Thus, assuming $h_t$ decays exponentially (which will be shown below), the regret would be bounded by a polylogarithmic multiple of
\[
\frac{1}{\delta^4} \sum_t h_t + cT\delta^2 = \frac{r}{\delta^4} + cT\delta^2 ~~\mbox{for some constant $r>0$.}
\]
One can balance the sampling regret of $\delta^{-4}$ with the stopping regret of $T\delta^2$ by choosing $\delta=T^{-1/6}$, which results in a regret bound of $\tilde{\mathcal{O}}(T^{2/3})$.

However, it turns out that we can do strictly better and obtain the near-optimal regret rate of $\tilde{\mathcal{O}}(\sqrt{T})$ by choosing the lag sizes {\it adaptively}. The key idea is that if the algorithm stops moving with a particular lag size $\delta$, then we reduce the lag size so that the algorithm can continue to proceed. This ensures that smaller lag sizes and correspondingly higher sampling rates are utilized only when the iterates are closer to the optimum when they do not result in high regret. 

This approach, however, presents a crucial challenge: when estimating the secant at a point $x_t$ and a lagged point $x_t-\delta$, the decision of whether the lag size must be reduced from $\delta$ to some smaller quantity must be made {\it before} we sample at $x_t$ to insure monotonicity of the iterates. To address this challenge, we design a novel algorithm that respects monotonicity while searching for the correct lag size. 

\paragraph{\it Adaptive Lagged Gradient Descent.} We develop a novel procedure called {\sc Adaptive Lagged Gradient Descent (Ada-Lgd)} (Alg. \ref{alg:continual-lgd-convex}) in this section. In this procedure, there are ``non-lagged'' iterates, denoted as $(x_t)_{t\in\mathbb{N}}$ (abusing notation), and ``lagged'' iterates denoted in relation to the non-lagged iterates, e.g., $x_t-\delta_i$ for some specified lag size $\delta_i$. For any non-lagged iterate $x_t$ such that $x_{t+1} = x_t - \frac{1}{\beta} \tilde{\tilde{\nabla}}_t - \delta_i$, we say that $\delta_i$ is the lag size of $x_t$. 

We now describe how {\sc Ada-Lgd} reduces the lag sizes in a monotonic manner. Suppose the current lag size is $\delta_i$, and we are sampling at $x_t-\delta_i$. Right after sampling at $x_t - \delta_{i}$, we sample at $x_t - \delta_{i+1}$ (where $\delta_{i+1}=q\delta_i$ for some $q<1$). This has the benefit of providing a gradient estimate at $x_t - \delta_i$, which in turn gives us an estimate of the gradient at $x_t$. The latter can be used in deciding whether or not the lag size should indeed be reduced to $\delta_{i+1}$ or lower. If yes, then we continue to sample at $x_t - \delta_{i+2}$ and continue the search for the right lag size; else, we finally sample at $x_t$ and continue the secant descent procedure. Such pre-emptive sampling to search for the correct lag size thus guarantees monotonicity.

\begin{figure}[t] 
\begin{center}
\scalebox{.8}{
    \begin{tikzpicture}
    \tikzstyle{vertex}=[circle, draw, fill = black, fill opacity = 1, text opacity = 1,inner sep=1.4pt]
    \draw[<->] (0,0) -- node[below] {} (13,0);
    \node[vertex,green](v) at (1, 0) {};
    \node at (1,-1){$x_{\min} = x_1-\delta_1$};
    \draw[->] (1,-.8) -- (1,-.2);
    \node[vertex](v) at (3-2*.7,0){};
    \draw[green] (3-2*.7,0) arc (0:180:.3);
    \node[vertex,blue](v) at (3-2*.7*.7,0){};
    \draw[blue] (3-2*.7*.7,0) arc (0:180:.5*2*.7-.5*2*.7*.7);
    \node[vertex,blue](v) at (3, 0) {};
    \node at (8-5*.7,-.3){$x_2-\delta_2$};

    \draw[fill=gray, fill opacity=.2, gray, opacity=.2] (2.6,.5) -- (8-5*.7*.7+.4,.5) -- (8-5*.7*.7+.4,-.5) -- (2.6,-.5) -- cycle;
    \node at (3,-.3){$x_1$};
    \draw[fill=gray, fill opacity=.2, gray, opacity=.2] (7.6,.5) -- (12-4*.7*.7+.4,.5) -- (12-4*.7*.7+.4,-.5) -- (7.6,-.5) -- cycle;
    \draw[fill=gray, fill opacity=.2, gray, opacity=.2] (11.6,.5) -- (12.4,.5) -- (12.4,-.5) -- (11.6,-.5) -- cycle;
    \draw[fill=gray, fill opacity=.2, gray, opacity=.2] (.8,.5) -- (3-2*.7+.2,.5) -- (3-2*.7+.2,-.5) -- (.8,-.5) -- cycle;

    \draw[dashed,->] (3-2*.7,.4) -- (3-2*.7,1.7) -- node[fill=white]{$x_2 = x_1- \delta_2 - \frac{1}{\beta} \tilde{\tilde{\nabla}}_1(x_1-\delta_2,x_1)$} (8,1.7) -- (8,.4);
    
    \draw[dashed,->] (8-5*.7^4,-.4) -- (8-5*.7^4,-1.7) -- node[fill=white]{\small $x_3 = x_2 - \delta_5 - \frac{1}{\beta} \tilde{\tilde{\nabla}}_2(x_2-\delta_5,x_2)$} (12,-1.7) -- (12,-.6);
    
    \draw[dashed] (12-4*.7,.4) -- (12-4*.7,1.7) -- (13,1.7);
    
    \foreach \a in {1,...,2}{
    \node[vertex,blue](v) at (8-5*.7^\a, 0) {};
    \draw[blue] (8-5*.7^\a,0) arc (0:180:.5*5*.7^\a/.7-.5*5*.7^\a);
    }
    \foreach \a in {3,...,5}{
    \node[vertex,orange](v) at (8-5*.7^\a, 0) {};
    \draw[orange] (8-5*.7^\a,0) arc (0:180:.5*5*.7^\a/.7-.5*5*.7^\a);
    }
    \node[vertex](v) at (8, 0) {};
    \node at (8,-.3){$x_2$};
    
    \node at (12-4*.7,-.3){$x_3-\delta_5$};
    \foreach \a in {1,...,2}{
    \node[vertex,orange](v) at (12-4*.7^\a, 0) {};
    \draw[orange] (12-4*.7^\a,0) arc (0:180:.5*4*.7^\a/.7-.5*4*.7^\a);
    }
    \node[vertex](v) at (12, 0) {};
    \node at (12,-.3){$x_3$};

    
    \tikzstyle{vertex}=[draw, color = black, fill = black, text opacity = 1, inner sep = 1.4pt]
    \node[vertex,green](v) at (1,0){};
    \node[vertex,green](v) at (3-2*.7,0){};
    \node[vertex,blue](v) at (3,0){};
    \node[vertex,blue](v) at (8-5*.7,0){};
    \node[vertex,blue](v) at (8-5*.7*.7,0){};
    \node[vertex,orange](v) at (8,0){};
    \node[vertex,orange](v) at (12-4*.7,0){};
    \node[vertex,orange](v) at (12-4*.7*.7,0){};
    \node[vertex,orange](v) at (12,0){};
    \end{tikzpicture}
}
\scalebox{.8}{
    \begin{tikzpicture}
    \tikzstyle{vertex}=[circle, draw, fill = black, fill opacity = 1, text opacity = 1,inner sep=1.4pt]
    \draw[<->] (0,0) -- node[below] {} (13,0);
    \node[vertex](v) at (2, 0) {};
    \node at (2,-.3){$x_3$};
    
    \draw[dashed,->] (0,1.7) -- node[fill=white]{$x_4 = x_3 - \delta_5 - \frac{1}{\beta} \tilde{\tilde{\nabla}}_3(x_3-\delta_5,x_3)$} (7.5,1.7) -- (7.5,.4);
    
    \draw[fill=gray, fill opacity=.2, gray, opacity=.2] (1.6,.5) -- (7.5-5.5*.7*.7+.4,.5) -- (7.5-5.5*.7*.7+.4,-.5) -- (1.6,-.5) -- cycle;

    \draw[fill=gray, fill opacity=.2, gray, opacity=.2] (7.1,.5) -- (11.5-4*.7*.7+.4,.5) -- (11.5-4*.7*.7+.4,-.5) -- (7.1,-.5) -- cycle;
    \draw[fill=gray, fill opacity=.2, gray, opacity=.2] (11.1,.5) -- (11.9,.5) -- (11.9,-.5) -- (11.1,-.5) -- cycle;

    \draw[dashed,->] (7.5-5.5*.7^4,-.4) -- (7.5-5.5*.7^4,-1.7) -- node[fill=white]{\small $x_5 = x_4 - \delta_8 - \frac{1}{\beta} \tilde{\tilde{\nabla}}_4(x_4-\delta_8,x_4)$} (11.5,-1.7) -- (11.5,-.6);
    
    \node at (7.5-5.5*.7,-.3){$x_4-\delta_5$};
    \foreach \a in {1,...,2}{
    \node[vertex,orange](v) at (7.5-5.5*.7^\a, 0) {};
    \draw[orange] (7.5-5.5*.7^\a,0) arc (0:180:.5*5.5*.7^\a/.7-.5*5.5*.7^\a);
    }
    \foreach \a in {3,...,5}{
    \node[vertex,purple](v) at (7.5-5.5*.7^\a, 0) {};
    \draw[purple] (7.5-5.5*.7^\a,0) arc (0:180:.5*5.5*.7^\a/.7-.5*5.5*.7^\a);
    }
    \node[vertex](v) at (7.5, 0) {};
    \node at (7.5,-.3){$x_4$};
    
    \node at (11.5-4*.7,-.3){$x_5-\delta_8$};
    \foreach \a in {1,...,2}{
    \node[vertex,purple](v) at (11.5-4*.7^\a, 0) {};
    \draw[purple] (11.5-4*.7^\a,0) arc (0:180:.5*4*.7^\a/.7-.5*4*.7^\a);
    }
    \foreach \a in {3,...,3}{
    \node[vertex,brown](v) at (11.5-4*.7^\a, 0) {};
    \draw[brown] (11.5-4*.7^\a,0) arc (0:180:.5*4*.7^\a/.7-.5*4*.7^\a);
    }
    \node[vertex](v) at (11.5, 0) {};
    \node at (11.5,-.3){$x_5$};
    
    
    \tikzstyle{vertex}=[draw, color = black, fill = black, text opacity = 1, inner sep = 1.4pt]
    \node[vertex,orange](v) at (2,0){};
    \node[vertex,orange](v) at (7.5-5.5*.7,0){};
    \node[vertex,orange](v) at (7.5-5.5*.7*.7,0){};
    \node[vertex,purple](v) at (7.5,0){};
    \node[vertex,purple](v) at (11.5-4*.7,0){};
    \node[vertex,purple](v) at (11.5-4*.7*.7,0){};
    \node[vertex,brown](v) at (11.5,0){};
    \end{tikzpicture}
}
\end{center}
\vspace{-7pt}
\caption{\small Illustration of the points: the algorithm starts exploring at $x_{\min} = x_1 - \delta_1$ followed by $x_1 - \delta_2$. In this case, Phase 1 consists of $x_1$, Phase 2 consists of $x_2$ and $x_3$, Phase 3 consists of $x_4$, and Phase 4 consists of $x_5$; the step-size indices are $n_1= 2, n_2=5, n_3=8$, and $n_4=9$. The computation of $x_{t+1}$ is given by approximate gradient from the {\it chosen} lagged point, as depicted by the dotted lines, using the estimate $\tilde{\tilde{\nabla}}_t(x_t - \delta_i,x_t)$ obtained by sampling at $x_t-\delta_i$ and $x_t$.
\vspace{-10pt}
} 
\label{fig:example-path} 
\end{figure}

For ease of exposition, it is useful to partition the non-lagged iterates into {\it phases} based on their lag sizes. More precisely, we say that the \emph{lag size} of non-lagged iterate $x_t$ is $\delta_i=q^{i-1}\delta_1$ if $x_{t+1} = x_t - \delta_i - \tilde{\tilde{\nabla}}_t(x_t-\delta_i,x_t)$, where $\tilde{\tilde{\nabla}}_t(x_t-\delta_i,x_t)$ is the gradient estimate obtained from samples at $x_t-\delta_i$ and $x_t$. The non-lagged iterates with the same lag size form a \emph{phase}, and the lag size of iterates in Phase $j$ is denoted $\delta_{n_j}=q^{n_j-1}\delta_1$.

For example, in Figure~\ref{fig:example-path}, $x_1$ is the only member of Phase 1, and the members of Phase 2 are $x_2$ and $x_3$. Note that multiple lag sizes can be skipped between $x_t$ and $x_{t+1}$, so it may be the case that $n_{i+1} > n_i + 1$ for phases $i$ and $i+1$. For example, the first jump in Figure~\ref{fig:example-path} is taken with a lag size of $\delta_2=q\delta_1$. While sampling at $x_2-\delta_2$ and $x_2-\delta_3$, since the estimated gradient is not steep enough, the algorithm begins sampling at $x_2-\delta_4$. The algorithm decreases the lag size twice more before deciding that $\delta_5$ is an appropriate lag size. 
Theorem~\ref{prop:lgd-noisy-bandit-dynamic-lags} formalizes the regret guarantee achieved: 

\begin{theorem}
\label{prop:lgd-noisy-bandit-dynamic-lags}
Assume that $f$ satisfies Assumption~\ref{asp:st-sm}, $x^* = \arg\,\min_{x \in \mathbb{R}} f(x) \in (x_{\min},x_{\max}]$, and assume the noise is mean zero, independent, and sub-Gaussian of bounded sub-Gaussian norm (Assumption~\ref{asp:noise}). Then {\upshape \textsc{Ada-Lgd}} (Algorithm \ref{alg:continual-lgd-convex}) satisfies EFTD and, with an input of $\delta_1 = 1/\log T$, $\gamma = 16\beta$, any $q \in (0,1)$, and $p = T^{-2}$, incurs regret of order $\mathcal{O}\left( (\log T)^2 T^{1/2} \right)$.
\end{theorem}

There are several steps in the proof of Theorem \ref{prop:lgd-noisy-bandit-dynamic-lags}, which are presented in Appendix~\ref{app:single-group}. Here, we present a proof sketch. 

{\it Proof sketch.} We first show that all the secant estimates are accurate using Hoeffding's inequality (Claim~\ref{claim:gradient-accuracy} in the proof). Then, by shifting the secant estimate slightly (line 11 in the algorithm), we obtain with high probability an underestimate (in magnitude) of the true secant. Thus, by the same smoothness argument used in the proof of Lemma~\ref{prop:lagged-gradient-descent-backtrack-convergence-pl}, jumping by the shifted secant estimate will avoid overshooting (Claim~\ref{claim:overshooting}). We then show that the iterates are monotonic (Claim~\ref{claim:monotonicity}), and that we achieve exponential convergence to the optimum (Claim~\ref{claim:convergence}) (again, with high probability) across all the iterates, regardless of lag size. 

With these facts in mind, we can proceed to the regret analysis. To that end, let $q^{n(t)}\delta_1$ be the lag size of $x_t$. Then, the number of samples taken at $x_t$ is approximately $1/q^{4n(t)} \approx 1/\nabla f(x_t)^4$. Putting all of this together, the expected regret is of order 
\begin{equation} \label{eqn:regret-noisy-1d-unseparated}
    \sum_t \frac{h_t}{|\nabla f(x_t)|^4}.
\end{equation} 
To bound (\ref{eqn:regret-noisy-1d-unseparated}), we consider cases where $|\nabla f(x_t)| \geq T^{-1/4}$ and $|\nabla f(x_t)| < T^{-1/4}$ separately. Letting $T'$ be the number of non-lagged iterates until the gradient is less than $T^{-1/4}$ in magnitude, one can show that $T' \in \mathcal{O}(\log T)$. Thus, we can bound regret as follows, where $g \preceq h$ means that $g \in \widetilde{\mathcal{O}}(h)$.
\begin{align*}
    \operatorname{regret}_T &\preceq \sum_{t=1}^{T'} \frac{h_t}{\nabla f(x_t)^4} + \sum_{\substack{\text{remaining} \\  \text{pts $x$}}} \big(f(x) - f(x^*) \big) \overset{\text{PL condition}}{\preceq} \sum_{t=1}^{T'} \frac{1}{\nabla f(x_t)^2} + \sum_{\substack{\text{remaining} \\  \text{pts $x$}}} \nabla f(x)^2\\
    &\preceq T^{1/2}T' + T^{-1/2}T \preceq T^{1/2}. \hfill \qed 
\end{align*}

\subsection{Algorithm design for \texorpdfstring{$N=2$}{N=2} group}
\label{subsec:noisy-two-group}

Recall from Section~\ref{subsec:noiseless-multi-group} that the EFTD constraints impose \emph{group-wise monotonicity} constraints and \emph{cross-coordinate} constraints. Our algorithmic approach to extend the algorithms in multiple dimensions to the bandit setting are similar at a high level to \textsc{NC}$^2$-\textsc{Lgd}, however there are critical modifications that are necessary to deal with noisy feedback. In particular, we incorporate adaptive lags, which allows us to increase the sampling rate as iterates approach the constrained optimum as we did in {\sc Ada-Lgd}. We build the approach from the case of $N=2$ for ease of exposition, and then argue convergence in the general case of $N>2$ by induction in Section \ref{subsec:noisy-n-group}. Recall that given two groups, the EF constraints are $\x_t(1) \geq \x_{t}(2) - s(1,2)$ and $\x_{t}(1) \geq \x_{t}(2) - s(2,1)$ for all $t \in [T]$. The latter constraints bound the decisions given to Group 1 from below by decisions given to Group 2 in the past, and vice versa. To denote the index different from ``$i$", we will use shorthand ``$-i$".

\begin{algorithm}[!t]
\scriptsize
\footnotesize
\DontPrintSemicolon
\SetKwInOut{Input}{input}
\Input{initial point $\x = (\x(1),\x(2))$, smoothness $\beta \geq 1$ of the objective function $f(\x(1),\x(2)) = f_1(\x(1)) + f_2(\x(2))$, horizon $T$, slacks $s(1,2), s(2,1)$, $x_{\min}$, $\gamma$, $q \in (0,1)$, $\xi = 1-q$, initial lag $\delta = \delta^{(1)} = \delta^{(2)}$, $\textsc{Grad}_i(x,y)= \frac{\overline{f}_i(y) - \overline{f}_i(x)}{y-x}$, where $\overline{f}_i$ is the average of $\ssize(y-x)$ function value samples, where $\ssize(d) = \frac{2 E_{\max}^2 \log \frac{2}{p}}{d^4}$, and gradient adjustments $\varepsilon_1(x) = (1+\beta)x$ and $\varepsilon_2(x) = x$} 
\textbf{\sg{Coordinate Descent Phase:}}\; 
\While{fewer than $T$ samples have been taken}{
    \textbf{if} there are no more points that can be sampled, go to {\bf Combined Phase} (line 11)\;
    \While{$\exists i$, so that $i$th coordinate can be sampled without violating EFTD constraints}{ 
        {\bf Sample:} \js{obtain samples at $f(\x)$, observing the incident samples at $f_i(\x(i))$} \;
        \textbf{Gradient checks}:\;
        \begin{enumerate}
            \item[(a)] \textbf{if} $\x(i)$ is a $j$th lagged point for $j>1$ \textbf{then} \;
                \hspace{1.3cm} \textbf{if} $-\textsc{Grad}_i((j-1)\text{st}~\mbox{lagged point},\x(i)) - \varepsilon_1(\delta^{(i)}) \leq \gamma \delta^{(i)}$:\;
                \hspace{2cm} Plan to sample next lagged point $\x(i)+\xi q \delta^{(i)}$, and update lag $\delta^{(i)} \leftarrow q\delta^{(i)}$ \;
            \item[(b)] \textbf{if} $\x(i)$ is a non-lagged point \textbf{then} \;
                \hspace{1.3cm} Let $g = \textsc{Grad}(\x(i)-\delta^{(i)},\x(i)) + \varepsilon_2(\delta^{(i)})$\;
                \hspace{1.3cm} \textbf{if} $-\frac{1}{\beta} g < T^{-1/4}$  \textbf{then} never move group $i$ in coordinate descent again\;
                \hspace{1.3cm} \textbf{else} 
                add the following points to the queue to be sampled:\; 
                \hspace{2cm} non-lagged point $y = \x(i) - \delta^{(i)} - \frac{1}{\beta} g$, and\;
                \hspace{2cm} first and second lagged points: $y-\delta^{(i)}, y-q\delta^{(i)}$ \;
                \hspace{1.3cm} \textbf{if} lagged size has dropped enough so that $\delta^{(i)} < \delta^{(-i)}$ \textbf{then} \; 
                \hspace{2cm} switch group and continue sampling
            \item[(c)] \textbf{if} $\x(i)$ is the second feasibility iterate \textbf{then} \;
                \hspace{1.3cm} Let $g = \textsc{Grad}(\mbox{previous feasibility iterate},\x(i))$\;
                \hspace{1.3cm} \textbf{if} $-\frac{1}{\beta} g \geq  \Big(\frac{(2+\gamma)\beta}{q \alpha}+1\Big) \delta^{(-i)}$ \textbf{then} \;
                \hspace{2cm} enter the combined phase \;
                \hspace{1.3cm} \textbf{else if} only one of the coordinates cannot be sampled \textbf{then} \; 
                \hspace{2cm} sample at $(\x(1),\x(2))$ for the remaining time until $T$\;
        \end{enumerate} 
        {\bf Add feasibility iterates:} if we are about to switch groups and group $i$ is at a lower lag size, then plan to sample at two feasibility iterates. 
    }
}
{\bf \sg{Combined phase to reduce to 1-dimensional problem:}}\\
\While{fewer than $T$ samples have been taken}{
    Run \textsc{Ada-Lgd} with previously sampled lagged points on function $\psi$ defined as: \\
        \hspace{1cm}\textbf{if} Group 2 is tight, set $\psi(x)=f_1(x)+f_2(x+s(2,1))$;\\
        \hspace{1cm} \textbf{else} set $\psi(x)=f_1(x)+f_2(x-s(1,2))$
}
\caption{Switch-then-Combine Adaptive LGD ({\sc SCAda-Lgd})
}
\label{alg:noisy-2segment-asymmetric-simplified-v2}
\end{algorithm}

In order to convert the continuous time dynamics of coordinate descent and cluster combination from 
Section~\ref{subsec:noiseless-multi-group} to the discrete bandit setting, we use a similar discretization as {\sc Ada-Lgd}, where for each coordinate, we compute {\it non-lagged iterates} and slowly move towards these iterates using {\it lagged iterates} while searching for the right lag size to estimate the gradient. However, due to the 2-group setting, a number of methodological adaptations must be made to overcome the following: 
\begin{itemize}
    \item[(1)] The unconstrained optimality for a particular group, i.e., $\nabla f_i(\x_{t}(i)) = 0$, can only be approximately detected. This means that (a) determining when to enter the combined phase must be carefully determined, and moreover (b) the trigger for switching between groups in the coordinate descent phase must be adapted as well (recall that reaching optimality for a group triggered the switch to the other group in the continuous-time method). 
    
    \item[(2)] Recall that the analysis of {\sc Ada-Lgd} depended on controlling the lag sizes and resulting gradient accuracy as the sampling process approached the optimum. Since the two groups may have different derivatives at the current point, their lag sizes may differ significantly as well. This raises an issue as the group with the smaller derivative will spend more time sampling, thus causing the other group to incur regret in the meantime. Moreover, if the two groups have different lag sizes, this disparity must be reconciled when entering the combined phase, where the groups must proceed jointly with a single lag size.

\end{itemize}  

We explain our novel discrete-time bandit algorithm {\sc SCAda-Lgd} below, which addresses the above-mentioned concerns. \js{A high-level description this algorithm can be found in Algorithm~\ref{alg:noisy-2segment-asymmetric-simplified-v2}, and a more detailed description can be found in Algorithm~\ref{alg:noisy-2segment-asymmetric-simplified-v2-long} in Appendix~\ref{app:noisy}. {\sc SCAda-Lgd} discretizes the continuous-time dynamics from Section~\ref{subsec:noiseless-multi-group}, resulting in the following phases:}


\noindent 
{\bf 1. Coordinate descent phase (}\textsc{SCAda-Lgd}\textbf{).} 
\js{In the coordinate descent phase of \textsc{SCAda-Lgd}, coordinates (groups) are updated in turn. That is, if $\x$ and $\mathbf{y}$ are consecutive points sampled by \textsc{SCAda-Lgd}, then $\x - \mathbf{y}$ will have at most one nonzero entry. The algorithm maintains a queue $Q_i$ \js{of points to sample} for each group $i$; to maintain monotonicity, the next point for Group $i$ to sample is always $\min Q_i$.} 
As in {\sc Ada-Lgd}, we call iterates computed using gradient jumps ``non-lagged iterates'' (i.e, $\x(i) \leftarrow \x(i) - \delta^{(i)} - g_i(\x(i))/\beta$ where $g_i(\x(i))$ denotes the estimate of the gradient at $\x(i)$), and we move towards non-lagged iterates using ``lagged iterates'' ($\x(i) - q^k\delta$) to control the accuracy of the gradients. We additionally introduce a new set of iterates, which we refer to as ``feasibility iterates," which are used to detect whether or not the combined phase should be initiated. These iterates are used to provide the group with larger lag size (and thus a lower-accuracy gradient estimate) with a higher-accuracy  gradient estimate, thus alleviating concern (4) if the combined phase is initiated.

Throughout the coordinate descent phase, gradients are estimated for the following purposes:

\begin{enumerate}
    \item[(a)] \textit{Triggering lag transitions:} If a gradient estimated at lagged point $\x(i)-\delta^{(i)}$ is of order $\delta^{(i)}$ in magnitude, then we know that we need to sample the next lagged point (with a lower lag size) as well. As was the case for \textsc{Ada-Lgd}, this check allows us to continue converging to the optimum despite the low-gradient condition being met.
   
    \item[(b)] {\it Detecting approximate optimality or triggering a switch to the other group:} If the point sampled is a non-lagged iterate in group $i$, and the gradient estimated is small enough (i.e., less than $\mathcal{O}(T^{-1/4})$), then we permanently switch to sampling group $-i$ and never optimize group $i$ in the coordinate descent phase again. This is to ensure that group $-i$ does not incur excessive regret while group $i$ is already close to its optimum and thus further improvements to group $i$ would take a large number of steps. Otherwise, the next non-lagged and lagged iterates are computed (as in \textsc{Ada-Lgd}) and added to the queue to sample. If the lag size of group $i$ is smaller than the lag size of group $-i$, then a switch to optimizing group $-i$ is triggered. This is to prevent the algorithm from devoting too much time to one group without optimizing the other. Intuitively, if the slacks were high enough that the constraints are never tight, then this would ensure that the lag sizes of the two groups differ by at most a factor of $q$, in the coordinate descent phase.
    
    \item[(c)] {\it Triggering the combined phase:} When the next points to sample for group $i$ are the two feasibility iterates, then a gradient estimate is computed using the function values at these two points. At this stage, we know that group $-i$ is close to its optimum, since feasibility iterates are only added in such a scenario. So, if the gradient is large enough (i.e., greater than $\Omega(\delta^{(-i)})$), then we can infer that group $i$ is far from its optimum, and this indicates that the EF constraint is tight at the joint optimum; hence, in this case, the combined descent phase is triggered.
\end{enumerate} 

After these checks, it may be the case that (i) a switch to the other group was triggered, or that (ii) there are no more feasible points for group $i$ to sample. If either of these situations occurs, then the algorithm will switch groups, adding feasibility iterates to $Q_{-i}$ if applicable (that is, if the lag size of group $i$ is smaller than the lag size of group $-i$). If neither (i) nor (ii) occur, then we remain on group $i$ and repeat this process.

\paragraph{\bf 2. Combined descent phase (}\textsc{SCAda-Lgd}\textbf{).} The decisions for the two groups are locked together once the combined phase is initiated. This means that the function $f(\x(1),\x(2))$ can be expressed as a single-variable function $\psi(x) = f_1(x) + f_2(x+r)$ for some $r \in \mathbb{R}$. Since feasibility iterates were sampled in the group with larger lag size, both groups enter the combined phase with similar-accuracy gradients, which provide the first gradient estimate of $h$. At this point, \textsc{Ada-Lgd} is run on $h$ for the remainder of the time horizon.

We next show that {\sc SCAda-Lgd} achieves an order-optimal (up to polylog factors) regret. 

\begin{figure}[!t]
\centering 
\begin{minipage}{\textwidth} 

  \begin{minipage}[c]{0.43\textwidth} 

    \scalebox{.55}{
\begin{tikzpicture}

\def\sone{.15}
\def\stwo{.15}
\def\scale{10}
\def\done{.12}
\def\dtwo{.12}
\def\q{.5}

\draw[white, fill=gray, fill opacity=.1] (0,\scale*\stwo) -- (\scale*1-\scale*\stwo,\scale) -- (\scale,\scale) -- (\scale,\scale*1-\scale*\sone) -- (\scale*\sone,0) -- (0,0) -- cycle;
\draw (\scale*\sone,0) -- (\scale,\scale*1-\scale*\sone);
\draw (\scale*\sone,0) -- (6.5,5) node [midway, below=1pt, sloped, fill=white] () {\Large $\x(1) \leq \x(2) + s(1,2)$};
\draw (0,\scale*\stwo) -- (\scale*1-\scale*\stwo,\scale) node [near end, above=1pt, sloped, fill=white] () {\Large $\x(2) \leq \x(1) + s(2,1)$};

\node at (4.5,9.3) {\begin{tabular}{c}
    \textcolor{magenta}{Combined phase} \\
    \textcolor{magenta}{initiated ($|f_1'(\cdot)|$} \\
    \textcolor{magenta}{estimated to be} \\
    \textcolor{magenta}{large w.r.t. $\delta^{(2)}$)}
\end{tabular}};

\node at (8.5,4.5) {\begin{tabular}{c}
    \textcolor{brown}{Feasibility} \\
    \textcolor{brown}{iterates}
\end{tabular}};

\node (purptext) at (7.8,1.3) {\begin{tabular}{c}
    \textcolor{purple}{Shaded iterates for} \\
    \textcolor{purple}{Group 1 to be sampled} \\
    \textcolor{purple}{in later rounds}
\end{tabular}};

\draw[purple,->] (6.5,.8) -- (5.3,.2);

\node at (2,6.5) {\begin{tabular}{c}
    \textcolor{cyan}{Low-gradient} \\
    \textcolor{cyan}{condition met}
\end{tabular}};

\begin{scope}[shift={(4.8-\scale*\done*\q, 0)}]
      \draw[purple,rotate=0] (0,0) ellipse [x radius = 1 cm, y radius = .5 cm];
\end{scope}

\begin{scope}[shift={(7.1-\scale*\done*\q^1/2 -\scale*\done*\q^2/2, 5.6)}]
      \draw[brown,rotate=0] (0,0) ellipse [x radius = .35 cm, y radius = .2 cm];
\end{scope}
\draw[brown,->] (8,5) -- (7.1,5.4);

\draw[->] (0,0) -- node[near end, sloped, fill=white] {Group 2} (0,\scale);
\draw[->] (0,0) -- node[near end, fill=white] {Group 1} (\scale,0);

\draw[dashed] (0,0) -- (1.5,0) -- (1.5,3) -- (4.5,3) -- (4.5,5.6) -- (7.1,5.6) -- (7.1,7.1) -- (8.6,7.1);

\tikzstyle{vertex}=[draw, circle, color = black, fill = black, text opacity = 1, inner sep = 1.4pt]
\foreach \a in {0,...,1}{
    \node[purple] at (1.2-\scale*\done*\q^\a, 0) {\small $\times$};
}
\node[vertex,purple,inner sep=2pt](v) at (1.2, 0) {};

\draw[purple,dashed,opacity=.3] (0,0) -- (4.8,0);

\tikzstyle{vertex}=[draw, circle, color = black, fill = black, text opacity = 1, inner sep = 1.4pt]
\node[purple,opacity=.3] (rx1s) at (4.8-\scale*\done, 0) {\small $\times$};
\node[purple,opacity=.3] (rx2s) at (4.8-\scale*\done*\q, 0) {\small $\times$};

\node[vertex,purple,inner sep=2pt,opacity=.3](rnl1s) at (4.8, 0) {};

\tikzstyle{vertex}=[draw, circle, color = black, fill = black, text opacity = 1, inner sep = 1.4pt]
\node[purple] (rx1) at (4.8-\scale*\done, 3) {\small $\times$};
\node[purple] (rx2) at (4.8-\scale*\done*\q, 3) {\small $\times$};

\node[vertex,purple,inner sep=2pt](rnl1) at (4.8, 5.6) {};

\tikzstyle{vertex}=[draw, circle, color = black, fill = black, text opacity = 1, inner sep = 1.4pt]
\foreach \a in {0,...,1}{
    \node[cyan] at (1.5, 1.2-\scale*\done*\q^\a) {\small $\times$};
}
\node[vertex,cyan,inner sep=2pt](v) at (1.5, 1.2) {};

\foreach \a in {0,...,0}{
    \node[cyan] at (1.5, 3.9-\scale*\done*\q^\a) {\small $\times$};
}
\node[cyan,opacity=.3] (bx1s) at (1.5, 3.9-\scale*\done*\q) {\small $\times$};
\node[vertex,cyan,inner sep=2pt,opacity=.3](bnl1s) at (1.5, 3.9) {};

\draw[cyan,dashed,opacity=.3] (1.5,3) -- (1.5,3.9);
\node[cyan] (bx1) at (4.5, 3.9-\scale*\done*\q) {\small $\times$};
\node[vertex,cyan,inner sep=2pt](bnl1) at (4.5, 3.9) {};

\foreach \a in {0,...,2}{
    \node[cyan] at (4.5, 5.6-\scale*\done*\q^\a) {\small $\times$};
}
\node[vertex,cyan,inner sep=2pt](bluenl) at (4.5, 5.6) {};
\draw[cyan,->] (3,6) -- (bluenl);

\draw[cyan,dashed,opacity=.3] (4.5,5.6) -- (4.5,7.1);
\node[cyan,opacity=.3] (bx2s) at (4.5, 7.1-\scale*\done*\q) {\small $\times$};
\node[cyan,opacity=.3] (bx3s) at (4.5, 7.1-\scale*\done*\q^2) {\small $\times$};
\node[cyan,opacity=.3] (bx4s) at (4.5, 7.1-\scale*\done*\q^3) {\small $\times$};

\node[vertex,cyan,inner sep=2pt,opacity=.3](bnl2s) at (4.5, 7.1) {};

\node[cyan] (bx2) at (7.1, 7.1-\scale*\done*\q) {\small $\times$};
\node[cyan] (bx3) at (7.1, 7.1-\scale*\done*\q^2) {\small $\times$};
\node[cyan] (bx4) at (7.1, 7.1-\scale*\done*\q^3) {\small $\times$};

\node[vertex,cyan,inner sep=2pt](bnl2) at (7.1, 7.1) {};

\foreach \a in {2,...,3}{
    \node[cyan,opacity=.3] at (7.1, 7.6-\scale*\done*\q^\a) {\small $\times$};
}
\node[vertex,cyan,inner sep=2pt,opacity=.3](v) at (7.1, 7.6) {};
\draw[cyan,dashed,opacity=.3] (7.1,7.1) -- (7.1,7.6);

\draw[purple,dashed,opacity=.3] (4.5,3) -- (4.8,3);
\node[vertex,purple,inner sep=2pt,opacity=.3](rnl1ss) at (4.8,3) {};

\foreach \a in {2,...,3}{
    \node[vertex,brown,inner sep=1.3pt](v) at (8.6-\scale*\done*\q^\a, 7.1) {};
}
\node[vertex,magenta,inner sep=2pt](magn1) at (8.6, 7.1) {};

\draw [magenta,->] (6,9.7) 
      .. controls ++(0:1.2) and ++(90:1.2) .. (magn1);

\foreach \a in {1,...,2}{
    \node[vertex,brown,inner sep=1.3pt](v) at (7.1-\scale*\done*\q^\a, 5.6) {};
}

\foreach \a in {0,...,0}{
    \node[purple] (rx3) at (9.4-\scale*\done*\q^\a, 7.1) {\small $\times$};
}

\node[purple,opacity=.3] at (9.4-\scale*\done*\q, 7.1) {\small $\times$};
\node[vertex,purple,inner sep=2pt,opacity=.3](v) at (9.4, 7.1) {};
\draw[purple,dashed,opacity=.3] (8.6,7.1) -- (9.4,7.1);

\foreach \a in {2,...,3}{
    \node[magenta] at (9.3-\scale*\done*\q^\a, 7.8-\scale*\done*\q^\a) {\small $+$};
}
\node[vertex,magenta,inner sep=2pt](v) at (9.3, 7.8) {};

\node[vertex,black,inner sep=2pt](v) at (9.6, 8.1) {};
\node[above=8pt,left=0pt] at (9.6, 8.4) {$\x^*_C$};

\node[vertex,black,inner sep=2pt](v) at (10, 7.8) {};
\node[above=-9pt,left=-9pt] at (10.1, 7.4) {$\x^*$};

\foreach \a in {1,...,1}{
    \node[purple,opacity=.3] at (9.4-\scale*\done*\q^\a, 5.6) {\small $\times$};
}
\node[purple,opacity=.3] (rx3s) at (9.4-\scale*\done, 5.6) {\small $\times$};

\node[vertex,purple,inner sep=2pt,opacity=.3](v) at (9.4, 5.6) {};
\draw[purple,dashed,opacity=.3] (7.1,5.6) -- (9.4,5.6);

\begin{scope}[every path/.style={->,dotted,cyan,line width=1pt,opacity=.3}, every node/.style={inner sep=1pt}]
\path[thin,every node/.style={}](bnl1s) edge[bend left=30] (bnl1);
\path[thin,every node/.style={}](bnl2s) edge[bend left=30] (bnl2);
\path[thin,every node/.style={}](bx1s) edge[bend left=30] (bx1);
\path[thin,every node/.style={}](bx2s) edge[bend left=30] (bx2);
\path[thin,every node/.style={}](bx3s) edge[bend left=30] (bx3);
\path[thin,every node/.style={}](bx4s) edge[bend left=30] (bx4);
\end{scope}

\begin{scope}[every path/.style={->,dotted,purple,line width=1pt,opacity=.3}, every node/.style={inner sep=1pt}]
\path[thin,every node/.style={}](rnl1s) edge[bend right=30] (rnl1ss);
\path[thin,every node/.style={}](rnl1ss) edge[bend right=30] (rnl1);
\path[thin,every node/.style={}](rx1s) edge[bend right=30] (rx1);
\path[thin,every node/.style={}](rx2s) edge[bend right=30] (rx2);
\path[thin,every node/.style={}](rx3s) edge[bend right=30] (rx3);
\end{scope}

\end{tikzpicture}
}
    
  \end{minipage}
  \hfill 
  \begin{minipage}[c]{0.5\textwidth} 
    \caption{\label{fig:combined-phase-avoided-successfully}
An illustration of a potential decision trajectory of Algorithm \ref{alg:noisy-2segment-asymmetric-simplified-v2}, drawn over a shaded decision space EF. The red iterates are sampled to estimate $f_1'$, the blue iterates are sampled to estimate $f_2'$, points marked with an ``x'' are lagged iterates, brown vertices are the ``feasibility check iterates'' described in the algorithm, and pink iterates in the combined phase. Points which are infeasible when calculated are shaded and will be sampled if and when they become feasible.}
   \end{minipage}

\end{minipage}
\end{figure}

\begin{theorem}
\label{thm:lgd-noisy-bandit-fpp}
Let $f_1(.)$ and $f_2(.)$ satisfy Assumption~\ref{asp:st-sm}, $(\x^*(1),\x^*(2)) = \arg\,\min_{(\x(1),\x(2)) \in \mathbf{R}^2} f(\x(1),\x(2)) > (\x_1(1),\x_1(2))$ where $(\x_1(1),\x_1(2)) \geq (x_{\min},x_{\min})$ is the initial point, and that the noise is mean zero, independent, and sub-Gaussian of bounded sub-Gaussian norm (Assumption \ref{asp:noise}). Then {\upshape {\sc SCAda-Lgd}} (Alg. \ref{alg:noisy-2segment-asymmetric-simplified-v2}) satisfies EFTD and, on input of $\delta = 1/\log T$, $\gamma = 1 + \frac{1}{\log T}$, any $q \in (0,1)$, and $p = T^{-2}$, incurs \js{$\mathcal{O}((\log T)^2 T^{1/2})$ regret.}
\end{theorem}

\paragraph{\bf Proof sketch of Theorem \ref{thm:lgd-noisy-bandit-fpp}.} As discussed above, {\sc SCAda-Lgd} (Alg. \ref{alg:noisy-2segment-asymmetric-simplified-v2}) is composed of (1) a coordinate descent phase, in which \textsc{Ada-Lgd} is run on each group, switching between groups when boundary or low-gradient conditions are met; and (2) a combined phase, where the decisions for one group are locked with respect to the decisions for the other, and the two are optimized simultaneously. Much of the analysis is identical to that of Algorithm \ref{alg:continual-lgd-convex}, so we focus on the differences.

\paragraph{\it Properly entering the combined phase.} Suppose we sample $f_2$ at the feasibility check iterates, where the lag size $\delta^{(1)}$ of group 1 is smaller than that of group 2, i.e., $\delta^{(2)}$. Entering the combined phase when the unconstrained optimum is feasible can result in high regret, as can failing to quickly enter the combined phase when the unconstrained optimum is infeasible. We show that for some constants $c_1$ and $c_2$, with high probability, if $\x^*(2) - \x_t(2) \geq c_1 \delta^{(1)}$, the algorithm enters the combined phase, and if $\x^*(2) - \x_t(2) \leq c_2 \delta^{(1)}$, it does not. This ensures that the algorithm enters the combined phase when it is known with high confidence that the unconstrained optimum is infeasible.

\paragraph{\it Controlling the waiting regret.} The other major difference from the analysis of Algorithm \ref{alg:continual-lgd-convex} is the presence of \emph{waiting regret}: the regret incurred by group 1 while the algorithm is optimizing over group 2, and vice versa. The waiting regret has the potential to be quite large without careful algorithm design. By limiting the amount of time spent on any group in the coordinate descent phase and by permanently moving to the combined phase when gradients become small enough, we can obtain a waiting regret bound of $\tilde{\mathcal{O}}(\sqrt{T})$. In particular, given lag sizes of $\delta_{n_1} > \cdots > \delta_{n_m}$ and sampling times $T_1,\ldots,T_m$, we can bound the waiting regret by $\sum_{i=1}^m T_i \delta_{n_i}^2$. By bounding the number of lag transitions and the number of gradient-scaled jumps taken in any round, we can bound this expression further by $\sum_{i=1}^m \delta_{n_i}^{-2}$. Finally, this expression can be bounded by $\tilde{\mathcal{O}}(\sqrt{T})$ since optimization stops when the gradient on either dimension is of order $T^{-1/4}$. With these two points sorted, the analysis of Algorithm \ref{alg:continual-lgd-convex} carries through. \qed

\subsection{Algorithm design for \texorpdfstring{$N>2$}{N>2} group}
\label{subsec:noisy-n-group}

The one- and two-group algorithms for the noisy bandits setting can be extended to the $N$-group case in a similar fashion to how the one-group noiseless algorithm was extended to $N$ groups. In particular, the $N$-group algorithm operates under the following mechanics:

\begin{itemize}
    \item \textbf{Coordinate descent mechanics.} Initially, each group is considered a cluster, and clusters are optimized in turn according to \textsc{Ada-Lgd}. Once a non-lagged point or the boundary of the feasible region is reached, the algorithm starts optimizing the next cluster.

    \item \textbf{Cluster combination mechanics.} Whenever it is necessary to combine clusters (i.e., when every cluster is at a tight constraint or near its optimum, and at least one cluster is suboptimal), two clusters are combined.
\end{itemize}

See Algorithm~\ref{alg:multi-segment}, \textsc{C}$^2$-\textsc{Lgd}, in Appendix~\ref{app:n-group-noisy} for the detailed description of the algorithm. Below, we give an inductive argument showing that \textsc{C}$^2$-\textsc{Lgd} attains $\widetilde{\mathcal{O}}(T^{1/2})$ regret.

\begin{theorem}
Let $\operatorname{Regret}_N(T)$ denote the expected regret of Algorithm~\ref{alg:multi-segment} run on $N$ groups with objective function satisfying Assumption~\ref{asp:st-sm}, and \sge{suppose that} the noise satisfies Assumption~\ref{asp:noise}. For every $N \geq 2$, we have that $\operatorname{Regret}_N(T) \in \mathcal{O}(N^2 (\log T)^2 T^{1/2})$, where the big-Oh hides constants dependent on the EF polyhedron.
\end{theorem}

\begin{proof} 
The $N=2$ case is proved in Theorem~\ref{thm:lgd-noisy-bandit-fpp}. Now suppose that the claim holds for some $N \geq 2$. In other words, there is some $T_N$ and some $c \geq 1$ such that for all $T \geq T_N$,
\[\operatorname{Regret}_N(T) \leq c (N\log T)^2 T^{1/2}.\]

Now, we argue that the claims holds for $N+1$ groups as well. To that end, suppose we run the algorithm on $N+1$ groups. Without loss of generality, suppose Groups 1 and 2 are the first to be combined, and that they are combined at after $T'$ total samples.  

First, we bound the regret incurred before Groups 1 and 2 are combined. Note that there is some $s$-dependent constant $M$ which bounds the number of times a constraint can be hit (that is, the number of times a cluster's pre-truncation jump is infeasible). Also note that with high probability, a cluster can only make $\mathcal{O}(\log T)$ jumps before meeting the small gradient condition, and at most $\mathcal{O}((\log T)T^{1/2})$ samples are taken during each of these jumps. Putting all of this together, we have that $T' \leq (\log T)^2 T^{1/2}(N+1)$ with high probability. Thus, with high probability, we can bound $T'$ (and thus the regret up to time $T'$) by $\mathcal{O}((\log T)^2 T^{1/2}(N+1))$.

Next, we bound the regret incurred after Groups 1 and 2 are combined. As noted above, with high probability, $T' \in \widetilde{\mathcal{O}}(T^{1/2})$. Thus, there is some $T_{N+1}$ for which any $T \geq T_{N+1}$ will satisfy $T-T' \geq T_N$ with high probability. It follows that for all $T \geq T_{N+1}$, the expected regret incurred after time $T'$ is at most $c (N\log T)^2 T^{1/2}$.

Putting all this together, we have that 
\[
\operatorname{Regret}_{N+1}(T) \leq (\log T)^2 T^{1/2}(N+1) +  c N^2(\log T)^2 T^{1/2} \leq c(N+1)^2 (\log T)^2 T^{1/2},
\]
as desired.
\end{proof}

\section{Numerical Experiments} 
\label{sec:numerical-experiments}

In this section, we present numerical experiments involving \textsc{Ada-Lgd} and \textsc{SCAda-Lgd}. These simulations aim to demonstrate the non-monotonic behavior of existing algorithms and to validate the convergence of our algorithms to the optimum. 

\subsection{One-Group Algorithm}
\begin{figure}[!t]
    \centering
    \begin{minipage}{.31\textwidth}
    \includegraphics[trim={0.2 0.2 0.2 .3cm},clip,width = 1.1\textwidth]{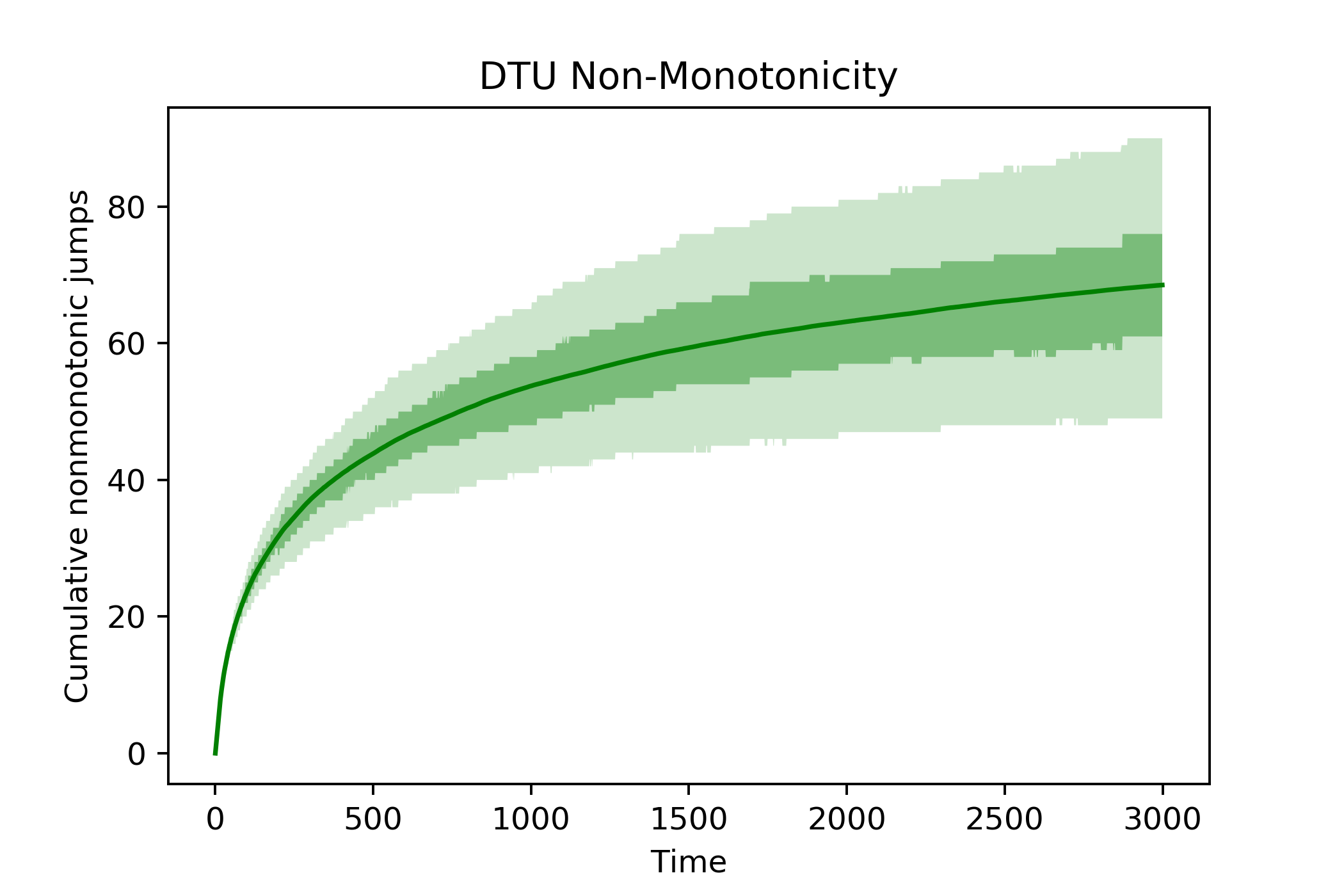}
    \end{minipage}
    \begin{minipage}{.31\textwidth}
    \includegraphics[trim={0.2 0.2 0.2 0.3cm},clip,width = 1.1\textwidth]{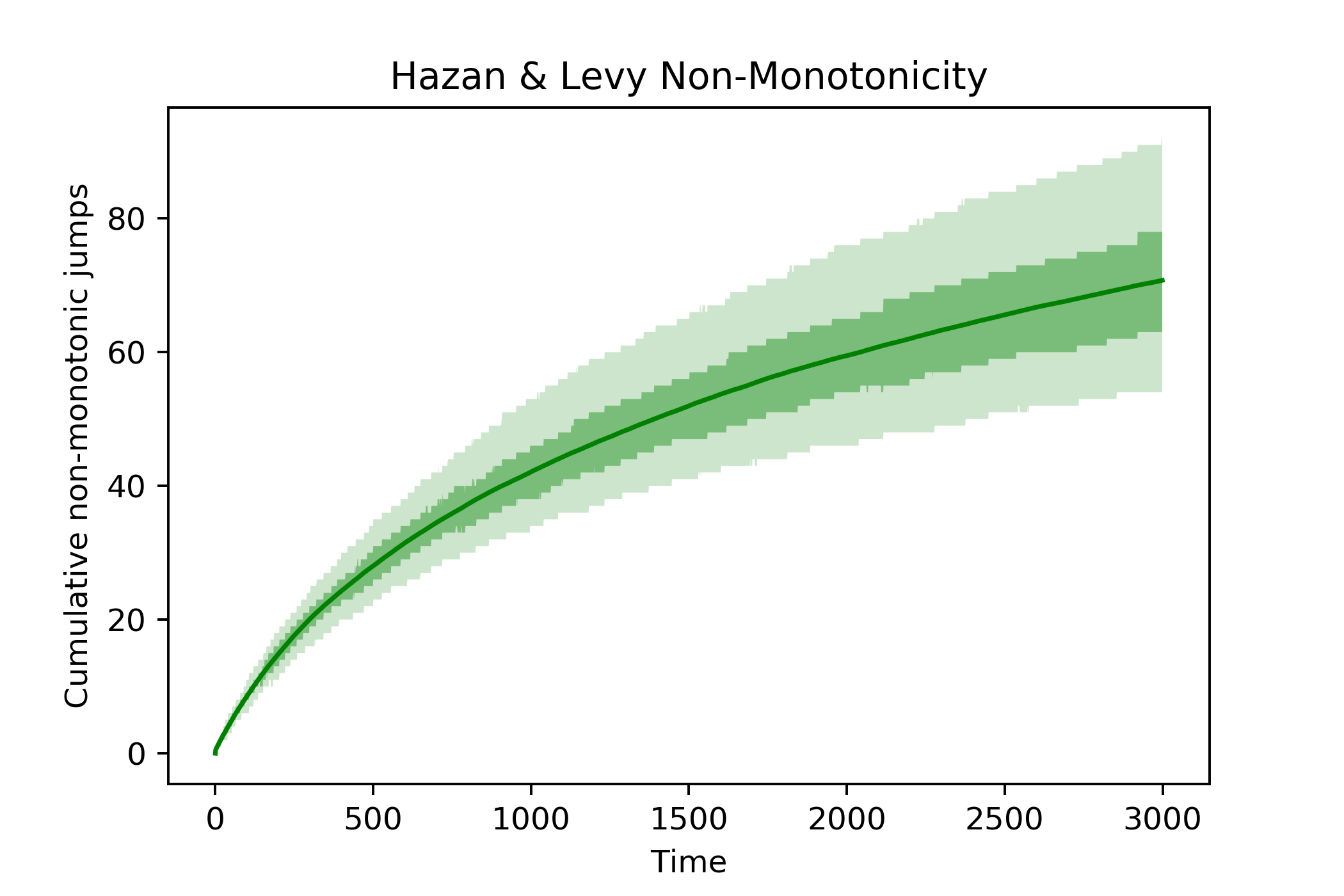}
    \end{minipage}
    \begin{minipage}{.35\textwidth}
    \includegraphics[trim={0 0 0 .1cm},clip,width = 1\textwidth]{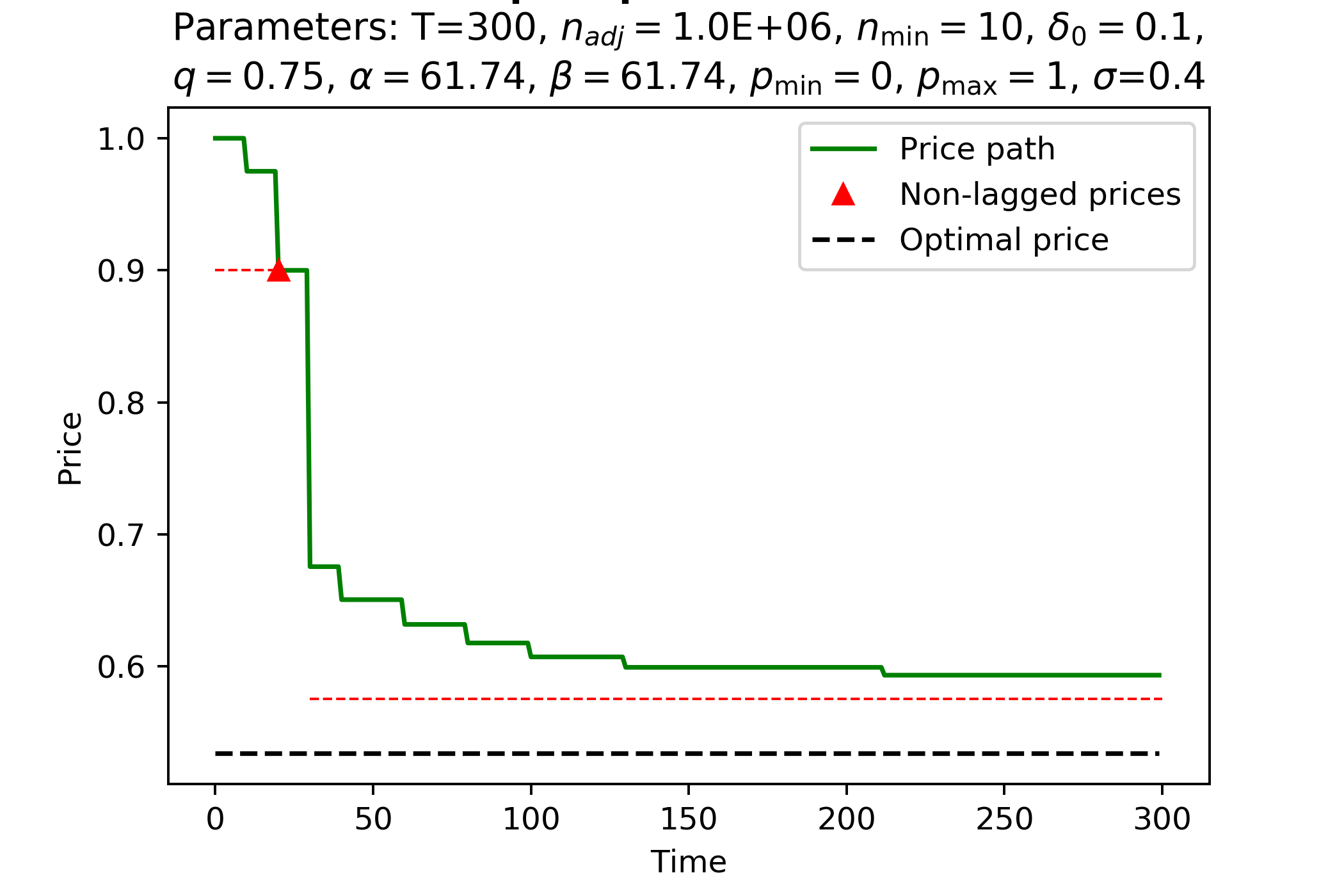}
    \end{minipage}
    \caption{\small (left) The cumulative number of non-monotonic steps taken by \textsc{Dtu}. (center) The cumulative number of non-monotonic steps (green) and the cumulative number of non-monotonic steps of size at least $0.02$ (blue) generated by \cite{hazan2014bandit}. Both simulations were on input of data-derived revenue curves corrupted by independent $N(0,.4)$ noise over a price space of $[0,1]$. (right) Price path generated by {\sc Ada-Lgd} (Alg. \ref{alg:continual-lgd-convex}). 
    }
    \label{fig:non-monotonicity}
\end{figure}
First, we compare \textsc{Ada-Lgd} to two standard state-of-the-art bandit algorithms (UCB by \cite{auer2002finite} and that of \cite{hazan2014bandit}) on data-derived revenue curves. 

We test the performance of a practical variant of \textsc{Ada-Lgd} using a trans-national UK-based online retail dataset comprised of various products, prices offered, quantities bought, purchase time, and other information.\footnote{This dataset is available online at \url{https://archive.ics.uci.edu/ml/datasets/online+retail}} We restrict the dataset to the products with at least 1000 entries (of which there were 37) and pre-process the data to account for returns, to remove 0-priced entries, and to normalize the prices to be in $[0,1]$. For each product, we remove all days for which multiple prices were offered, and use linear regression (fitting a demand curve is a typical task faced by an online retailer; linear regression is a reasonable choice here as the $R^2$-values of the products have mean $0.6475$ and standard deviation $0.2403$, with 13 products having $R^2$-value greater than $0.7$) to generate a demand curve based on per-day average quantities. After this processing, 35 of the 37 products have rich enough data to produce meaningful demand curves, so we restrict to these products. We use the demand curve for product $i$ to calculate a revenue curve $\textup{Rev}_i(\cdot)$, which we then scale so that its minimum over $\mathcal{X}$ is 0 and its maximum over $\mathcal{X}$ is 1. The feedback generated for $x_t$ on item $i$ is $\textup{Rev}_i(x_t) + \varepsilon_t$, where $\varepsilon_1,\ldots,\varepsilon_T \sim N(0,\sigma)$ are independent noise with $\sigma = 0.4$. 

While {\sc Ada-LGD} (Alg.~\ref{alg:continual-lgd-convex}) has optimal (up to logarithmic factors) regret as $T \to \infty$, its regret can be high for small values of $T$, since the number of samples taken at a point can be large relative to $T$ in this regime. To address this issue, we introduce an adjustment factor $n_{\text{adj}}$ and a minimum sampling rate $n_{\min}$. We re-define the sample size in {\sc Ada-LGD} (Alg.~\ref{alg:continual-lgd-convex}) as follows: if $\ssize(d)$ is the number of samples prescribed by the algorithm, then the number of samples taken in the experiments is $\ssize'(d) = \max\Big\{ \ssize(d)/n_{\text{adj}} , n_{\min}\Big\}$. 

We benchmark our algorithm against two bandit algorithms: a bandit convex optimization algorithm of  \cite{hazan2014bandit} which is a sample-based gradient descent algorithm using a self-concordant barrier, and a UCB algorithm \textsc{Discretize then UCB} (\textsc{Dtu}) based on a discretization of $\mathcal{X}$. \textsc{Dtu} on input of $(\log T/T)^{1/3}$-discretization obtains a regret of $\tilde{\mathcal{O}}(T^{2/3})$, which is sub-optimal, but nonetheless can have good performance. In our simulations, we use a discretization of 15 prices.

In Figure~\ref{fig:non-monotonicity}, we first plot the cumulative number of non-monotonic jumps for Hazan \& Levy and \textsc{Dtu}. We note that both algorithms ultimately make a similar number of non-monotonic jumps at time 3000, with \textsc{Dtu} making most of these jumps in early time periods. A sample monotonic price path generated by {\sc Ada-LGD} (Alg.~\ref{alg:continual-lgd-convex}) is shown in Figure~\ref{fig:non-monotonicity} (right).

\begin{figure}[!t] 
\centering
\begin{minipage}{\textwidth} 

  \begin{minipage}[c]{0.4\textwidth} 
    \includegraphics[trim = 100 250 130 265, clip, width=\textwidth]{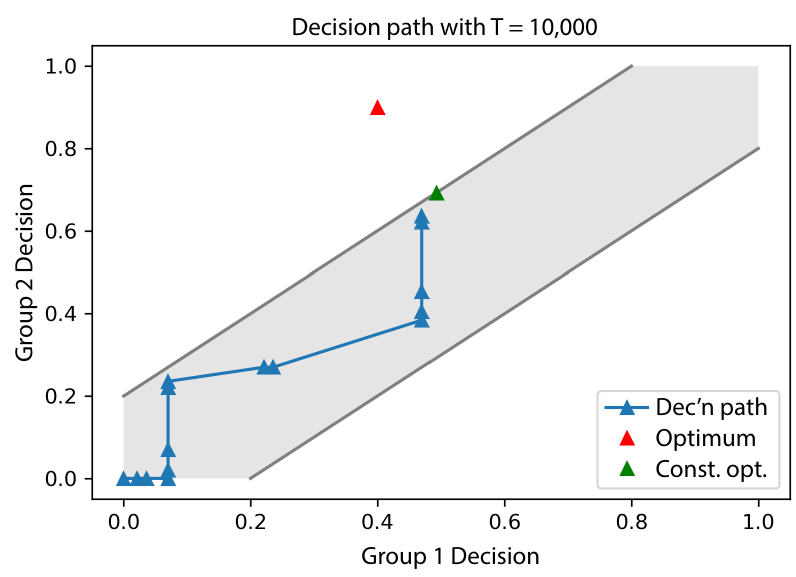}
  \end{minipage}
  \hfill
  \begin{minipage}[c]{0.40\textwidth} 
    \caption{An actual decision path generated by {\sc SCAda-Lgd} (Alg. \ref{alg:noisy-2segment-asymmetric-simplified-v2}), as discussed in Section \ref{sec:numerical-experiments}.}
    \label{fig:combined-phase-avoided-successfully-exp}
  \end{minipage}
\end{minipage}
\vspace{-0.1cm}
\end{figure}

\subsection{Two-Group Algorithm}
To validate the EFTD behavior of {\sc SCAda-Lgd}, we run it on a synthetic objective function. The functions being optimized are $f_1(x) = (x - 0.6)^2/0.36$ and $f_2(x) = (x- 0.1)^2/0.81$, which are chosen so that $\min_{x\in [0,1]} f_1(x) = \min_{x\in [0,1]} f_2(x) = 0$ and $\max_{x\in [0,1]} f_1(x) = \max_{x\in [0,1]} f_2(x) = 1$. A sample decision path for this input over $T =$ 10,000 time periods is shown in Figure~\ref{fig:combined-phase-avoided-successfully-exp}, with initial decision at (0, 0). Observe that the decisions are monotone and they converge to constrained optimum. 

\section{Discussion and Conclusion}\label{sec:discussion}

There has been a surge of interest in the academic community and among policymakers alike on the topic of {\it algorithmic fairness}. As this discourse develops, it is getting increasingly important to understand how the notions of fairness can be adapted in online decision-making scenarios. In this paper, we illustrated the rich interplay between temporal fairness desiderata and stochastic bandit convex optimization, through the novel meta-objective of ensuring \sge{envy}-freeness at the time of decision. There are several possibilities for extension of our analysis, as we discuss below. 

\begin{enumerate}

\item {\bf General polyhedra:} First, we note that our techniques heavily rely on the structure of the EF polyhedron, and algorithms for monotone stochastic convex optimization for general polyhedra remain open. In particular, we exploit the property of the EF polyhedron that if one avoids overshooting the unconstrained optimum, then one automatically avoids overshooting the constrained optimum (Claim~\ref{claim:continuous-time-overshooting}). This property does not hold in general---see Figure \ref{fig:monotonicity} for an example.


\begin{figure}[t]
    \centering
    \begin{tikzpicture}

    \draw[white, fill=gray!20] (0,2) -- (0,2.7) -- (1.3,2.7) -- (2,2) -- cycle;
    \draw[white, fill=gray!20] (2,0) -- (2.7,0) -- (2.7,1.3) -- (2,2) -- cycle;
    \draw[thin, gray, dashed] (-.3,2.7) -- (2.7, 2.7) -- (2.7,-.3);
    \draw[thin, gray, dashed] (-.3,2) -- (2, 2) -- (2,-.3);
    \draw (0,0) -- (3,0) -- (3,1) -- (1,3) -- (0,3) -- cycle;
    
    \node at (.8, 2.35) {$R_1$};
    \node at (2.35, .8) {$R_2$};
    
    \tikzstyle{vertex}=[draw, circle, color = black, fill=black, text opacity = 1, inner sep = 1.4pt]
    
    \node[vertex] at (2.7,2.7) {};
    \node at (3, 3) {$\x^*$};
    
    \node[vertex] at (2,2) {};
    \node at (1.7, 1.7) {$\x^*_C$};
    
    \draw[dotted] (6,.5) -- (9,3.5);
    \draw[dotted] (7,-.5) -- (10,2.5);
    \draw (6.3,.8) -- (8.7,3.2);
    \draw (7.3,-.2) -- (9.7,2.2);
    
    \draw[thin, gray, dashed] (5.7,2.7) -- (6.5, 2.7) -- (6.5,-.3);
    \draw[thin, gray, dashed] (5.7,1.85) -- (7.35, 1.85) -- (7.35,-.3);
    
    \node[vertex] at (6.5,2.7) {};
    \node at (6.2, 3) {$\x^*$};
    
    \node[vertex] at (7.35,1.85) {};
    \node at (7.65, 1.55) {$\x^*_C$};
    
    \node at (1.5,-.9) {\textbf{(a) Non-EF Polyhedron}};
    \node at (8,-.9) {\textbf{(b) EF Polyhedron}};
    
    \end{tikzpicture}
    \caption{Note the positions of the constrained optima $\x^*_C$ and unconstrained optima $\x^*$ in the two polyhedra. Regions $R_1$ and $R_2$ are monotone with respect to the unconstrained optimum, but not with respect to the constrained optimum in the non-EF polyhedron (left). This does not happen for the EF polyhedron on the right.}
    \label{fig:monotonicity}
\end{figure}













\item {\bf Other First-Order Optimization Methods:} Next, we note that most of the iterative first-order methods can be viewed as some form of descent over a constrained region. In fact, for $N=1$, {\sc Ada-Lgd} can be viewed as a discretization of the continuous-time dynamics of projected gradient descent or even Frank-Wolfe \cite{frank1956algorithm}. In fact, one way to ensure monotonicity is by adding additional constraints in each iteration to ensure that the next iterate is monotone. This strategy would work if one could ensure that the original constrained optimum is always contained in the intersection of the original constrained region and the monotonicity constraints imposed in each iteration. However, to ensure this, the iterates of the algorithm must never overshoot the constrained optimum in any dimension. While smoothness can be leveraged to avoid overshooting the unconstrained optimum, it is unclear how to avoid overshooting when the optimum is constrained, and the face that contains the optimum is not known.

\item {\bf Assumptions on the Function Class $\mathcal{F}$:} We presented order-optimal EFTD algorithms for smooth convex functions that satisfy the PL condition. It is open to extend the analysis or show no impact of monotonicity when the function is only smooth and convex.

We discuss the assumption of smoothness first. If the function is only assumed to be convex, then the local gradient yields no information about the proximity to the optimal point, and thus, it is impossible to control overshooting. In fact, \cite{jia2022} and \cite{chen2021multi} have leveraged this intuition to show that $\Omega(T^{3/4})$ regret is inevitable for piecewise-linear convex functions, when monotonicity is enforced. Having said that, some form of lower bound on the distance $|x-x^*|$ to the unconstrained optimum should be sufficient to control overshooting. Smoothness provides one such gradient-based bound on this distance (i.e., $|x-x^*| \geq \frac{1}{\beta} |\nabla f(x)|$). However, it might be interesting to explore other conditions. For example, if it is known that $|x-x^*| \geq c_1 |\nabla f(x)|^{c_2}$, then one can avoid overshooting by simply jumping by at most $c_1 |\nabla f(x)|^{c_2}$.

As for the PL inequality, our algorithms can still be employed in the absence of this condition, but the regret bound will no longer hold. In particular, smoothness alone implies the following in all of our algorithms:
   \vspace{-0.1cm} \[
    h_{t+1}-h_t \leq -c | \nabla f(x_t)|^2, ~\mbox{where $h_t = f(x_t) - f(x^*)$.}
    \]
From here, the PL condition implies that $h_{t+1} \leq (1-c)h_t$, which results in exponential convergence to the optimum. If only convexity and smoothness are assumed, then one can only show convergence at the slower rate of $1/t$ (Appendix \ref{sec:relaxing-PL}). This convergence rate is too slow to compensate for the sampling required to estimate gradients with enough accuracy to get a $\widetilde{\mathcal{O}}(\sqrt{T})$ regret. Whether these conditions can be relaxed to obtain this near-optimal regret bound remains an open question.

Lastly, it is unclear if the condition on separability of the function $f(\x)$ can be dropped. We include an example in Appendix~\ref{app:separability} to show that gradient descent will not converge monotonically to the optimum, even in the unconstrained setting with perfect gradient feedback.

\item {\bf Extensions of EFTD:} We believe extensions of EFTD to different applications would be very interesting, and these may involve application specific challenges. For example, one can imagine an inventory constraint at every time period that prohibits from increasing the price discount by too much. More generally, it would be interesting to explore the extensions of other static notions of fairness to temporal notions, by requiring that the decisions become more conducive over time. It might also be interesting to see how \sge{envy}-free policies might induce strategic behavior from the users. Qualitatively speaking, we can show that if the amount of delay that people are willing to strategically make is bounded, then this has little impact on the regret of our {\sc Ada-Lgd} and {\sc SCAda-Lgd} (Appendix \ref{sec:strategic}). However, we leave the broader question of strategic behavior open.

\item {\bf Legal Constraints:} Finally, we note that temporal considerations can be required or suggested for compliance with laws or policy. For example, applicant-screeners often ensure approximate demographic parity for the purpose of avoiding disparate impact litigation \cite{raghavan2020mitigating}, and this constraint constrains current decisions by prior decisions. So-called ``mandated progress'' legislation, such as affirmative action---which has been adopted and, in some cases, mandated in the U.S., Canada, and France---can similarly be thought of as a temporal constraint in which goals must be set and a current decision is thus constrained by past decisions \cite{klarsfeld2021equality}. In the context of loan-granting, Wells Fargo was recently sued for racial discrimination in mortgage lending, including offering different average interest rates to Black applicants than white applicants over a period of time \cite{waters2022wells}, which again points to the potential for the use of temporal fairness constraints as a preventative measure for avoiding litigation. Some times these constraints can be enforced due to necessity of protecting consumer rights in the prevalent societal context, e.g., price gouging laws may dictate bounded increase in decisions, during the duration of the pandemic ({\it McQueen and Ballinger v. Amazon.com}). In general, the case law surrounding algorithmic approaches to ensuring fairness is not yet well-developed. However, we believe that it is important to get ahead of such restrictions and to understand the limitations of algorithms to provide guidance to legal scholarship on the possibilities.
\end{enumerate}

\bibliographystyle{abbrv}
\bibliography{bibfile}

\begin{thebibliography}{10}

\bibitem{agarwal2011stochastic}
A.~Agarwal, D.~P. Foster, D.~J. Hsu, S.~M. Kakade, and A.~Rakhlin.
\newblock Stochastic convex optimization with bandit feedback.
\newblock {\em Advances in Neural Information Processing Systems}, 24, 2011.

\bibitem{auer2002finite}
P.~Auer, N.~Cesa-Bianchi, and P.~Fischer.
\newblock Finite-time analysis of the multiarmed bandit problem.
\newblock {\em Machine learning}, 47(2):235--256, 2002.

\bibitem{balcan2019envy}
M.-F.~F. Balcan, T.~Dick, R.~Noothigattu, and A.~D. Procaccia.
\newblock Envy-free classification.
\newblock {\em Advances in Neural Information Processing Systems}, 32, 2019.

\bibitem{bansal2017topic}
S.~Bansal, A.~Srivastava, and A.~Arora.
\newblock Topic modeling driven content based jobs recommendation engine for
  recruitment industry.
\newblock {\em Procedia computer science}, 122:865--872, 2017.

\bibitem{atlantic}
C.~Beam.
\newblock Welcome to pricing hell: The ubiquitous rise of add-on fees and
  personalized pricing has turned buying stuff into a game you can’t win,
  2024.

\bibitem{bertsimas2013fairness}
D.~Bertsimas, V.~F. Farias, and N.~Trichakis.
\newblock Fairness, efficiency, and flexibility in organ allocation for kidney
  transplantation.
\newblock {\em Operations Research}, 61(1):73--87, 2013.

\bibitem{bolton2003consumer}
L.~E. Bolton, L.~Warlop, and J.~W. Alba.
\newblock Consumer perceptions of price (un) fairness.
\newblock {\em Journal of consumer research}, 29(4):474--491, 2003.

\bibitem{moulin_2016}
F.~Brandt, V.~Conitzer, U.~Endriss, J.~Lang, and A.~D. Procaccia.
\newblock {\em Handbook of computational social choice}.
\newblock Cambridge University Press, 2016.

\bibitem{bubeck2015convex}
S.~Bubeck et~al.
\newblock Convex optimization: Algorithms and complexity.
\newblock {\em Foundations and Trends{\textregistered} in Machine Learning},
  8(3-4):231--357, 2015.

\bibitem{calders2009building}
T.~Calders, F.~Kamiran, and M.~Pechenizkiy.
\newblock Building classifiers with independency constraints.
\newblock In {\em 2009 IEEE International Conference on Data Mining Workshops},
  pages 13--18. IEEE, 2009.

\bibitem{ccpa}
CCPA.
\newblock {TITLE 1.81.5. California Consumer Privacy Act of 2018 (Title 1.81.5
  added by Stats. 2018, Ch. 55, Sec. 3.)}.

\bibitem{chen2021multi}
N.~Chen.
\newblock Multi-armed bandit requiring monotone arm sequences.
\newblock {\em Advances in Neural Information Processing Systems},
  34:16093--16103, 2021.

\bibitem{chen2022fair}
Q.~Chen, N.~Golrezaei, F.~Susan, and E.~Baskoro.
\newblock Fair assortment planning.
\newblock {\em Available at SSRN 4072912}, 2022.

\bibitem{chen2021fairness}
X.~Chen, X.~Zhang, and Y.~Zhou.
\newblock Fairness-aware online price discrimination with nonparametric demand
  models.
\newblock {\em arXiv preprint arXiv:2111.08221}, 2021.

\bibitem{cohen2022price}
M.~C. Cohen, A.~N. Elmachtoub, and X.~Lei.
\newblock Price discrimination with fairness constraints.
\newblock {\em Management Science}, 2022.

\bibitem{cohen2021dynamic}
M.~C. Cohen, S.~Miao, and Y.~Wang.
\newblock Dynamic pricing with fairness constraints.
\newblock {\em Available at SSRN 3930622}, 2021.

\bibitem{cope2009regret}
E.~W. Cope.
\newblock Regret and convergence bounds for a class of continuum-armed bandit
  problems.
\newblock {\em IEEE Transactions on Automatic Control}, 54(6):1243--1253, 2009.

\bibitem{darke2003fairness}
P.~R. Darke and D.~W. Dahl.
\newblock Fairness and discounts: The subjective value of a bargain.
\newblock {\em Journal of Consumer psychology}, 13(3):328--338, 2003.

\bibitem{Dwork2012}
C.~Dwork, M.~Hardt, T.~Pitassi, O.~Reingold, and R.~Zemel.
\newblock Fairness through awareness.
\newblock In {\em {Proceedings of the 3rd Innovations in Theoretical Computer
  Science (ITCS)}}, pages 214--226. ACM, 2012.

\bibitem{flaxman2005online}
A.~D. Flaxman, A.~T. Kalai, and H.~B. McMahan.
\newblock Online convex optimization in the bandit setting: gradient descent
  without a gradient.
\newblock In {\em Proceedings of the sixteenth annual ACM-SIAM symposium on
  Discrete algorithms}, pages 385--394, 2005.

\bibitem{foley1966resource}
D.~K. Foley.
\newblock {\em Resource allocation and the public sector}.
\newblock Yale University, 1966.

\bibitem{frank1956algorithm}
M.~Frank, P.~Wolfe, et~al.
\newblock An algorithm for quadratic programming.
\newblock {\em Naval research logistics quarterly}, 3(1-2):95--110, 1956.

\bibitem{gupta2019individual}
S.~Gupta and V.~Kamble.
\newblock Individual fairness in hindsight.
\newblock In {\em Proceedings of the 2019 ACM Conference on Economics and
  Computation}, pages 805--806, 2019.

\bibitem{haws2006dynamic}
K.~L. Haws and W.~O. Bearden.
\newblock Dynamic pricing and consumer fairness perceptions.
\newblock {\em Journal of Consumer Research}, 33(3):304--311, 2006.

\bibitem{hazan2016introduction}
E.~Hazan.
\newblock Introduction to online convex optimization.
\newblock {\em Foundations and Trends in Optimization}, 2(3-4):157--325, 2016.

\bibitem{hazan2014bandit}
E.~Hazan and K.~Y. Levy.
\newblock Bandit convex optimization: Towards tight bounds.
\newblock In {\em NIPS}, pages 784--792, 2014.

\bibitem{heidari2018preventing}
H.~Heidari and A.~Krause.
\newblock Preventing disparate treatment in sequential decision making.
\newblock In {\em Proceedings of the 27th International Joint Conference on
  Artificial Intelligence}, pages 2248--2254, 2018.

\bibitem{Heyman2008PerceptionsOF}
J.~Heyman and B.~A. Mellers.
\newblock Perceptions of fair pricing.
\newblock 2008.

\bibitem{jiaetal}
S.~Jia, A.~Li, and R.~Ravi.
\newblock Markdown pricing under unknown demand.
\newblock {\em Available at SSRN 3861379}, 2021.

\bibitem{jia2022}
S.~Jia, A.~Li, and R.~Ravi.
\newblock Dynamic pricing with monotonicity constraint under unknown parametric
  demand model.
\newblock {\em Advances in Neural Information Processing Systems},
  35:19179--19188, 2022.

\bibitem{joseph2016fairness}
M.~Joseph, M.~Kearns, J.~Morgenstern, and A.~Roth.
\newblock Fairness in learning: classic and contextual bandits.
\newblock In {\em Proceedings of the 30th International Conference on Neural
  Information Processing Systems}, pages 325--333, 2016.

\bibitem{kallus2021fairness}
N.~Kallus and A.~Zhou.
\newblock Fairness, welfare, and equity in personalized pricing.
\newblock In {\em Proceedings of the 2021 ACM Conference on Fairness,
  Accountability, and Transparency}, pages 296--314, 2021.

\bibitem{karimi2016linear}
H.~Karimi, J.~Nutini, and M.~Schmidt.
\newblock Linear convergence of gradient and proximal-gradient methods under
  the polyak-{\l}ojasiewicz condition.
\newblock In {\em Machine Learning and Knowledge Discovery in Databases:
  European Conference, ECML PKDD 2016, Riva del Garda, Italy, September 19-23,
  2016, Proceedings, Part I 16}, pages 795--811. Springer, 2016.

\bibitem{keskin2014dynamic}
N.~B. Keskin and A.~Zeevi.
\newblock Dynamic pricing with an unknown demand model: Asymptotically optimal
  semi-myopic policies.
\newblock {\em Operations research}, 62(5):1142--1167, 2014.

\bibitem{kiefer1953sequential}
J.~Kiefer.
\newblock Sequential minimax search for a maximum.
\newblock {\em Proceedings of the American mathematical society},
  4(3):502--506, 1953.

\bibitem{kiefer1952stochastic}
J.~Kiefer and J.~Wolfowitz.
\newblock Stochastic estimation of the maximum of a regression function.
\newblock {\em The Annals of Mathematical Statistics}, 23(3):462--466, 1952.

\bibitem{klarsfeld2021equality}
A.~Klarsfeld and G.~Cachat-Rosset.
\newblock Equality of treatment, opportunity, and outcomes: mapping the law.
\newblock In {\em Oxford Research Encyclopedia of Business and Management}.
  2021.

\bibitem{kleinberg2004nearly}
R.~Kleinberg.
\newblock Nearly tight bounds for the continuum-armed bandit problem.
\newblock {\em Advances in Neural Information Processing Systems}, 17:697--704,
  2004.

\bibitem{liu2017calibrated}
Y.~Liu, G.~Radanovic, C.~Dimitrakakis, D.~Mandal, and D.~C. Parkes.
\newblock Calibrated fairness in bandits.
\newblock {\em arXiv preprint arXiv:1707.01875}, 2017.

\bibitem{ma2020group}
W.~Ma, P.~Xu, and Y.~Xu.
\newblock Group-level fairness maximization in online bipartite matching.
\newblock {\em arXiv preprint arXiv:2011.13908}, 2020.

\bibitem{manshadi2021fair}
V.~Manshadi, R.~Niazadeh, and S.~Rodilitz.
\newblock Fair dynamic rationing.
\newblock In {\em Proceedings of the 22nd ACM Conference on Economics and
  Computation}, pages 694--695, 2021.

\bibitem{nesterov2017random}
Y.~Nesterov and V.~Spokoiny.
\newblock Random gradient-free minimization of convex functions.
\newblock {\em Foundations of Computational Mathematics}, 17(2):527--566, 2017.

\bibitem{raghavan2020mitigating}
M.~Raghavan, S.~Barocas, J.~Kleinberg, and K.~Levy.
\newblock Mitigating bias in algorithmic hiring: Evaluating claims and
  practices.
\newblock In {\em Proceedings of the 2020 conference on fairness,
  accountability, and transparency}, pages 469--481, 2020.

\bibitem{salem2019closing}
J.~Salem and S.~Gupta.
\newblock Closing the gap: Group-aware parallelization for the secretary
  problem with biased evaluations.
\newblock {\em Available at SSRN 3444283}, 2019.

\bibitem{sen20136}
A.~Sen.
\newblock Equality of what?
\newblock {\em Globalization and International Development: The Ethical
  Issues}, page~61, 2013.

\bibitem{sinha2021resume}
A.~K. Sinha, A.~K. Akhtar, A.~Kumar, et~al.
\newblock Resume screening using natural language processing and machine
  learning: A systematic review.
\newblock {\em Machine Learning and Information Processing}, pages 207--214,
  2021.

\bibitem{spall1992multivariate}
J.~C. Spall et~al.
\newblock Multivariate stochastic approximation using a simultaneous
  perturbation gradient approximation.
\newblock {\em IEEE transactions on automatic control}, 37(3):332--341, 1992.

\bibitem{stromquist1980cut}
W.~Stromquist.
\newblock How to cut a cake fairly.
\newblock {\em The American Mathematical Monthly}, 87(8):640--644, 1980.

\bibitem{vershynin2018high}
R.~Vershynin.
\newblock {\em High-dimensional probability: An introduction with applications
  in data science}, volume~47.
\newblock Cambridge university press, 2018.

\bibitem{waters2022wells}
T.~Waters.
\newblock Wells fargo bank sued for race discrimination in mortgage lending
  practices, 2022.
\newblock Published Apr. 26, 2022. Last accessed June 7, 2022.

\bibitem{xia2004the}
L.~Xia, K.~Monroe, J.~Cox, B.~Kent, K.~Monroe, and J.~Jones.
\newblock The price is unfair! a conceptual framework of price fairness
  perceptions.
\newblock {\em Journal of Marketing}, 68:1--15, 11 2004.

\bibitem{zafar2017fairness}
M.~B. Zafar, I.~Valera, M.~Gomez~Rodriguez, and K.~P. Gummadi.
\newblock Fairness beyond disparate treatment \& disparate impact: Learning
  classification without disparate mistreatment.
\newblock In {\em Proceedings of the 26th International Conference on World
  Wide Web}, pages 1171--1180, 2017.

\end{thebibliography}

\newpage

\appendix

\section{Analysis for Section \ref{sec:noiseless}}
\label{app:noiseless}

\subsection{Analysis for a single group} \label{app:single-group}

\textsc{Proof of Lemma \ref{prop:lagged-gradient-descent-backtrack-convergence-pl}.} We begin by proving ``1.'' To show that we never overshoot, we will exploit smoothness. In particular, for any $t$ such that $x_t \leq x^*$, we have 
\begin{align*}
    x_{t+1}-x_t' &= -\frac{1}{\beta} \tilde \nabla_t = -\frac{1}{\beta} \nabla f(\overline{x_t}) &\mbox{for some $\overline{x_t} \in [x_t',x_t]$}\\
    &= \frac{1}{\beta} \Vert \nabla f(\overline{x_t}) \Vert &\mbox{since $x_t \leq x^*$}\\
    &\leq x^* - \overline{x_t} &\mbox{since $\nabla f(x^*) = 0$} \\
    &\leq x^* - x_t' ~. \text{This proves ``1.''}
\end{align*} 
To show the convergence (``2''), we begin by bounding the gap $h_t = f(x_t) - f(x^*)$. For any $t \geq 2$, we have
\begin{align*}
    h_{t+1}-h_t &= f(x_{t+1})-f(x_t) \\
    &\leq \nabla_t^\top (x_{t+1}-x_t) + \frac{\beta}{2} ( x_{t+1}-x_t )^2 &\mbox{$\beta$-smooth} \\
    &\leq (\tilde \nabla_t^\top + \beta \delta) (x_{t+1}-x_t) + \frac{\beta}{2} ( x_{t+1}-x_t )^2 & \mbox{$\beta$-smooth} \\
    &= - \frac{1}{2\beta} \Vert \tilde{\nabla}_t \Vert^2 - \delta \tilde{\nabla}_t - \frac{\beta}{2} \delta^2 &\mbox{definiton of $x_{t+1}$} \\
    &\leq - \frac{1}{2\beta} \Vert \tilde{\nabla}_t \Vert^2 - \left(-\frac{\tilde{\nabla}_t}{(1+\gamma)\beta}\right) \tilde{\nabla}_t - \frac{\beta}{2} \delta^2 & \mbox{since $-\frac{1}{\beta} \tilde{\nabla}_t \geq (1+\gamma) \delta$} \\
    &= - \left( \frac{1}{2\beta} - \frac{1}{(1+\gamma)\beta} \right) \Vert \tilde{\nabla}_t \Vert^2 - \frac{\beta}{2} \delta^2 \leq - \underbrace{\left( \frac{1}{2\beta} - \frac{1}{(1+\gamma)\beta} \right)}_{=: c} \Vert \tilde{\nabla}_t \Vert^2.
\end{align*}
By the mean value theorem, there is some $\overline{x_t} \in [x_t',x_t] \subset [x_{t-1},x_t]$ such that $\nabla f(\overline{x_t}) = \tilde \nabla_t$. Using the PL inequality, we get:
$h_{t+1}-h_t \leq -c\Vert \tilde \nabla_t \Vert^2  = -c\Vert \nabla f(\overline{x_t}) \Vert^2  
    \leq - 2\alpha c[f(\overline{x_t}) - f(x^*)] 
    \leq - 2 \alpha c  h_t.$  
Note that for $\gamma > 1$ as specified in the lemma, $2\alpha c \in (0,1)$. So, 
\begin{align*}
h_{t+1} &\leq \big(1-2 \alpha c\big) h_t  \leq \cdots \leq \big(1-2\alpha c\big)^{t}h_1 \leq h_1 \exp\big(-2\alpha c t\big)  ~. \qed
\end{align*}

\textsc{Proof of Theorem \ref{cor:lgd-one-stopping-criteria-noiseless-pl}.}
As stated in the theorem, fix $\delta = T^{-1/2}$ 
as the lag size. Since overshooting never occurs (by Lemma \ref{prop:lagged-gradient-descent-backtrack-convergence-pl}), we need only calculate the exploration and stopping regret. The exploration regret is bounded by Lemma \ref{prop:lagged-gradient-descent-backtrack-convergence-pl}.

Now we analyze the stopping regret. If the algorithm stops at some time $t$, then it must be that 
\[
-\frac{1}{\beta} \tilde \nabla_t \leq (1+\gamma)\delta ~,
\]  
which allows us to bound the gradient: $\Vert \nabla_t \Vert \leq \Vert \tilde \nabla_t \Vert \leq \beta(1+\gamma)\delta$. In other words, if we stop at time $t$, then $\Vert \nabla_t \Vert \in \mathcal{O}(\delta)$, since $\gamma = 1 + \frac{1}{\log T} \in \mathcal{O}(1)$. Thus, by the PL inequality, the instantaneous regret after stopping is $\mathcal{O}(\delta^2)$. So, 
\begin{align*}
    \operatorname{regret} &\in \mathcal{O} \left( \sum_{t=2}^T h_1 e^{-c(t-1)} + \underbrace{T\delta^2}_{=1} \right) = \mathcal{O}(1).
\end{align*}
So, we get constant regret.
\qed\\

\subsubsection{Relaxing PL-inequality}
\label{sec:relaxing-PL}

In this section, we show that lagged gradient descent (Algorithm~\ref{alg:Lagged-2pt-Gradient-Descent-Backtracking}) achieves a regret guarantee of $\widetilde{\mathcal{O}}(1)$ for smooth and convex functions in the single dimensional case. This result relies on the following lemma.
\begin{lemma} \label{prop:lagged-gradient-descent-backtrack-convergence}
Let $f : \mathcal{X} \to \mathbf{R}$ be a $\beta$-smooth and convex function. 
Let $x_1,\ldots,x_{T/2}$ be the non-lagged points generated by {\sc Lagged Gradient Descent} (Algorithm \ref{alg:Lagged-2pt-Gradient-Descent-Backtracking}), and assume that $x_1 \leq x^* = \arg\,\min_{x \in \mathbb{R}} f(x)$. Then, for $\gamma > 1$, the following hold:
\begin{enumerate}
    \item Decisions increase monotonically toward the optimum: $x_1 \leq x_2 \leq x_3 \leq \cdots \leq x_{T/2} \leq x^*$;
    \item The convergence rate up to halting is given by: $h_t = f(x_t) - f(x^*) \leq \frac{\Vert x_1 - x^* \Vert^2}{(t-1)c}$
    , where $c = \frac{1}{2\beta} - \frac{1}{(1+\gamma)\beta}$. 
\end{enumerate}
\end{lemma}

\textsc{Proof of Lemma \ref{prop:lagged-gradient-descent-backtrack-convergence}.} 
We begin by showing that we avoid overshooting the optimum. To show that we never overshoot, we will exploit smoothness. In particular, for any $t$ such that $x_t \leq x^*$, we have 
\begin{align*}
    x_{t+1}-x_t' &= -\frac{1}{\beta} \tilde \nabla_t = -\frac{1}{\beta} \nabla f(\overline{x_t}) &\mbox{for some $\overline{x_t} \in [x_t',x_t]$}\\
    &= \frac{1}{\beta} \Vert \nabla f(\overline{x_t}) \Vert &\mbox{since $x_t \leq x^*$}\\
    &\leq x^* - \overline{x_t} &\mbox{assuming $\nabla f(x^*) = 0$} \\
    &\leq x^* - x_t' ~. 
\end{align*} 

This proves that we avoid overshooting. Note that this, along with the fact that we only move monotonically, implies the following:
\begin{equation} \label{eqn:monotonic-improvements}
    \Vert x_{t+1} - x^* \Vert^2 \leq \Vert x_t - x^* \Vert^2 ~~\mbox{for all $t \geq 1$.}
\end{equation}

Next, we bound the convergence rate. To that end, let $h_t = f(x_t) - f(x^*)$. For any $t \geq 2$, we have
\begin{align*}
    h_{t+1}-h_t &= f(x_{t+1})-f(x_t) \\
    &\leq \nabla_t^\top (x_{t+1}-x_t) + \frac{\beta}{2} ( x_{t+1}-x_t )^2 &\mbox{$\beta$-smooth} \\
    &\leq (\tilde \nabla_t^\top + \beta \delta) (x_{t+1}-x_t) + \frac{\beta}{2} ( x_{t+1}-x_t )^2 & \mbox{$\beta$-smooth} \\
    &= - \frac{1}{2\beta} \Vert \tilde{\nabla}_t \Vert^2 - \delta \tilde{\nabla}_t - \frac{\beta}{2} \delta^2 &\mbox{definition of $x_{t+1}$} \\
    &\leq - \frac{1}{2\beta} \Vert \tilde{\nabla}_t \Vert^2 - \left(-\frac{\tilde{\nabla}_t}{(1+\gamma)\beta}\right) \tilde{\nabla}_t - \frac{\beta}{2} \delta^2 & \mbox{since $-\frac{1}{\beta} \tilde{\nabla}_t \geq (1+\gamma) \delta$} \\
    &= - \left( \frac{1}{2\beta} - \frac{1}{(1+\gamma)\beta} \right) \Vert \tilde{\nabla}_t \Vert^2 - \frac{\beta}{2} \delta^2 \leq - \underbrace{\left( \frac{1}{2\beta} - \frac{1}{(1+\gamma)\beta} \right)}_{=: c} \Vert \tilde{\nabla}_t \Vert^2.
\end{align*}
Also note that by convexity,
\begin{equation} \label{eqn:convexity-cs}
    h_t \leq \nabla f(x_t) (x_t-x^*) \leq \Vert x_t - x^* \Vert \cdot \Vert \nabla f(x_t) \Vert .
\end{equation}
So, by (\ref{eqn:monotonic-improvements}), (\ref{eqn:convexity-cs}), and the above bound on $h_{t+1}-h_t$,
\begin{equation} \label{eqn:ht_gap_smooth}
    h_{t+1}-h_t \leq -c \Vert \tilde{\nabla}_t\Vert^2 \leq -c \Vert \nabla f(x_t) \Vert^2 \leq -c \frac{h_t^2}{\Vert x_t - x^* \Vert^2} ~.
\end{equation}
Letting $w_t = \frac{c}{\Vert x_t - x^* \Vert^2}$, we have that $w_th_t^2 + h_{t+1} \leq h_t$. This is equivalent to
\[
w_t \frac{h_t}{h_{t+1}} + \frac{1}{h_t} \leq \frac{1}{h_{t+1}}.
\]
In turn, since $h_t \geq h_{t+1}$, this implies that
\[
\frac{1}{h_{t+1}} - \frac{1}{h_t} \geq w_t.
\]
Thus, we have that
\begin{align*}
    \frac{1}{h_t} &\geq \frac{1}{h_t} - \frac{1}{h_1}  \\
    &= \sum_{s=1}^{t-1} \frac{1}{h_{s+1}} - \frac{1}{h_s} \\
    &\geq \sum_{s=1}^{t-1} w_s \\
    &\geq (t-1) w_1 &\mbox{since $w_1 \leq w_2 \leq \cdots$}
\end{align*}
Thus, $h_t \leq \frac{1}{(t-1) w_1}$. \qed 

Using this lemma, we can prove the following regret guarantee.

\begin{theorem} 
\label{cor:lgd-one-stopping-criteria-noiseless}
Assume that $x^* = \arg\,\min_{x \in \mathbf{R}} f(x)$
, the initial lagged point satisfies $x_1' < x^*$, and fix $\delta = T^{-1}$ and $\gamma = 1 + \frac{1}{\log T}$. Then {\sc Lagged Gradient Descent} (Alg. \ref{alg:Lagged-2pt-Gradient-Descent-Backtracking}) is a $\widetilde{\mathcal{O}}(1)$-regret EFTD algorithm for optimizing a convex and $\beta$-smooth function in the noiseless bandit setting.
\end{theorem}

\textsc{Proof of Theorem \ref{cor:lgd-one-stopping-criteria-noiseless}.}
As stated in the theorem, fix $\delta = T^{-1}$ 
as the lag size. Since overshooting never occurs (by Lemma \ref{prop:lagged-gradient-descent-backtrack-convergence}), we need only calculate the exploration and stopping regret. The exploration regret is bounded by Lemma \ref{prop:lagged-gradient-descent-backtrack-convergence}.

Now we analyze the stopping regret. If the algorithm stops at some time $t$, then it must be that 
\[
-\frac{1}{\beta} \tilde \nabla_t \leq (1+\gamma)\delta ~,
\]  
which allows us to bound the gradient: $\Vert \nabla_t \Vert \leq \Vert \tilde \nabla_t \Vert \leq \beta(1+\gamma)\delta$. In other words, if we stop at time $t$, then $\Vert \nabla_t \Vert \in \mathcal{O}(\delta)$, since $\gamma = 1 + \frac{1}{\log T} \in \mathcal{O}(1)$. Note that by convexity, we have that $h_t \leq \Vert \nabla_t \Vert (x^* - x_t) \leq \Vert \nabla_t \Vert (x^* - x_1) \leq \Vert \nabla_t \Vert \sqrt{c/w_1}$, where $c = \frac{1}{2\beta} - \frac{1}{(1+\gamma)\beta}$ and $w_1 = c/\Vert x_1 - x^*\Vert^2$. So,   
\begin{align*}
    \operatorname{regret} &\in \mathcal{O} \left( \sum_{t=2}^T \frac{1}{(t-1)w_1} + \underbrace{T\delta}_{=1} \right) = \mathcal{O}(\log T).
\end{align*}
So, we get logarithmic regret.
\qed\\

\subsection{\texorpdfstring{$N$}{N}-dimensional algorithm}
\label{app:n-dim-noiseless}

\begin{algorithm}[h]
\footnotesize
\DontPrintSemicolon
\SetKwInOut{Input}{input}
\Input{Number of groups $N$, smoothness parameter $\beta$ of $f(\x(1),\ldots,\x(N)) = \sum_{i=1}^N f_i(\x(i))$, time horizon $T$, non-negative EF slacks $s$, $x_{\min} \in \mathbb{R}$} 
Initialize queue $Q_i \leftarrow \emptyset$ for $i \in [N]$ \tcp*[f]{will contain points to query} \; 
Initialize the partition $\Pi = \big\{ \{1\},\ldots,\{N\} \big\}$ of groups, and set $\textup{active}_A = \textup{Yes}$ and $g_A = -\infty$ for each $A \in \Pi$ \;
For any cluster $A \subseteq [N]$, define $\psi_A(x) = \sum_{j \in A} f_j(x + b_j)$, where $b_j$ for $j \in A$ is defined by the tight constraints of cluster $A$ (see Appendix~\ref{app:n-dim-noiseless});  
initialize $A = \{1\}$ \;
Add $x_{\min}$ to $Q_i$ for $i \in [N]$ \;
\While{fewer than $T$ samples have been taken}{
    \If{$\textup{active}_B = \textup{No}$ for each cluster $B \in \Pi$ (i.e., no clusters can move)}{
        \uIf{each cluster is constrained by some other cluster}{
            Find clusters $C,D$ which are locked in place with respect to each other 
        }
        \Else{
            Find clusters $C \ne D \in \Pi$ such that $C$ is at a tight constraint imposed by some group in $D$, and $|\psi'_D(\x(\min D))| < T^{-1}$ \;
        }
        Update $\Pi \leftarrow \Pi \cup \{C \cup D\} \setminus \{C,D\}$ \;
        Let $A = C \cup D$, and set $g_{A} = -\infty$ and $\textup{active}_A = \textup{Yes}$\;
        Let $i = \min A$, and add $\x(i)$ to $Q_i$ \tcp*[f]{add points to the representative group's queue}
    }
    Let $i = \min A$ \tcp*[f]{choose a representative group} \;
    $\mathcal{X}_A = [\x(i), \x(i) + d(A;\x)]$ \tcp*[f]{current feasible points for cluster $A$}\;
    \If{$Q_i$ contains any elements in $\mathcal{X}_A$}{
        Let $\x(i) \leftarrow \min Q_i$, remove $\x(i)$ from $Q_i$, and update $\x(j)$ (for $j \in A \setminus \{i\}$) accordingly (i.e., letting $d = \min Q_i - \x(i)$, we set $\x(j) \leftarrow \x(j) + d$ for all $j \in A$)\; 
        Update $\textup{active}_B$ for $B \in \Pi$ as necessary (formerly tight clusters may now be active; in particular, if $B$ is at a tight constraint imposed by some group in $A$ and $B$ is not at any tight constraints imposed by any group in any other cluster, then $B$ will no longer be at a tight constraint) \;
        Calculate $\psi_A'(\x(i))$ \;
               Add $\x(i) - \frac{1}{\beta} \psi_A'(\x(i))$ to $Q_i$ \;
    }
    \If{$Q_i$ contains no elements in $\mathcal{X}_A$}{
        Set $\x(i) = \max \mathcal{X}_A$ and update $\x(j)$ (for $j \in A$) accordingly \;
        Set $\textup{active}_A = \textup{No}$ and update $\textup{active}_B$ for $B \in \Pi$ (formerly tight clusters may now be active) 
    }   
    \If{$|\psi'_B(\x(\min B))| < T^{-1}$ for every cluster $B \in \Pi$}{
        Exit the while loop and remain at point $\x$ for the remaining iterations
    }
    $A \leftarrow \varphi(A;\Pi)$ \tcp*[f]{move to the next group}\;
}
\caption{Noiseless Cycle-then-Combine Lagged Gradient Descent ({\sc NC$^2$-Lgd})}
\label{alg:multi-segment-noiseless-long}
\end{algorithm}

We begin by proving the convergence of our continuous-time algorithm, which is the backbone of our multi-dimensional algorithms (Algorithms~\ref{alg:multi-segment-noiseless} and \ref{alg:multi-segment}). Convergence of the continuous-time algorithm follows from Claim~\ref{claim:existence-of-low-gradient}, which we prove here.

{\it Proof of Claim~\ref{claim:existence-of-low-gradient}.} Suppose, for the sake of contradiction, that (1) does not hold. For each $j \in [k]$, let $i_j = \min C_j$ be a representative group from cluster $C_j$. We can therefore represent the constraints placed on $C_j$ by linear inequalities on $\x({i_j})$; i.e., for each $j \ne m \in [k]$, there exist $b_{j,m} \in \mathbb{R}$ such that
\[
\x({i_j}) \leq \x({i_m}) + b_{j,m}
\]
is the constraint placed on $C_{j}$ by $C_{m}$. Moreover, since the slacks $s(\cdot,\cdot)$ are nonnegative, $b_{j,m} \geq 0$. 

Now consider the directed graph $G = (\{C_1,\ldots,C_k\},A)$, where the arc $(C_{j}, C_{m})$ means that $C_{j}$ is constrained by $C_{m}$ (i.e., $\x({i_j}) = \x({i_m}) + b_{j,m}$). We now argue that this graph must be acyclic. For the sake of contradiction, suppose that there is a cycle $(j_1,\ldots,j_{\ell})$; i.e.,
\begin{align*}
\x({i_{j_1}}) &= \x({i_{j_2}}) + b_{j_1, j_2} \\
&\vdots \\
\x({i_{j_{\ell-1}}}) &= \x({i_{j_{\ell}}}) + b_{j_{\ell-1}, j_{\ell}} \\
\x({i_{j_{\ell}}}) &= \x({i_{j_{1}}}) + b_{j_{\ell}, j_{1}}. 
\end{align*}
This implies that $\x({i_{j_1}}) \geq \cdots \geq \x({i_{j_{\ell}}}) \geq \x({i_{j_1}})$. If these inequalities are all equalities, then $i_{j_1},\ldots,i_{j_\ell}$ can be combined into a cluster; otherwise, if one of the inequalities is strict, this also yields a contradiction. Thus, $G$ is acyclic.

Now, choose any cluster $C$ which has at least one out-going arc (which must exist, since not all clusters are at a low gradient). Find a maximal directed path beginning at $C$, and let $(C_{j}, C_{m})$ be the terminal arc. Since $G$ is acyclic, there are no out-going arcs from $C_{m}$, which means that $C_{m}$ is at a low gradient and is constraining the movement of $C_{j}$. This proves the claim. $\hfill \qed_{\text{Claim}}$ 

Next, we introduce some notation that will be used in the multi-group algorithms (Algorithms~\ref{alg:multi-segment-noiseless} and \ref{alg:multi-segment}). First, recall that a \emph{cluster} is a set of groups $C=\{i_1,\ldots,i_m\}$ which are optimized jointly. That is, if this cluster is formed at time $t$, then for any $t' \geq t$ and any $i,j \in C$,
\[
\x_{t'}(i) - \x_{t'}(j) = \x_t(i) - \x_t(j).
\]
Thus, beyond time $t$, we can express the objective of cluster $C$ as a single-variable function. More precisely, let $i = \min C$, and let $b_j = \x_t(j) - \x_t(i)$. Then for any $t' \geq t$, we can express the objective for cluster $C$ as
\[
\psi_C(\x_{t'}(i)) = \sum_{j \in C} f(\x_{t'}(i) + b_j) = \sum_{j \in C} f(\x_{t'}(j)).
\]

Next, we define the \emph{succession function}, which cycles between the different clusters of groups.

\begin{definition}[succession function]
Let $N\in \mathbb{N}$ and let $\Pi$ be a partition of $[N]$. For any $A \in \Pi$, define the successor $\varphi(A;\Pi) := \begin{cases} \argmin_{B \in \Pi : \min B > \min A} \min B &\mbox{if $\exists B \in \Pi$ with $\min B > \min A$} \\ \argmin_{B \in \Pi} \min B &\mbox{else.} \end{cases}$
\end{definition}
Note that $\varphi(\,\cdot\,;\Pi)$ corresponds to the linear ordering of $\Pi$ by the minimum elements of its blocks.

Next, we define the \emph{feasibility distance}, which quantifies the extent to which a cluster $A$ can move without violating any of the constraints.

\begin{definition}[feasibility distance]
Let $N \in \mathbb{N}$, let $\emptyset \ne A \subsetneq [N]$, let $s : [N] \times [N] \to \mathbb{R}^{\geq 0}$, and let $x \in \mathbb{R}^N$ satisfy. Then the \emph{feasibility distance} of $A$ is $d(A;x) = \min_{i \in A, j \not \in A} x_j + s(i,j) - x_i$.
\end{definition}

Thus, Algorithm \ref{alg:multi-segment-noiseless-long} (which is the more detailed version of Algorithm~\ref{alg:multi-segment-noiseless}) can be thought of as coordinate descent algorithm where (1) one cycles through coordinates according to the succession function, (2) jumps are truncated at the boundary of the feasible region, (3) the distance of a point to the boundary along a particular axis is the feasibility distance, and (4) coordinates are combined whenever doing so allows for further optimization. As discussed below, this algorithm yields $\widetilde{\mathcal{O}}(1)$ regret.

\textsc{Theorem~\ref{prop:multi-group-noiseless}.} The multi-segment algorithm assuming perfect gradient access, Algorithm~\ref{alg:multi-segment-noiseless} (i.e., Alg.~\ref{alg:multi-segment-noiseless-long}), attains a regret of $\widetilde{\mathcal{O}}(1)$, ignoring factors of $N$.

{\it Proof of Theorem \ref{prop:multi-group-noiseless}.} We first argue that the algorithm is correct; i.e., that the cluster-combining procedure converges to the optimum in continuous time. In other words, we must show that whenever no clusters can move in isolation and the constrained optimum has not been attained, there are two clusters who's combination will allow for further optimization. This, however, follows from Claim~\ref{claim:existence-of-low-gradient}.

Next, we claim that the number of incomplete gradient descent jumps (i.e., those which cannot be fully made due to the EFTD constraints) can be bounded by $\mathcal{O}(N^2 \log T)$. To see this, note that each cluster converges to its optimum at an exponential rate when jumps are complete. Thus, it takes a cluster $C$ $\mathcal{O}(\log T)$ complete gradient descent jumps until it reaches the small-gradient condition. In the worst case, each of these jumps causes all other clusters to have incomplete jumps, resulting in $\mathcal{O}(N \log T)$ incomplete jumps due to cluster $C$. Thus in total, the number of incomplete jumps can be bounded by $\mathcal{O}(N^2 \log T)$.

Now note that the small-gradient threshold is set to $T^{-1}$; since this threshold is positive, the algorithm may erroneously combine clusters. However, when this happens, the joint optimum for the two clusters must be at most $\mathcal{O}(T^{-1})$ distance away from the current point. Thus, adding the new constraint from combining these clusters, our new constrained optimum shifts by at most $\mathcal{O}(T^{-1})$ each time two clusters are combined. Thus, the constrained optimum shifts by at most $\mathcal{O}(N^2T^{-1})$ in total, meaning that the constrained optimal function value shifts by at most $\mathcal{O}(N^4T^{-2})$. The difference in regret between converging to the true constrained optimum and the shifted constrained optimum is therefore negligible in $T$.

Since we have exponential convergence at all but $\mathcal{O}(N^2 \log T)$ iterations, the total regret is $\mathcal{O}(N^2 \log T)$. \qed

\section{Analysis for Section ~\ref{sec:noisy}}
\label{app:noisy}

In this section, we provide analysis for our noisy algorithms (i.e., the algorithms from Section~\ref{sec:noisy}). In our noisy setting, function value observations are corrupted with additive independent sub-Gaussian noise. As such, we can bound the error in function value estimates using Hoeffding's inequality, presented below.

\begin{lemma}[general Hoeffding's inequality \cite{vershynin2018high}]
\label{lem:hoeffding}
Let $\varepsilon_1,\ldots,\varepsilon_n$ be independent mean-zero sub-Gaussian random variables. Then, there is a constant $C$ such that for every $s \geq 0$, we have
\[
\mathbb{P} \Big( \Big\vert \sum_{t=1}^n \varepsilon_i \Big\vert \geq s \Big) \leq 2 \exp \bigg( - \frac{Cs^2}{\sum_{t=1}^n \Vert \varepsilon_i \Vert_{\psi_2}^2} \bigg).
\]
\end{lemma}

\subsection{One dimensional algorithm}

Here, we analyze the single-group noisy algorithm (Algorithm~\ref{alg:continual-lgd-convex}). For the proof of Theorem \ref{prop:lgd-noisy-bandit-dynamic-lags}, we introduce some additional notation for ease of exposition. Following our notion of {\it phases}, we can uniquely associate to each $x_t$ a pair $(s,i)$ such that $x_t$ is the $s$th iterate in the $i$th phase; we denote such an iterate as $x_t = y_s^{(i)}$. For example, in Figure~\ref{fig:example-path}, we see that $x_1 = y_1^{(1)}, x_2 = y_1^{(2)}, x_3 = y_2^{(2)}, x_4 = y_1^{(3)}$ and $x_5 =y_1^{(4)}$. We now proceed with the proof.

{\it Proof of Theorem~\ref{prop:lgd-noisy-bandit-dynamic-lags}.} We break the proof into several claims. 

\begin{claim}[gradient accuracy]
\label{claim:gradient-accuracy}
Let $\frac{\overline{f}(x_t) - \overline{f}(x_t - \delta_{i})}{\delta_i}$ and $\frac{\overline{f}(x_t-\delta_{i+1}) - \overline{f}(x_t - \delta_{i})}{\xi\delta_i}$ be estimated secants at epoch $t$, where the former is the estimate at $x_t$ used to calculate a gradient-scaled jump, and the latter is an estimate at a lagged point. Then 
\begin{equation}
\label{eqn:gradient-bounds1-2-4-21}
\begin{tabular}{c}
$\left\vert \frac{\overline{f}(x_t) - \overline{f}(x_t - \delta_{i})}{\delta_i} - \frac{f(x_t) - f(x_t-\delta_i)}{\delta_i} \right\vert < \delta_i \quad\mbox{and}$ \\ \hspace{0.2cm}
$\left\vert \frac{\overline{f}(x_t-\delta_{i+1}) - \overline{f}(x_t - \delta_{i})}{\delta_i - \delta_{i+1}} -  \frac{f(x_t-\delta_{i+1})-f(x_t - \delta_i)}{\delta_i - \delta_{i+1}} \right\vert < \delta_i.$
\end{tabular}
\end{equation}
each with probability at least $(1-p)^2$, where $p = T^{-2}$.
\end{claim}

Consider the first bound. We sample at $x_t-\delta_i$ and $x_t$ on the order of $1/\delta_i^4$ times each. Thus, by Hoeffding's inequality,
\[
|\overline f(x_t) - f(x_t) | \in \mathcal{O}(\delta_i^2) ~~~\mbox{and}~~~ |\overline f(x_t-\delta_i) - f(x_t-\delta_i) | \in \mathcal{O}(\delta_i^2) ,
\]
each with probability at least $1-p$. Thus,
\begin{align*}
\left\vert \frac{\overline{f}(x_t) - \overline{f}(x_t - \delta_{i})}{\delta_i} - \frac{f(x_t) - f(x_t-\delta_i)}{\delta_i} \right\vert &= \left\vert \frac{\big[\overline{f}(x_t)- f(x_t)\big] - \big[\overline{f}(x_t - \delta_{i})  - f(x_t-\delta_i)\big]}{\delta_i} \right\vert \\
&\leq \frac{|\overline{f}(x_t)- f(x_t)| + |\overline{f}(x_t - \delta_{i})  - f(x_t-\delta_i)|}{\delta_i} \\
&\in \mathcal{O}(\delta_i).
\end{align*}
The proof for the second inequality is similar. $\hfill \qed_{\text{Claim \ref{claim:gradient-accuracy}}}$

\begin{claim}[overshooting]
\label{claim:overshooting}
Assume that (\ref{eqn:gradient-bounds1-2-4-21}) holds for all estimated gradients (this happens with probability at least $(1-p)^T$) and that $x_1 = x_{\min} + \delta_1 < x^*$, where $x^*$ is the unconstrained optimum. Then all the iterates $x_1,x_2,\ldots x_k$ (for $k\leq T$) generated by the lagged secant movements in the outer loop of the algorithm do not overshoot the optimum; that is, $x_t \leq x^*$ for all $t \leq k$. Note that this implies that all lagged 
points sampled by the algorithm (i.e., those sampled \emph{between} $x_t$ and $x_{t+1}$, for some $t$) also do not overshoot.
\end{claim}

{\sc Proof of Claim \ref{claim:overshooting}.} We show this by induction, where the base case follows from assumption that $x_1 \leq x^*$. Suppose $x_t \leq x^*$, and that $x_{t+1}$ was chosen based on a lag size of $\delta_i$. In other words, $x_{t+1} = x_t - \frac{1}{\beta}\tilde{\tilde{\nabla}}_t - \delta_i$, where $\tilde{\tilde{\nabla}}_t = \frac{\overline{f}(x_t) - \overline{f}(x_t - \delta_{i}) }{\delta_i} +  \delta_i$. By (\ref{eqn:gradient-bounds1-2-4-21}), we have that
\begin{equation} \label{eqn:ucb-higher-than-lagged}
    \nabla f(x_t-\delta_i)\leq \frac{f(x_t) - f(x_t-\delta_i)}{\delta_i} \leq  \tilde{\tilde{\nabla}}_t.
\end{equation}
So,
    \begin{align*}
        x_{t+1} - (x_t-\delta_i) &= - \frac{1}{\beta}\tilde{\tilde{\nabla}}_t  \\
        &\leq -\frac{1}{\beta}  \nabla f(x_t-\delta_i)  &\mbox{by (\ref{eqn:ucb-higher-than-lagged})} \\
        &= \frac{1}{\beta}  |\nabla f(x_t-\delta_i) | &\mbox{since $x_t \leq x^*$} \\
        &\leq x^* - (x_t-\delta_i) &\mbox{since $\nabla f(x^*) = 0$ and $f$ is smooth.} 
    \end{align*}

Thus, the next iterate $x_{t+1}$ does not overshoot the unconstrained optimum $x^*$.
$\hfill \qed_{\text{Claim \ref{claim:overshooting}}}$

\begin{claim}[monotonicity]
\label{claim:monotonicity}
Assuming (\ref{eqn:gradient-bounds1-2-4-21}) holds for all estimated gradients, and that $x_1 < x^*$, samples taken by the algorithm (including lagged and non-lagged iterates) are non-decreasing.
\end{claim}

{\sc Proof of Claim \ref{claim:monotonicity}.} Again, suppose that $x_{t+1}$ was chosen based on a lag size of $\delta_i$. In other words, $x_{t+1} = x_t - \frac{1}{\beta}\tilde{\tilde{\nabla}}_t - \delta_i$, where $\tilde{\tilde{\nabla}}_t = \frac{\overline{f}(x_t) - \overline{f}(x_t - \delta_{i})}{\delta_i} + \delta_i$ and $g_t^{(i)} = \frac{\overline{f}(x_t - \delta_{i+1}) - \overline{f}(x_t - \delta_{i})}{\xi\delta_i} + (1+\beta)\delta_i$, where $\delta_i=q^{i-1}\delta_1$ is the $i$th lag sizes and $\xi = 1-q$. Note that $\xi \delta_i = (x_t-\delta_{i+1}) - (x_t-\delta_i)$ is the domain gap between $x_t-\delta_{i}$ and $x_t-\delta_{i+1}$. To show monotonicity, we need to show that the next \emph{lagged} point exceeds the current point; i.e., we must show that $x_t \leq x_{t+1}-\delta_i$. 

First, note that by (\ref{eqn:gradient-bounds1-2-4-21}) and smoothness, we have that
\begin{equation} \label{eqn:gti-shallow}
    \nabla f(x_t) \leq g_t^{(i)}.
\end{equation}
So, since $-g_t^{(i)}/16\beta > \delta_i$, it follows that $-\nabla f(x_t)/(16\beta) > \delta_i$. So,

\begin{align*}
    x_{t+1} - x_t - \delta_i &= - \frac{1}{\beta}\tilde{\tilde{\nabla}}_t - 2\delta_i \\
    &\geq -\frac{1}{\beta} \Big[ \frac{f(x_t) - f(x_t - \delta_i)}{\delta_i} + 2\delta_i\Big]- 2\delta_i \\
    &\geq -\frac{1}{\beta} \Big[\nabla f(x_t)  + 2\delta_i\Big]- 2\delta_i \\
    &= -\frac{1}{\beta} \nabla f(x_t) - 2\delta_i(1 + 1/\beta) \\
    &> 16\delta_i - 2\delta_i(1 + 1/\beta) \\
    &= 14\delta_i - \frac{2\delta_i}{\beta} \\
    &\geq 12 \delta_i &\mbox{since $\beta \geq 1$} \\
    &> 0.
\end{align*}
This proves Claim \ref{claim:monotonicity}. $\hfill \qed_{\text{Claim \ref{claim:monotonicity}}}$

\begin{claim}[convergence rate]
\label{claim:convergence}
Let $h_t = f(x_t) - f(x^*)$ be the instantaneous regret at the $t$th non-lagged iterate. Then $h_t \in \mathcal{O}(e^{-ct})$ for some $c > 0$.
\end{claim}

Let $G = -\beta(x_{t+1} - x_t) =  \frac{\overline{f}(x_t) - \overline{f}(x_t-\delta_i) }{\delta_i} + (1+\beta)\delta_i$. Thus, $x_{t+1} = x_t - \frac{1}{\beta} G$. This rescaling will be useful in the proof of this claim. Before proceeding with the proof of this claim, we make three observations.

First, an immediate consequence of (\ref{eqn:ucb-higher-than-lagged}) is that 
\begin{equation} \label{eqn:smooth-claim4-G-underestimate}
    \nabla f(x_t) \leq \nabla f(x_t - \delta_i) + \beta \delta_i \leq \tilde{\tilde{\nabla}}_t + \beta \delta_i = G.
\end{equation}
In other words, $G$ is an underestimate (in magnitude) of $\nabla_t$.

Second, suppose that we jump from $x_t$ to $x_{t+1}$ with lag size $\delta_i$, so that $-\frac{1}{16\beta}g_t^{(i)} > \delta_i$. Note that the lagged gradient estimate $g_t^{(i)}$ is an underestimate (in magnitude) of $\nabla_t$. Thus, the following inequalities hold:
\begin{equation} \label{eqn:smooth-claim4-delta-small}
    -\frac{1}{4\beta} \nabla_t > -\frac{1}{16\beta} \nabla_t > \delta_i. 
\end{equation}

Third, in order to show fast convergence, we show that $G$ is large in magnitude. To that end, observe that
\begin{align}
    G &= \frac{\overline{f}(x_t) - \overline{f}(x_t-\delta_i) }{\delta_i} + (1+\beta)\delta_i \\
    &\leq \frac{f(x_t) -f(x_t-\delta_i) }{\delta_i} + (2+\beta)\delta_i \\
    &< \nabla_t + (2+\beta)\delta_i \\
    &< \nabla_t - \frac{2+\beta}{4\beta} \nabla_t &\mbox{by (\ref{eqn:smooth-claim4-delta-small})} \\
    &= \Big( 1 - \frac{2+\beta}{4\beta} \Big) \nabla_t \\
    &\leq \nabla_t / 4 &\mbox{since $\beta \geq 1$.} \label{eqn:G-is-large}
\end{align}

With these bounds in mind, we proceed to bound the convergence rate. Letting $h_t = f(x_t) - f(x^*)$,
\begin{align*}
    h_{t+1} - h_t &= f(x_{t+1}) - f(x_t) \\
    &\leq \nabla_t (x_{t+1}-x_t) + \frac{\beta}{2} (x_{t+1}-x_t)^2 &\mbox{smoothness}\\
    &= -\frac{1}{\beta} \nabla_t G + \frac{1}{2\beta} G^2 \\
    &\leq -\frac{1}{\beta} G^2 + \frac{1}{2\beta} G^2 &\mbox{by (\ref{eqn:smooth-claim4-G-underestimate})} \\
    &= -\frac{1}{2\beta} G^2.
\end{align*}

So, by (\ref{eqn:monotonic-improvements}) and the above bound on $h_{t+1}-h_t$,
\begin{equation*} 
    h_{t+1}-h_t \leq - \frac{1}{2\beta} \Vert G\Vert^2 \overset{(*)}{\leq} -\frac{1}{32\beta} \Vert \nabla f(x_t) \Vert^2 \overset{\text{PL}}{\leq} -c h_t ~.
\end{equation*}
where $(*)$ follows from (\ref{eqn:G-is-large}) and $c = \frac{\alpha}{32\beta}$. Letting $w_t = \frac{c}{\Vert x_t - x^* \Vert^2}$, we have that $w_th_t^2 + h_{t+1} \leq h_t$. This is equivalent to $h_{t+1} \leq (1-c)h_t$, which implies an exponential convergence rate of $h_{t+1} \leq h_1 e^{-ct}$. \qed 

Finally, we can put the previous claims together to bound the regret. We will consider the cases where $|\nabla f(x_t)| \geq T^{-1/4}$ and $|\nabla f(x_t)| < T^{-1/4}$ separately. Letting $T'$ be the number of non-lagged iterates until the gradient is less than $T^{-1/4}$ in magnitude, one can show that $T' \in \mathcal{O}(\log T)$ due to the exponential convergence rate. Thus, ignoring logarithmic factors, we have
\begin{align*}
    \operatorname{regret}_T &= \sum_{t=1}^{T'} \frac{h_t}{\nabla f(x_t)^4} + \sum_{\substack{\text{remaining} \\  \text{pts $x$}}} \big(f(x) - f(x^*) \big) \\
    &\leq \sum_{t=1}^{T'} \frac{1}{\nabla f(x_t)^2} + \sum_{\substack{\text{remaining} \\  \text{pts $x$}}} \nabla f(x)^2  &\mbox{by the PL condition} \\
    &\leq T^{1/2}T' + T^{-1/2}T \\
    &\leq T^{1/2}. \qed 
\end{align*}

\subsection{Two-dimensional algorithm}

Algorithm \ref{alg:noisy-2segment-asymmetric-simplified-v2-long} (i.e., the more detailed description of Algorithm~\ref{alg:noisy-2segment-asymmetric-simplified-v2}) is composed of two phases: (1) a coordinate descent phase, in which {\sc Ada-LGD} (Alg.~\ref{alg:continual-lgd-convex}) is run separately on the two groups, and the group being optimized changes each time a lag size is contracted; and (2) a combined phase, where the decisions for one group are locked with respect to the decisions for the other, and the two are optimized simultaneously. For simplicity, we will assume that at each queried point $(x_1,x_2)$, feedback is observed and regret is incurred for both groups. Without this assumption, the regret will increase by a $\tilde{\mathcal{O}}(1)$ factor dependent on $\min_{i \in \{1,2\}} a_i$, where $a$ is the distribution over groups. Much of the analysis is identical to that of Algorithm \ref{alg:continual-lgd-convex}; as such, we focus on the differences, and break up the argument into claims.

\begin{algorithm}[!t]
\scriptsize
\footnotesize
\DontPrintSemicolon
\SetKwInOut{Input}{input}
\Input{initial point $(x_1,x_2)$, smoothness $\beta \geq 1$ of the objective function $f(x_1,x_2) = f_1(x_1) + f_2(x_2)$, horizon $T$, slacks $s(1,2), s(2,1)$, $x_{\min}$, $\gamma$, $q \in (0,1)$, $\xi = 1-q$, 
initial lag $\delta = \delta^{(1)} = \delta^{(2)}$,  
$\textsc{Grad}(x,y)= \frac{\overline{f}(y) - \overline{f}(x)}{y-x}$, where $\overline{f}$ is the average of $\ssize(y-x)$ function value samples, where $\ssize(d) = \frac{2 E_{\max}^2 \log \frac{2}{p}}{d^4}$, and gradient adjustments $\varepsilon_1(x) = (1+\beta)x$ and $\varepsilon_2(x) = x$}
Initialize queues to maintain tuples of (sampling point, number of samples, and the type of sampling point $(S)$):
$Q_1 = Q_2 = \{\big(x_{\min}, \ssize(\xi \delta), S=1\big)$, $\big(x_{\min}+\xi\delta, \ssize(\xi q\delta), S=2\big)$, $\big(x_{\min}+\delta, \ssize(\delta),S=0\big)$\}\; 
\tcp*{$S=k>0$ represents the $k$th lagged point, $S=0$ represents a non-lagged point, and $S\in\{-1,-2\}$ represents the two feasibility iterates.}
\textbf{Coordinate Descent Phase:}\; 
\While{fewer than $T$ samples have been taken}{

    $\mathcal{X}_i = [x_i, x_{-i}+s(i,-i)]$ for $i=1,2$ \tcp*[f]{current feasible points for Group $i= 1,2$}\;
    \textbf{if} $\exists j: Q_j \cap \mathcal{X}_j \neq \emptyset$ \textbf{then} Set $i \in \{j ~|~ Q_j \cap \mathcal{X}_j \ne \emptyset\}$; 
    {\bf else} go to {\bf Combined Phase} (line 11)\;
    set \texttt{switch} = False\;
    \While{$Q_i \cap \mathcal{X}_i \ne\emptyset$}{ 
        {\bf Sample:} Let $(x_i, n_x, S_i)$ be the lowest point to sample in $Q_i \cap \mathcal{X}_i$ by first coordinate (with ties broken arbitrarily) and obtain $n_x$ 
        samples \js{of $f_j(x_j)$ for $j\in \{1,2\}$ (in particular, $f_i(x_i)$)}; Update $Q_i \leftarrow Q_i \setminus (x_i, n_x, S_i)$\;
        \textbf{Gradient checks based on type ($S_i$) of $x_i$}:\;
        \begin{enumerate}
            \item[(a)] \textbf{if} $x_i$ is a $j$th lagged point (i.e., $S_i=j$) and $j>1$ \textbf{then} \;
                \hspace{1.3cm} \textbf{if} $-\textsc{Grad}((j-1)\text{st}~\mbox{lagged point},x_i) - \varepsilon_1(\delta^{(i)}) \leq \gamma \delta^{(i)}$:\;
                \hspace{2cm} Plan to sample next lagged point: $Q_i \leftarrow Q_i \cup$ $\{\big(x_i+\xi q \delta^{(i)}, \ssize(\xi q \delta^{(i)}), S=j+1\big)\}$ \;
                \hspace{2cm} $\delta^{(i)} \leftarrow q\delta^{(i)}$ and adjust the sample size for the next non-lagged iterate in $Q_i$ to $\ssize(\delta^{(i)})$\; 
            \item[(b)] \textbf{if} $x_i$ is a non-lagged point (i.e., $S_i = 0$) \textbf{then} \;
                \hspace{1.3cm} Let $g = \textsc{Grad}(x_i-\delta^{(i)},x_i) + \varepsilon_2(\delta^{(i)})$\;
                \hspace{1.3cm} \textbf{if} $-\frac{1}{\beta} g < T^{-1/4}$  \textbf{then} set $Q_i \leftarrow \emptyset$ (never move group $i$ in coordinate descent again)\;
                \hspace{1.3cm} \textbf{else} 
                populate queue with the next non-lagged and lagged points:\; 
                \hspace{2cm} Let $y = x_i - \delta^{(i)} - \frac{1}{\beta} g$, and\;
                \hspace{2cm} $Q_i \leftarrow Q_i \cup \{(y-\delta^{(i)},\ssize(\xi \delta^{(i)}), S=1), ~(y-q\delta^{(i)},\ssize(\xi q \delta^{(i)}), S=2)$, $(y,\ssize(\delta^{(i)}), S=0)$\}\;
                \hspace{1.3cm} \textbf{if} lagged size has dropped enough so that $\delta^{(i)} < \delta^{(-i)}$, then switch group: \texttt{switch} = True\;
            \item[(c)] \textbf{if} $x_i$ is the second feasibility iterate (i.e., $S_i = -2$) \textbf{then} \;
                \hspace{1.3cm} Let $g = \textsc{Grad}(\mbox{previous feasibility iterate},x_i)$\;
                \hspace{1.3cm} \textbf{if} $-\frac{1}{\beta} g \geq  \Big(\frac{(2+\gamma)\beta}{q \alpha}+1\Big) \delta^{(-i)}$ \textbf{then} \;
                \hspace{2cm} set $Q_1,Q_2 \leftarrow \emptyset$ (i.e., enter the combined phase) \;
                \hspace{1.3cm} \textbf{else if} $Q_{-i} = \emptyset$ \textbf{then} sample at $(x_1,x_2)$ for the remaining time until $T$\;
        \end{enumerate} 
        {\bf Add feasibility iterates:} \textbf{if} 
        (\texttt{switch} = True or $Q_i \cap \mathcal{X}_i = \emptyset$) AND $\delta^{(i)} < \delta^{(-i)}$  
        \textbf{then}\; 
        \hspace{1.3cm}set $\mathcal{X}_{-i} \leftarrow [x_{-i}, x_i + s(-i,i)]$\;
        \hspace{1.3cm}$Q_{-i} \leftarrow Q_{-i} \cup \{\big(\max \mathcal{X}_{-i} - \delta^{(i)}, \ssize(\xi \delta^{(i)}), S=-1\big), \big(\max \mathcal{X}_{-i} - q\delta^{(i)}, \ssize(\xi q\delta^{(i)}), S = -2\big)\}$\;
        \hspace{1.3cm}switch groups: $i \leftarrow -i$ 
        }
    }
{\bf Combined phase:}\\
\While{fewer than $T$ samples have been taken}{
    Run \textsc{Ada-Lgd} with previously sampled lagged points (with lags $\delta^{(-i)}$ and $q\delta^{(-i)}$) on function $\psi$: \\
        \hspace{1cm}\textbf{if} Group 2 is tight, set $\psi(x)=f_1(x)+f_2(x+s(2,1))$; \textbf{else} set $\psi(x)=f_1(x)+f_2(x-s(1,2))$
}
\caption{Switch-then-Combine Adaptive Lagged Gradient Descent ({\sc SCAda-Lgd})}
\label{alg:noisy-2segment-asymmetric-simplified-v2-long}
\end{algorithm}

\begin{figure}
\begin{center}
\scalebox{.7}{
\begin{tikzpicture}
\def\sone{.15}
\def\stwo{.15}
\def\done{2}
\def\q{.5}

\draw[white, fill=gray, fill opacity=.1] (-6.5,3) -- (-5,3) -- (-5,2.5) -- (-6.5,2.5) -- cycle;
\node at (-3.5,2.75) {Feasible region};
\draw[white, fill=gray, fill opacity=.1] (-4,-2.9) -- (2,3.1) -- (4,1.1) -- (-2,-4.9) -- cycle;
\draw (-4,-2.9) -- (2,3.1);
\draw (-2,-4.9) -- (4,1.1);

\tikzstyle{vertex}=[draw, circle, color = black, fill = black, text opacity = 1, inner sep = 1.4pt]


\draw[dashed] (-1.4,-4.3) -- (-1.4,-1.4);
\draw[dashed] (-1.4,-1.4) -- (1.5,-1.4);
\draw[dashed] (-2,-4.3) -- (-1.4,-4.3);

\node[vertex] (z2) at (-1.4,-1.4) {};

\foreach \a in {0,...,2}{
    \node[cyan] at (-1.4, -1.4-\done*\q^\a) {\small $\times$};
}
\node[vertex,cyan,inner sep=2pt](v) at (-1.4,-1.4) {};

\foreach \a in {1,...,2}{
    \node[purple] at (1.5-\done*\q^\a, -1.4) {\small $\circ$};
}

\draw[dotted,orange,fill=orange,fill opacity=.1] (1.5,-1.4) -- (1.5,.5) -- (3.4,.5) -- (3.4,-1.4) -- cycle;

\node[orange] at (4.2,-.45) {$D\big(\delta^{(2)}\big)$};

\node at (2.3,-3.1) {$(x_1,x_2)$};
\begin{scope}[every path/.style={-{Latex[length=2mm]},line width=.2pt}, every node/.style={inner sep=1pt}]
\path (2.3,-2.9) edge[bend right=20] (1.5,-1.4);
\end{scope}


\draw[gray] (.5,-.2) -- (.7,-.2);
\draw[gray] (1.3,-.2) -- (1.5,-.2);
\draw[gray] (.5,-.3) -- (.5,-.1);
\draw[gray] (1.5,-.3) -- (1.5,-.1);
\node[gray] at (1,-.2) {\small $\delta^{(2)}$};

\draw[gray] (-2.35,-1.4) -- (-2.35,-1.6);
\draw[gray] (-2.35,-2.2) -- (-2.35,-2.4);
\draw[gray] (-2.45,-1.4) -- (-2.25,-1.4);
\draw[gray] (-2.45,-2.4) -- (-2.25,-2.4);
\node[gray] at (-2.35,-1.9) {\small $\delta^{(2)}$};

\draw [purple,decorate,decoration={brace,amplitude=5pt,mirror},xshift=0pt,yshift=6pt] (1.5-\done*\q^2,-1.4) -- (1.5-\done*\q,-1.4) node [purple,midway,xshift=0pt,yshift=12pt] {$g_1$};

\draw [cyan,decorate,decoration={brace,amplitude=5pt},xshift=6pt,yshift=0pt] (-1.8,-1.4-\done*\q) -- (-1.8,-1.4-\done*\q^2) node [cyan,midway,xshift=-13pt,yshift=0pt] {$g_3$};

\draw [cyan,decorate,decoration={brace,amplitude=5pt},xshift=6pt,yshift=0pt] (-1.8,-1.4-\done) -- (-1.8,-1.4-\done*\q) node [cyan,midway,xshift=-13pt,yshift=0pt] {$g_4$};

\draw [cyan,decorate,decoration={brace,amplitude=5pt,mirror},xshift=17pt,yshift=0pt] (-1.8,-1.4-\done*\q) -- (-1.8,-1.4) node [cyan,midway,xshift=13pt,yshift=0pt] {$g_2$};


\end{tikzpicture}
}
\end{center}
\caption{\label{fig:combined-phase-avoided-successfully-v2}
Decision space $\mathcal{X}_F^N$ with the following scenario in Alg. \ref{alg:noisy-2segment-asymmetric-simplified-v2}: Group 2 has undergone at least one lag size transition and Group 1 is at a tight constraint. X's represent lagged points, filled-in circles the non-lagged points, and empty circles the feasibility iterates; gradients of $f_1$ are estimated at red nodes, and gradients of $f_2$ are estimated at blue nodes. $g_1,\ldots,g_4$ denote gradient estimates obtained from the two nodes indicated by the brackets. The orange square of side length $D\big(\delta^{(2)}\big)$ is relevant to Claims~\ref{claim:transitioning-to-combined}-\ref{claim:combined-phase-monotonicity}. }
\end{figure}

\begin{claim}
\label{claim:transitioning-to-combined}
Suppose $\delta^{(2)} < \delta$ is the lag size for Group 2, and Group 1 has hit the boundary of the feasible region (see Fig. \ref{fig:combined-phase-avoided-successfully-v2}) at $(x^{(1)},x^{(2)})$. Then the following hold with probability at least $(1-p)^4$:
\begin{itemize}
    \item if $x^*_1- x_1\leq D(\delta^{(2)})$, then the algorithm does not enter the combined phase; and
    \item if $x^*_1- x_1\geq D(\delta^{(2)}) + H(\delta^{(2)})$, then the algorithm enters the combined phase.
\end{itemize}
\end{claim}

{\it Proof of Claim \ref{claim:transitioning-to-combined}.} First suppose that $x^*_1- x_1\leq D(\delta^{(2)})$. Then by smoothness, we have that $|f_1'(x_1)| \leq \beta D(\delta^{(2)}) = (2+\gamma)\frac{\beta^2}{\alpha} \cdot \frac{\delta^{(2)}}{q}$. Now let $g_1$ be the gradient estimate obtained from points $x_1- \delta^{(2)}$ and $x_1- q\delta^{(2)}$, as shown in Figure \ref{fig:combined-phase-avoided-successfully-v2}. Then
\begin{align*}
    -\frac{1}{\beta} g_1 &= \frac{1}{\beta}|g_1| \leq \frac{1}{\beta}|f_1'(x_1- \delta^{(2)})| \leq \frac{1}{\beta}|f_1'(x_1)| + \delta \leq D(\delta^{(2)}) + \delta^{(2)} = (2+\gamma) \frac{\beta}{\alpha} \cdot \frac{\delta^{(2)}}{q} + \delta^{(2)}
\end{align*}
with probability at least $(1-p)^2$. In this case, the algorithm does not enter the combined phase.

Now suppose that $x^*_1- x_1\geq D(\delta^{(2)}) + H(\delta^{(2)})$. Expanding this expression, we get that $x^*_1- x_1\geq (2+\gamma) \frac{\beta^2}{\alpha^2} \cdot \frac{\delta^{(2)}}{q} + \frac{\beta \delta^{(2)}}{\alpha}$. This implies by the PL inequality (in particular, by the error bound), that
\[
|f_1'(x)| \geq (2+\gamma) \frac{\beta^2}{\alpha} \cdot \frac{\delta^{(2)}}{q} + \beta \delta^{(2)}.
\]
Since $|g_1| > |f_1'(x)|$ with probability at least $(1-p)^2$, it follows that $-\frac{1}{\beta} g_1 \geq (2+\gamma) \frac{\beta}{\alpha} \cdot \frac{\delta^{(2)}}{q} + \delta^{(2)}$ with the same probability. Hence, in this case, the algorithm will enter the combined phase. $\hfill \qed_{\text{Claim \ref{claim:transitioning-to-combined}}}$ 

\begin{claim}
\label{claim:combined-phase-monotonicity}
Let $g_1$ be the gradient estimated at the feasibility iterates (see Figure~\ref{fig:combined-phase-avoided-successfully-v2}), and let $\x^*=(x_1^*,x_2^*)$ be the unconstrained optimum.
\begin{itemize}
    \item If $\x^* \in \text{EF}$, then the algorithm never enters the combined phase.
    \item If the algorithm enters the combined phase based on the gradient estimate $g_1$ (as in Fig. \ref{fig:combined-phase-avoided-successfully-v2}), then $-\frac{1}{\beta} g_1 \geq (2+\gamma)\delta^{(2)}$.
\end{itemize}
\end{claim}

{\it Proof of Claim \ref{claim:combined-phase-monotonicity}.} We begin with the first claim. It is enough to show that Group 1 sampling will not induce a transition to the combined phase, given that Group 2 has already undergone at least one lag size transition. As in Claim \ref{claim:combined-phase-monotonicity}, let $\delta^{(2)} < \delta$ be the current lag size of Group 2. We will argue that the Group 2 gradient must be small, since this group has transitioned from the previous lag size $\delta^{(2)}/q$; thus, the optimum $x_2^{*}$ of $f_2$ must be close to the current point $x_2$. 

To that end, let $g_4$ be the gradient which induced the transition to $\delta^{(2)}$ (see, e.g., Fig. \ref{fig:combined-phase-avoided-successfully-v2}). It follows that $-\frac{1}{\beta} g_4 < (2+\gamma) \frac{\delta^{(2)}}{q}$. Let $y$ be the non-lagged iterate corresponding to $g_4$ (so, in Fig. \ref{fig:combined-phase-avoided-successfully-v2}, $y = x_2$). Then $\alpha |x^*_2- y| \leq |f_2'(y)| < |g_4| < (2+\gamma) \beta \delta^{(2)}/q$. It follows that 
\begin{equation}
    |x^*_2- x_2| \leq |x^*_2- y| < (2+\gamma) \frac{\beta}{\alpha} \cdot \frac{\delta^{(2)}}{q} = D\big(\delta^{(2)}\big).
\end{equation}
Since we are assuming that $\x^* \in \text{EF}$, it must be that $|x^*_1- x_1| \leq D\big(\delta^{(2)}\big)$ as well. Thus, by Claim \ref{claim:transitioning-to-combined}, the algorithm will not enter the combined phase with high probability.

To prove the second statement, let $g_1$ be the gradient estimate (in the Group 1 direction) which induced the transition to the combined phase (see, e.g., Fig. \ref{fig:combined-phase-avoided-successfully-v2}). Then, since $\alpha |x-x^*| \leq |\nabla f_i(x)| \leq \beta |x-x^*|$ for $x$, it follows that $\alpha \leq \beta$. So, since $0 < q < 1$, we have
\[
-\frac{1}{\beta} g_1 \geq \left(\frac{(2+\gamma)\beta}{q\alpha} + 1\right) \delta^{(2)} > (2+\gamma)\delta.
\]
In other terms, since the algorithm transitions to the combined phase \emph{only when} the gradient in the Group 1 direction is large enough, a lag size transition will not be triggered. $\hfill \qed_{\text{Claim \ref{claim:combined-phase-monotonicity}}}$ 

\begin{claim}
\label{claim:checking-feasibility check-is-fast}
The feasibility check iterates (e.g., the brown nodes in Figure \ref{fig:combined-phase-avoided-successfully}) increase the overall regret during the coordinate phase by at most a factor of two.
\end{claim}

{\it Proof of Claim \ref{claim:checking-feasibility check-is-fast}.} Feasibility check iterates are sampled $\tilde{\mathcal{O}}(1/\delta_i^4)$ times, where $\delta_i$ is the current lag size of the other group. Thus the regret incurred at the two feasibility check iterates is strictly less than the regret incurred at the previous two lagged iterates in the other group. $\hfill \qed_{\text{Claim \ref{claim:checking-feasibility check-is-fast}}}$ 

\begin{claim}
\label{claim:regularity-of-transitions}
Let $\delta_{n_1}> \cdots > \delta_{n_m}$ be the distinct non-trivial lag sizes of Group $j$ (excluding $\delta$ if $\delta$ is a non-trivial lag size), where $\delta_i = q^i \delta$. Defining $n_0 = 0$, we have that $n_{i+1}-n_i \in \mathcal{O}(1)$ for $0 \leq i < m$. Moreover, the number of gradient-scaled jumps taken with lag size $\delta_{n_i}$ is constant as well, for $1 \leq i \leq m$.
\end{claim}

{\it Proof of Claim \ref{claim:regularity-of-transitions}.} 
For ease of exposition, we drop all constants in the proof of this claim. First, observe that with high probability, we jump from the lagged point $x_t-\delta_{n_i}$ by an amount proportional to $|\nabla f(x_t)|$. Thus, with high probability, the gradient increases by $\mathcal{O}(|\nabla f(x_t)|)$ from $x_t$ to the next non-lagged point $x_{t+1}$. In other words, with high probability,
\[
\frac{\nabla f(x_{t+1})}{\nabla f(x_t)} \geq c' ~~\mbox{for some constant $c' > 0$.}
\]
Thus, since $\delta_{n_i} \approx |\nabla f(x_t)|$ and the lag sizes $\delta_{n_i}$ decay exponentially in $i$, this proves that $n_{i+1} - n_i \in \mathcal{O}(1)$.

The above argument can be extended to bound $n_1$ as well. If the initial lag size of $\delta$ was non-trivial (i.e., some gradient-scaled jump was made using a lag size of $\delta$), then the above argument holds verbatim. Otherwise, the initial non-lagged iterate $(x_{\min} + \delta)$ has a gradient of magnitude $\mathcal{O}(\delta)$. However, since $\delta = 1/\log T$ and the optimum is assumed to be strictly greater than zero, we may assume that $T$ is large enough so that the gradient at $x_{\min} + \delta$ is large with respect to $\delta$.

Finally, we argue the ``moreover'' claim. While $\delta_{n_i}$ is the lag size, the gradient has magnitude $\Theta(\delta_{n_i})$. Once the gradient drops below $\mathcal{O}(q\delta_{n_i})$, a lag transition is initiated. Since the gradient-scaled jumps yield exponential convergence, $t$ jumps with lag size $\delta_{n_i}$ would yield instantaneous regret of order $e^{-t} \delta_{n_i}^2$. For $t \in \mathcal{O}(1)$, this would result in an instantaneous regret of order $(q\delta_{n_i})^2$, which would trigger a lag size transition. $\hfill \qed_{\text{Claim \ref{claim:regularity-of-transitions}}}$ 

\begin{claim}
\label{claim:waiting-regret-convex}
The waiting regret (i.e., the regret incurred by Group 1 over the blue nodes and by Group 2 over the red nodes in Figure \ref{fig:combined-phase-avoided-successfully}) is $\tilde{\mathcal{O}}(T^{1/2})$.
\end{claim}

{\it Proof of Claim \ref{claim:waiting-regret-convex}.} Note that smaller lag sizes require more sampling: if $\delta^{(2)} < \delta^{(1)}$, then more time will be spend sampling $f_2$ at $x_2 - \delta^{(2)}$ and $x_2 - q\delta^{(2)}$ than will be spent sampling $f_1$ at $x_1 - \delta^{(1)}$ and $x_1 - q\delta^{(1)}$. In order to capture worst-case waiting regret for a group, we will consider the case where $\delta^{(2)} \leq \delta^{(1)}$ at all times and bound the waiting regret of Group 1.

Let $\delta_{n_1} > \cdots > \delta_{n_m}$ be the non-trivial lag sizes used by Group 2 (excluding $\delta$, if $\delta$ is a non-trivial lag size). Note that $\delta_{n_1}$ is not necessarily $\delta$, and $\delta_{n_{i+1}}/\delta_{n_i}$ is not necessarily $q$. For each $i$, let $T_i$ denote the number of samples taken where $\delta_{n_i}$ is the lag size \emph{or} in transitioning to $\delta_{n_{i+1}}$. Thus, letting $T_0$ denote the number of samples prior to $\delta_{n_1}$, we have that $\sum_{i=0}^m T_i = T$.

We will now bound the waiting regret of Group 1 during these $m+1$ phases. First, consider the time period associated with $T_i$, for any $1 \leq i \leq m$. During this phase, Group 2 has already transition to a lag size of $\delta_{n_i}$ from a lag size of $\delta_{n_i}/q$. Note that once this transition is made, the algorithm switches to optimizing over Group 1. Since the algorithm did not enter the combined phase in the previous round, it must be that the gradient in the Group 1 direction is of order $\delta_{n_i}$. Thus, the total waiting regret for Group 1 is of order
\[
T_0 + \sum_{i=1}^m T_i \delta_{n_i}^2 ~.
\]

Now let us consider bounds on $T_i$, for $0 \leq i \leq m$. By Claim \ref{claim:regularity-of-transitions}, we know that the number of gradient-scaled jumps that Group 2 takes during this phase is constant, and similarly the number of lags transitions (i.e., $n_{i+1}-n_i$) is also constant. It follows that $T_i \in \tilde{\mathcal{O}}(\delta_{n_i}^{-4})$. 

We can similarly bound $T_0$. As in the previous calculations, we will ignore constants for simplicity. To bound $T_0$, we first bound the number of gradient-scaled jumps before the current gradient is of order $q\delta$ (as this would trigger a lag size transition). To that end, the linear convergence rate of {\sc Ada-LGD} (Alg.~\ref{alg:continual-lgd-convex}) implies that the gradient after $t$ gradient-scaled jumps is of order $e^{-t} M^2$, where $M = \max_{(x,y) \in \text{EF}} |f_2'(y)|$. It follows that the gradient is of order $q\delta$ after $\ln \frac{M}{q\delta}$ jumps; since $\delta = 1/\log T$, the total number of gradient-scaled jumps before the first lag size transition is of order $\log \log T$. Finally, by Claim \ref{claim:regularity-of-transitions}, $n_1$ is constant as well, which implies that $T_0 \in \mathcal{O}\big((\log T)^5 \log \log T\big)$. 

Putting all of this together, the waiting regret for Group 1 is of soft order $\sum_{i=1}^m \delta_{n_i}^{-2}$. Letting $x_{2,t_i}$ denote the first non-lagged iterates for Group 2 to admit a lag size of $\delta_{n_i}$, we have that $|\nabla f_2(x_{2,t_i})| \in \Theta(\delta_{n_i})$. Thus we can rewrite the waiting regret for Group 1 as $\sum_{i=1}^m 1/|\nabla f_2(x_{2,t_i})|^2$. Finally, since we stop optimizing over Group 2 when its gradient is bounded in magnitude by $T^{-1/4}$, we can bound this further by $m T^{1/2}$. Since there can be at most logarithmically many gradient-scaled jumps before the gradient is of order $T^{-1/4}$, the waiting regret is $\tilde{\mathcal{O}}(T^{1/2})$. $\hfill \qed_{\text{Claim \ref{claim:waiting-regret-convex}}}$ 

Since the combined function $h$ is $\alpha$-strongly convex and $\beta$-smooth, we have the same convergence rate for the combined phase as before. Moreover, since the algorithm does not erroneously enter the combined phase with high probability, the optimum of $h$ is the same as the constrained optimum with high probability. Thus, the regret analysis of Algorithm \ref{alg:continual-lgd-convex} carries over. \qed

\subsection{\texorpdfstring{$N$}{N}-group algorithm}
\label{app:n-group-noisy}

\begin{algorithm}[h]
\footnotesize
\DontPrintSemicolon
\SetKwInOut{Input}{input}
\Input{Initial point $(\x(1),\ldots,\x(N))$, initial lags $(\delta_1,\ldots,\delta_N)$, sub-Gaussian norm bound $E_{\max}$, $\ssize(d) = \frac{2 E_{\max}^2 \log \frac{2}{p}}{d^4}$, smoothness parameter $\beta \geq 1$ of $f(\x(1),\ldots,\x(N)) = \sum_{i=1}^N f_i(\x(i))$, time horizon $T$, non-negative EF slacks $s$, $x_{\min}$, $\xi = 1-q$, $\gamma$, and gradient adjustments $\varepsilon_1(x) = (1+\beta)x$ and $\varepsilon_2(x) = x$} 
Initialize queue $Q_i \leftarrow \emptyset$ for $i \in [N]$ \tcp*[f]{will contain points to query} \; 
Initialize the partition $\Pi = \big\{ \{1\},\ldots,\{N\} \big\}$ of groups, and set $\textup{active}_A = \textup{Yes}$ and $\psi'_A = -\infty$ for each $A \in \Pi$ \;
For any $A \subseteq [N]$, define $\psi_A(x)$ as in Algorithm~\ref{alg:multi-segment-noiseless}
; initialize $A = \{1\}$ \;
Set $\x(i) \leftarrow x_{\min}$ for $i \in [N]$, and add $\x(i)$, $\x(i) + \xi \delta_i$, $\x(i)+\delta_i$ to $Q_i$ for $i \in [N]$ \;
\While{fewer than $T$ samples have been taken}{
    \If{$\textup{active}_B = \textup{No}$ for each cluster $B \in \Pi$ (i.e., no clusters can move)}{
        \uIf{each cluster is constrained by some other cluster}{
            Find clusters $C,D$ which are locked in place with respect to each other (see Claim~\ref{claim:existence-of-low-gradient})
        }
        \Else{
            Find clusters $C \ne D \in \Pi$ such that $C$ is at a tight constraint imposed by some group in $D$, and $D$ has met the small gradient condition ($|\psi'_D| < T^{-1/4}$) (see Claim~\ref{claim:existence-of-low-gradient}) \;
        }
        Update $\Pi \leftarrow \Pi \cup \{C \cup D\} \setminus \{C,D\}$ \;
        Let $A = C \cup D$, and set $\psi'_{A} = -\infty$ and $\textup{active}_A = \textup{Yes}$\;
        Let $i = \min A$, set $\delta_i$ to be the smaller of the two lag sizes, and add $\x(i), \x(i) + \xi \delta_i, \x(i)+\delta_i$ to $Q_i$ \tcp*[f]{add points to the representative group's queue}
    }
    Let $i = \min A$ \tcp*[f]{choose a representative group} \;
    $\mathcal{X}_A = [\x(i), \x(i) + d(A;x)]$ \tcp*[f]{current feasible points for cluster $A$}\;
    Add the appropriate feasibility check iterates \;
    \While{$Q_i$ contains any elements in $\mathcal{X}_A$ and no non-lagged points have been sampled}{
        Let $\x(i) \leftarrow \min Q_i$, remove $\x(i)$ from $Q_i$, and update $\x(j)$ (for $j \in A \setminus \{i\}$) accordingly (i.e., letting $d = \min Q_i - \x(i)$, we set $\x(j) \leftarrow \x(j) + d$ for all $j \in A$)\; 
        Update $\textup{active}_B$ for $B \in \Pi$ as necessary (formerly tight clusters may now be active; in particular, if $B$ is at a tight constraint imposed by some group in $A$ and $B$ is not at any tight constraints imposed by any group in any other cluster, then $B$ will no longer be at a tight constraint) \;
        Sample $\ssize(\xi\delta)$ times at $\x(j)$ for $j \in A$ \;
        \uIf{$\x(i)$ is a lagged iterate for cluster $A$, but not the first lagged iterate}{
            \textbf{if} the estimated gradient is less than $\gamma\delta_i$ in magnitude, set $\delta_i \leftarrow q\delta_i$ and add $\x(i)+\xi \delta_i$ to $Q_i$ 
        }
        \ElseIf{$x_i$ is a non-lagged iterate for cluster $A$}{
            Update the gradients $g^{(i)}$ for $i \in A$  and set $\psi'_A = \sum_{i \in A} g^{(i)} + \varepsilon_2(\delta_i)$ \;
            Add $\x(i) - 2\delta_i - \frac{1}{\beta}\psi'_A$, $\x(i) - (1+q)\delta_i - \frac{1}{\beta}\psi'_A$, and $\x(i) - \delta_i - \frac{1}{\beta}\psi'_A$ to $Q_i$ \;
        }
    }
    \If{$Q_i$ contains no elements in $\mathcal{X}_A$}{
        Set $\x(i) = \max \mathcal{X}_A$ and update $\x(j)$ (for $j \in A$) accordingly \;
        Set $\textup{active}_A = \textup{No}$ and update $\textup{active}_B$ for $B \in \Pi$ (formerly tight clusters may now be active) 
    }

    \If{$|\psi'_B| < \beta (1+\gamma) T^{-1/4}$ for every cluster $B \in \Pi$}{
        Exit the outer while loop and remain at the current point for the remaining iterations
    }
    $A \leftarrow \varphi(A;\Pi)$ \tcp*[f]{move to the next group}\;
}
\caption{Cycle-then-Combine Lagged Gradient Descent ({\sc C$^2$-Lgd})}
\label{alg:multi-segment}
\end{algorithm}

See Section \ref{app:n-dim-noiseless} for definitions of the succession function and the feasibility distance, which are used in \textsc{Cycle-then-Combine Lagged Gradient Descent} ({\sc C$^2$-Lgd}) (Alg.~\ref{alg:multi-segment}). This algorithm can be thought of as an inductive extension of \textsc{Ada-Lgd} and \textsc{SCAda-Lgd}, and thus operates similarly. In this section, we further description of the algorithm. The high-level idea of the algorithm is to maintain a clustering (i.e., a partition) of the groups, optimize each cluster separately in coordinate-descent fashion, and combine clusters whenever none can move and doing so would loosen a tight constraint. These ideas are described in more detail below.

{\bf Making gradient-scaled jumps.} As with all other algorithms presented in this paper, gradients are estimated using a lagged and non-lagged point. As is the case with \textsc{Ada-Lgd} and \textsc{SCAda-Lgd}, {\sc C$^2$-Lgd} uses adaptive lag sizes: whenever the local gradient becomes shallow compared to the lag size, the lag size is reduced.

{\bf Switching between clusters.} The algorithm switches from the current cluster to the next when a non-lagged point is reached or the boundary is hit.

{\bf Combining clusters.} Suppose we have reached a point where no cluster can move. If all clusters are at a low gradient, then we have reached an approximate optimum, and no further movements are made. Otherwise, there are clusters at tight constraints. In this can, we argue that there must be a low-gradient cluster which is preventing a tight cluster from moving without violated the EF constraint. In this case, we combine two such clusters and continue.

\section{Strategic sampling sequences}\label{sec:strategic}

In a scenario where decisions are expected to generally become more conducive over time, the various entities receiving these decisions may delay their arrivals. Such delays may introduce biases in the function value estimates constructed by our algorithms under bandit feedback. Note that if the entity delays are uncorrelated with the entity-specific function values (e.g., the delay of a customer under dynamic pricing is uncorrelated with whether or not the customer purchases at any price, i.e., uncorrelated with the customer's willingness to pay for the item), then this is not a concern since the function estimates will continue to be unbiased under such delays. In general, though, these delays could be correlated with the function values, in which case the estimates could be biased. However, we can show that if entities are only willing to delay their arrivals by a short amount of time, then the impact of such strategic behavior on Algorithms \ref{alg:continual-lgd-convex}-\ref{alg:multi-segment} is small. 

We assume a model of strategic behavior in which delays in entity arrivals correspond to permutations of the order of samples. We allow for any adversarial change in the sequence of samples except that the position of each sample can only shift by at most $\Lambda$ slots. We call such an arrival sequence $\Lambda$-strategic arrival sequence.

The main obstacle in directly analyzing the regret of our algorithms under $\Lambda$-strategic arrivals is that the function-value samples are no longer independent, which complicates the use of concentration inequalities. However, since the permutation in question is bounded, we can directly compare the performance of our algorithms on the permuted and unpermuted sequences. In particular, given a point $x$ at which $\ssize(\delta)$ samples are taken, one can show that $\big\vert \overline{f}_{\text{perm}}(x) - \overline{f}(x) \big\vert \in \tilde{\mathcal{O}}(\Lambda/\ssize(\delta))$ with high probability, where $\overline{f}_{\text{perm}}(x)$ is the average over the permuted sequence, and $\overline{f}(x)$ is the average over the unpermuted sequence. This is due to the fact that the averages $\overline{f}_{\text{perm}}(x)$ and $\overline{f}(x)$ only differ on $\mathcal{O}(\Lambda)$ of their samples. This observation then implies accurate gradient estimates (with high probability). Ultimately, one can show that the same regret bound holds for strategic arrivals as with non-strategic arrivals, as long as the maximum shift $\Lambda$ is ``small.''

\begin{theorem}[Regret under strategic arrival sequences] \label{thm:lgd-regret-strategic-behavior}  Assume that (1) $\x^*(i) = \arg\,\min_{x \in \mathbf{R}} f_i(x) \in (x_{\min},x_{\max}]$ for each group $i$, (2) that the noise satisfies Assumption~\ref{asp:noise}, and (3) that arrivals are strategic with delay parameter $\Lambda \in o\big((\log T)^{5/2} \big)$. Consider the modification of Algorithms~\ref{alg:continual-lgd-convex}-\ref{alg:multi-segment} which doubles the sample size: $\ssize(d) = \frac{4 E_{\max}^2 \log \frac{2}{p}}{d^4}$ and keeps all other inputs the same. Under this modification, the algorithms incur $\tilde{\mathcal{O}}(\sqrt{T})$ regret.
\end{theorem}
\begin{proof}{Proof.}
We begin by demonstrating the accuracy of the gradient estimates produced by the modified algorithm. For a given decision $x$, group $i$, and time $t$, the function value observation can be written as $f(x) + \varepsilon_{i,t}$, where $\varepsilon_{i,t}$ has a mean-0, sub-Gaussian distribution. Note that both of these algorithms operate using function value estimates which are obtained by repeatedly sampling at the point in question. Thus, when the noise sequence is permuted, this leaves most of the function value observations at a point $x$ unchanged, with the exception of noise variables whose permuted position belongs to a different decision point. It thus suffices to show that these edge cases do not have a large impact on function value estimates.

The first useful observation is that the sub-Gaussian assumption on noise translates to a high-probability uniform bound on the noise sequence. In particular, with probability at least $1-p$, we have that $|\varepsilon_{i,1}| < \sqrt{\frac{E_{\max}^2 \ln \frac{2}{p}}{C}}$. Since $(1-p)^T \to 1$ for $p = T^{-2}$, this implies that

\begin{equation} \label{eqn:high-probability-noise-bound}
    \mathbb{P} \left( |\varepsilon_{i,t}| < \sqrt{\frac{E_{\max}^2 \ln \frac{2}{p}}{C}} ~\mbox{for all $t \leq T$} \right) \geq (1-p)^T \to 1.
\end{equation}

Using the above, we will bound the function value estimates produced by the algorithm. To do so, consider any two points $a < b$ at which the algorithm will estimate the gradient. Let $t=s,\ldots,s+\ssize(d)-1$ be the epochs at which we sample at $a$, where $d = b-a$. Since the average observed on the unpermuted sequence is accurate (see the proof of Theorem \ref{prop:lgd-noisy-bandit-dynamic-lags}), we will proceed by showing that the average $\overline{f}_{\text{perm}}(a)$ observed on the permuted sequence is close to the average $\overline{f}(a)$ observed on the unpermuted sequence. In particular, let
\[
\overline{f}(a) = f(a) + \frac{1}{\ssize(d)} \sum_{t = s}^{s + \ssize(d) - 1} \varepsilon_{i,t} ~~~~\mbox{and}~~~~\overline{f}_{\text{perm}}(a) = f(a) + \frac{1}{\ssize(d)} \sum_{t = s}^{s + \ssize(d) - 1} \varepsilon_{i,\pi^{-1}(t)}.
\]

\noindent 
Since the (multi)-sets $\{\varepsilon_{i,t}\}_{t=s}^{s + \ssize(d) - 1}$ and $\{\varepsilon_{i,\pi^{-1}(t)}\}_{t=s}^{s + \ssize(d) - 1}$ differ by at most $2\Lambda$ entries, (\ref{eqn:high-probability-noise-bound}) implies:
\[
\big\vert \overline{f}_{\text{perm}}(a) - \overline{f}(a) \big\vert \leq \frac{2 \Lambda}{\ssize(d)} \sqrt{\frac{E_{\max}^2 \ln \frac{2}{p}}{C}} =: A
\]
with high probability. Now note that $A$, ignoring constants, is $\frac{\Lambda d^4}{(\log T)^{1/2}}$. By observing that $d \leq 1/\log T$ for any gap $d$ chosen by the algorithm, one can guarantee that $\big\vert \overline{f}_{\text{perm}}(a) - \overline{f}(a) \big\vert \leq d^2$ with high probability, as long as $\Lambda \in o\big((\log T)^{5/2} \big)$. Additionally, by Lemma \ref{lem:hoeffding}, we have that $| \overline{f}(a) - f(a)| \in \mathcal{O}(d^2)$ with high probability. It follows that
\begin{equation}
\label{eqn:}
\big\vert \overline{f}_{\text{perm}}(a) - f(a) \big\vert \in \mathcal{O}(d^2) ~~\mbox{with high probability.}
\end{equation}

Since this is the same error bound proved under Algorithms~\ref{alg:continual-lgd-convex}-\ref{alg:multi-segment}, the rest of the analysis holds.  
\end{proof}

\section{On the PL assumption}

In Sections~\ref{sec:noiseless}-\ref{sec:noisy}, we considered optimization of smooth and convex functions satisfying the PL inequality (or the stronger assumption of strong convexity). While these assumptions are common in convex optimization, their suitability to applications such as pricing merits discussion. We argue in this section that there are indeed many pricing scenarios where these assumptions may be reasonable. For example, linear demand functions give smooth and strongly concave (SSC) revenue curves.

Moreover, our algorithms only require the assumptions to hold \emph{within the feasible region}. In terms of pricing, the exponential and logit revenue curves are smooth and strongly concave in an open interval around their optima (see, e.g., Figure~\ref{fig:logit-ssc}, which quantifies the smoothness and strong concavity parameters over various feasible regions). Thus if a firm can obtain a narrow price range $\mathcal{P} = [p_{\min}, p_{\max}]$ containing the optimum using data on similar items (or past versions of the same item), then SSC (and, therefore, Assumption~\ref{asp:st-sm}) may be reasonable.

\begin{figure}
\centering
\begin{tikzpicture}
    \def\scale{9}
    \def\shift{7.5}
    \def\econst{2.718}
    
    \draw[teal!7,fill = teal!7] (0,-2) -- (2.4,-2) -- (2.4,3.5) -- (0,3.5) -- cycle;
    
    \draw[<->] (-1, 0) -- (5, 0) node[right] {$p$};
    \draw[<->] (0, -2) -- (0, 3.5) node[above] {Rev};
    
    \foreach[evaluate={\b=int(1*\a)},evaluate={\c=1*\a}] \a in {1,...,4}{
        \draw[black] (\c,-.08) -- (\c,.08);
        \node[black] at (\c,-.25){{\small \b}};
    }
    \draw[scale=1, domain=0:4, smooth, variable=\x, blue] plot ({\x}, {\scale*\x*\econst^(-\x)/(1+\econst^(-\x))});
    \node at (3.25,1.8) {\textcolor{blue}{$\operatorname{Rev}(p)$}};
    \draw[scale=1, domain=1.262:4, smooth, variable=\x, red] plot ({\x}, {
    \scale*2*(\econst^(-2*\x)/((1+\econst^(-\x))^2) - \econst^(-\x)/(1+\econst^(-\x))) + \scale*\x*(\econst^(-\x)*(2*\econst^(-2*\x)/((1+\econst^(-\x))^3) - \econst^(-\x)/((1+\econst^(-\x))^2)) + \econst^(-\x)/(1+\econst^(-\x)) - 2*\econst^(-2*\x)/((1+\econst^(-\x))^2))});
    \node at (2.35,-1.4) {\textcolor{red}{$\operatorname{Rev}''(p)$}};
    \draw[ultra thick,teal,dashed] (0,0) -- node [above=-2pt, sloped] () {\small SSC region} (2.4,0);
    
    \draw[<->] (-1+\shift, 0) -- (5.5+\shift, 0) node[right] {$p_{\max}$};
    \draw[<->] (\shift, -1) -- (\shift, 3.5) node[above] {};
    \foreach[evaluate={\b=int(1*\a)},evaluate={\c=1*\a},evaluate={\d=int(floor(\b/2))},evaluate={\e=int(Mod(5*\b,10))}] \a in {1,...,5}{
        \draw[black] (\c+\shift,-.08) -- (\c+\shift,.08);
        \node[black] at (\c+\shift,-.25){{\small \d.\e}};
    }
    \foreach[evaluate={\b=int(1*\a)},evaluate={\c=1*\a},evaluate={\d=int(floor(\b/5))},evaluate={\e=int(Mod(2*\b,10))}] \a in {1,...,3}{
        \draw (-.08+\shift,\c) -- (.08+\shift,\c);
        \node at (-.4+\shift,\c){{\small \d.\e}};
    }
    \def\scale{-5}
    \def\scalee{.5}
    \draw[scale=1, domain=\shift:4.8+\shift, smooth, variable=\x, violet] plot ({\x}, {\scale*2*(\econst^(-2*\scalee*(\x-\shift))/((1+\econst^(-\scalee*(\x-\shift)))^2) - \econst^(-\scalee*(\x-\shift))/(1+\econst^(-\scalee*(\x-\shift)))) + \scale*\scalee*(\x-\shift)*(\econst^(-\scalee*(\x-\shift))*(2*\econst^(-2*\scalee*(\x-\shift))/((1+\econst^(-\scalee*(\x-\shift)))^3) - \econst^(-\scalee*(\x-\shift))/((1+\econst^(-\scalee*(\x-\shift)))^2)) + \econst^(-\scalee*(\x-\shift))/(1+\econst^(-\scalee*(\x-\shift))) - 2*\econst^(-2*\scalee*(\x-\shift))/((1+\econst^(-\scalee*(\x-\shift)))^2))});
    
    \draw[cyan] (\shift,.5 * 5) -- node [above=0pt, sloped] () {$\beta$} (4.8+\shift,.5 * 5);
    \node[violet] at (2.4+\shift,1.53) {$\alpha$};
\end{tikzpicture}
\caption{(left) Illustration of the revenue $\operatorname{Rev}(p)= p e^{-p}/(1 + e^{-p})$ and the second derivative of revenue derived from the logit demand function $e^{-p}/(1+e^{-p})$. The smooth and strongly concave (SSC) region is indicated by the teal dashed line. (right) The strong concavity parameter $\alpha$ and the smoothness parameter $\beta$ of $\operatorname{Rev}(p)$ over the price space $\mathcal{P}=[0,p_{\max}]$ are plotted over $p_{\max}$.}
\label{fig:logit-ssc}
\end{figure}

That said, if the firm chooses a $p_{\max}$ greater than or equal to the inflection point (thus breaking strong concavity), the monotonic algorithms we propose may still perform well. As we discuss in Section~\ref{sec:noiseless}, only smoothness is required to avoid overshooting the optimum, so there is no risk of overshooting while decreasing prices starting from $p_{\max}$ even if $p_{\max}$ lies outside the region where SSC is satisfied. Of course, our regret guarantees will not translate to this setting since it may take a considerable amount of time for the price path to enter the SSC region in bad cases.

Finally, we note that Assumption~\ref{asp:st-sm} has allowed us to make algorithmic innovations that are expected to be useful in addressing more general scenarios, where it may be possible to obtain good regret guarantees under EFTD by utilizing gradient information, such as potentially the open case of concave and smooth revenue functions. 

\section{On the separability assumption}
\label{app:separability}

All regret bounds proved in this paper assume that the objective function $f$ is separable; that is, $f(\x) = \sum_{i=1}^N f_i(\x(i))$ for some one-dimensional functions $f_1,\ldots,f_N$. In this section, we provide an example to show the necessity of this assumption.

To that end, we will observe the behavior of two algorithms given a nonseparable objective function. The two algorithms are \textsc{Gradient Descent} and \textsc{Monotonic Projected Gradient Descent}, defined below:\\

\begin{tabular}{ll}
    $\x_{t+1} = \x_t - \eta \nabla f(\x_t)$ &~~~~ \textsc{Gradient Descent} \\[.5em]
    $\x_{t+1} = \x_t - \eta \min\{\mathbf{0}, \nabla f(x_t)\}$ &~~~~ \textsc{Monotonic Projected Gradient Descent}\\[.5em]
\end{tabular}

\noindent for some $\eta > 0$. Note that \textsc{Monotonic Projected Gradient Descent} can be described as follows: letting $P_{\x} = \{\mathbf{y} : \mathbf{y} \geq \x\}$, the updates are
\[
\x_{t+1} = \Pi_{P_{\x_t}}\big(\x_t - \eta \nabla f(\x_t)\big).
\]
Thus, \textsc{Monotonic Projected Gradient Descent} is monotonic by design.

\begin{figure}
    \centering
    \includegraphics[width=.5\textwidth]{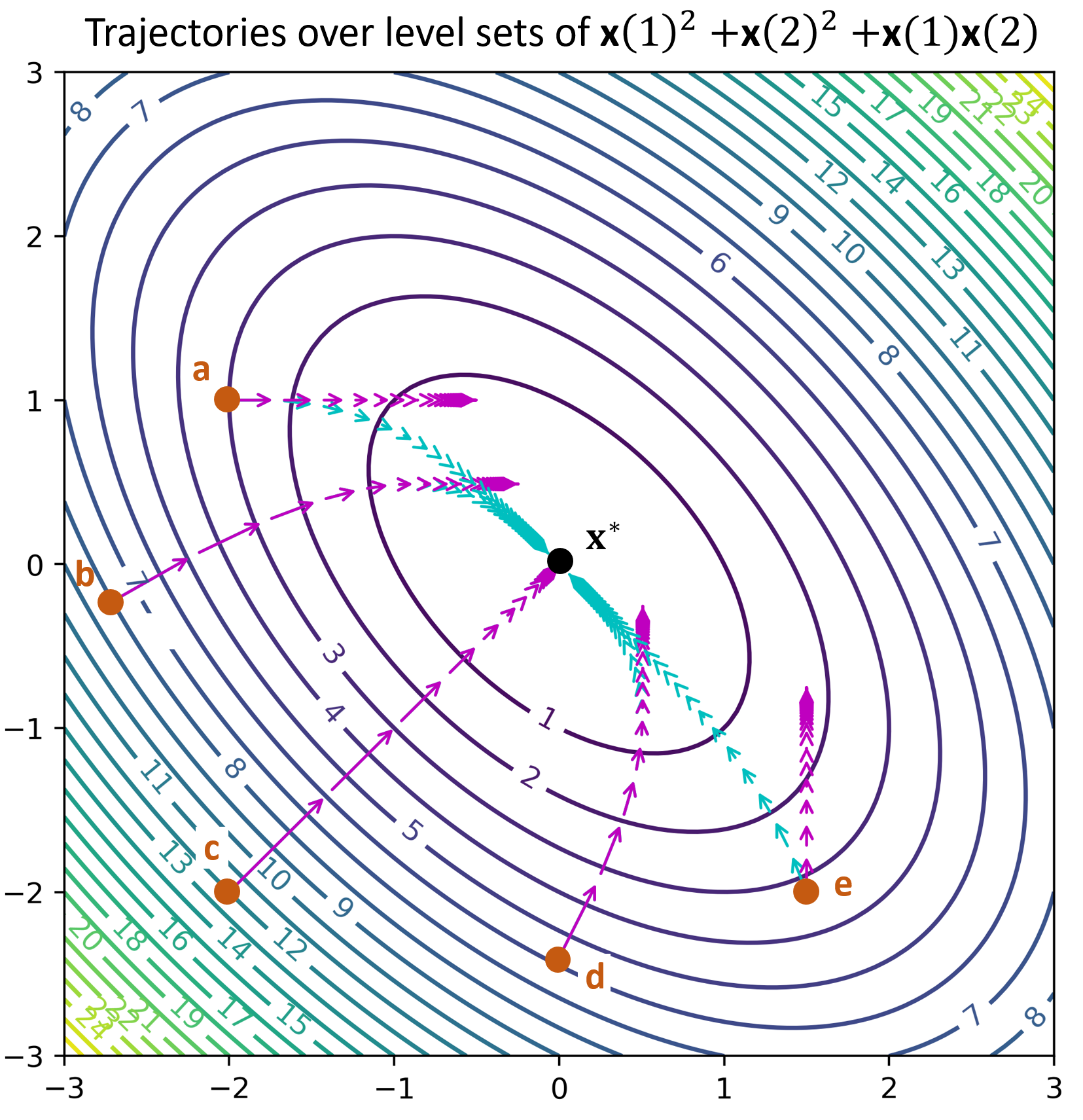}
    \caption{Trajectories of \textsc{Gradient Descent} (blue) and  \textsc{Monotonic Projected Gradient Descent} (purple) with step size $\eta=0.1$. Here, starting at points $\mathbf{b}$ and $\mathbf{d}$ will result in overshooting the optimum $\x^*$, even though $\mathbf{b} \leq \x^*$ and $\mathbf{d} \leq \x^*$.}
    \label{fig:nonseparable-overshoots}
\end{figure}

In Figure~\ref{fig:nonseparable-overshoots}, we run both algorithms with step size $\eta = 0.1$ on the objective function $f(\x) = \x(1)^2 + \x(2)^2 + \x(1)\x(2)$. Note that $0.1 < 1/\beta$ in this case, so both algorithms would converge monotonically to the optimum (given an initial point $\x_1 \leq \x^*$) if the objective function were separable. In this case, however, we see that both algorithms can overshoot the optimum. Notably, both algorithms overshoot the optimum starting from points $\mathbf{b}$ and $\mathbf{d}$, even though $\mathbf{b},\mathbf{d} \leq \x^*$.

\end{document}